\documentclass[11pt]{article}

\usepackage[a4paper,top=2.6cm,bottom=2.8cm,left=2.8cm,right=2.8cm,headsep=0.8cm]{geometry}
\usepackage[T1]{fontenc}
\usepackage[utf8]{inputenc}
\usepackage{amsmath}
\usepackage{libertinus}
\usepackage{libertinust1math}
\usepackage[scaled=0.95]{inconsolata}
\usepackage[final]{microtype}
\usepackage[table]{xcolor}
\definecolor{accent}{HTML}{6B2D5C}
\definecolor{accentlink}{HTML}{1F4E8C}
\definecolor{rulegray}{gray}{0.55}

\usepackage{titlesec}
\titleformat{\section}{\sffamily\Large\bfseries}{\textcolor{accent}{\thesection}}{0.8em}{}
\titleformat{\subsection}{\sffamily\large\bfseries}{\textcolor{accent}{\thesubsection}}{0.8em}{}
\titleformat{\subsubsection}{\sffamily\normalsize\bfseries}{\textcolor{accent}{\thesubsubsection}}{0.8em}{}
\titleformat{\paragraph}[runin]{\sffamily\normalsize\bfseries}{}{0pt}{}
\titlespacing*{\section}{0pt}{3.2ex plus 1ex minus .2ex}{1.4ex plus .2ex}
\titlespacing*{\subsection}{0pt}{2.6ex plus 1ex minus .2ex}{1.1ex plus .2ex}
\titlespacing*{\subsubsection}{0pt}{2.2ex plus .8ex minus .2ex}{0.8ex plus .2ex}
\titlespacing*{\paragraph}{0pt}{1.6ex plus .6ex minus .2ex}{0.8em}

\usepackage[font=small,labelfont={sf,bf,color=accent},labelsep=period,margin=0.5cm]{caption}
\usepackage{fancyhdr}
\fancypagestyle{plain}{\fancyhf{}\fancyfoot[C]{\small\sffamily\thepage}}

\newif\ifblindrev\blindrevfalse
\newcommand{\weavecite}{\citep{campagne2026weave}}
\usepackage{xspace}
\usepackage{scalefnt}
\newcommand{\dsetstyle}[1]{{\scalefont{1.05}\textsc{\MakeLowercase{#1}}}}
\newcommand{\REGIONSFR}{\dsetstyle{REGIONS-FR}\xspace}
\newcommand{\WEAVEUK}{\dsetstyle{WEAVE-UK}\xspace}
\newcommand{\ELECTRICITY}{\dsetstyle{Electricity}\xspace}
\usepackage{enumitem}
\usepackage{url}
\usepackage{tikz}
\usepackage{graphicx}
\usepackage{booktabs}
\usepackage{makecell}
\usepackage{multirow}
\usetikzlibrary{arrows,positioning,shapes}
\usetikzlibrary{shapes.multipart}
\usetikzlibrary{arrows.meta}
\usetikzlibrary{decorations.pathreplacing}
\usepackage{rotating}
\usepackage{natbib}
\bibpunct[, ]{(}{)}{,}{a}{}{,}

\definecolor{darkgreen}{rgb}{0.00,0.70,0.00}
\definecolor{darkblue}{rgb}{0.00,0.00,0.70}
\definecolor{darkred}{rgb}{0.70,0.00,0.00}
\definecolor{bluecb}{rgb}{0.10,0.30,0.75}
\definecolor{nodeblue}{HTML}{3B6FD0}
\definecolor{nodebluef}{HTML}{CFE0F5}
\definecolor{nodegreen}{HTML}{4E9A51}
\definecolor{nodegreenf}{HTML}{D8ECD0}
\definecolor{nodedark}{HTML}{2B4C8F}
\definecolor{gainred}{HTML}{A53A32}
\definecolor{gaingreen}{HTML}{2F7D4A}
\definecolor{blue1}{HTML}{2851CC}
\definecolor{red1}{HTML}{FF0000}
\definecolor{green1}{HTML}{3BB273}
\definecolor{orange1}{HTML}{F96C39}
\definecolor{pink1}{HTML}{F056BF}
\definecolor{gray1}{HTML}{696A62}
\definecolor{gray2}{HTML}{E3E4DB}

\usepackage{dsfont}
\newcommand{\Ind}[1]    {\mathds{1}{\{#1\}}}

\newcommand{\V}{\ensuremath{\mathcal{V}}\xspace}

\newcommand{\EE}{\ensuremath{\mathbb E}\xspace}
\newcommand{\bX}{\ensuremath{\mathbf X}\xspace}
\newcommand{\bA}{\ensuremath{\mathbf A}\xspace}
\newcommand{\bL}{\ensuremath{\mathbf L}\xspace}
\newcommand{\bh}{\ensuremath{\mathbf h}\xspace}
\newcommand{\bW}{\ensuremath{\mathbf W}\xspace}

\newcommand{\rka}[1]{\cellcolor{gray!30}#1}
\newcommand{\rkb}[1]{\cellcolor{gray!18}#1}
\newcommand{\rkc}[1]{\cellcolor{gray!8}#1}
\newcommand{\vr}[1]{{\footnotesize\,($#1$)}}
\newcommand{\worse}[1]{{\footnotesize\textcolor{gainred}{(#1)}}}
\newcommand{\better}[1]{{\footnotesize\textcolor{gaingreen}{(#1)}}}
\newcommand{\Qitem}[2]{\par\noindent\hangindent=2.6em\hangafter=1\makebox[2.6em][l]{(#1)}\emph{#2} }
\newcommand{\Qflat}[2]{\par(#1)~\emph{#2} }

\usepackage{hyperref}
\hypersetup{
  colorlinks=true, breaklinks,
  linkcolor=accentlink, citecolor=accentlink, urlcolor=accent,
  pdftitle={A Diagnostic Protocol for GNNs in Electricity Load Forecasting: Separating Parameter Sharing from Neighborhood Access},
  pdfauthor={Eloi Campagne, Yvenn Amara-Ouali, Yannig Goude, Itai Zehavi, Argyris Kalogeratos},
}

\newcommand{\aff}[1]{\textsuperscript{\,#1}}
\makeatletter
\renewcommand{\maketitle}{%
  \thispagestyle{plain}%
  \begingroup\centering
  {\sffamily\LARGE\bfseries\color{accent}%
   A Diagnostic Protocol for GNNs in\\[0.2ex] Electricity Load Forecasting\par}\vspace{1ex}
  {\sffamily\Large\color{accent}Separating Parameter Sharing from Neighborhood Access\par}\vspace{3ex}
  {\large\mbox{Eloi Campagne\aff{1,2,*}}\quad
   \mbox{Yvenn Amara-Ouali\aff{2}}\quad
   \mbox{Yannig Goude\aff{2}}\quad
   \mbox{Itai Zehavi\aff{1,2}}\quad
   \mbox{Argyris Kalogeratos\aff{1}}\par}\vspace{1.4ex}
  {\small\itshape
   \aff{1}Centre Borelli, École Normale Supérieure Paris-Saclay, Gif-sur-Yvette, France\\
   \aff{2}EDF Lab, Palaiseau, France\\[0.6ex]
   \upshape\aff{*}Corresponding author: \href{mailto:eloi.campagne@ens-paris-saclay.fr}{\texttt{eloi.campagne@ens-paris-saclay.fr}}\par}\vspace{3.5ex}
  \endgroup
  \setcounter{footnote}{0}%
}
\makeatother
\renewenvironment{abstract}{%
  \par\noindent{\color{rulegray}\rule{\linewidth}{0.4pt}}\par\vspace{1ex}
  \begin{list}{}{\leftmargin=1.2cm\rightmargin=1.2cm}\item\relax\small
  {\sffamily\bfseries\color{accent}Abstract.}\enspace\ignorespaces}{%
  \end{list}}
\newcommand{\keywords}[1]{%
  \begin{list}{}{\leftmargin=1.2cm\rightmargin=1.2cm}\item\relax\small
  {\sffamily\bfseries\color{accent}Keywords.}\enspace #1\end{list}
  \vspace{0.6ex}\noindent{\color{rulegray}\rule{\linewidth}{0.4pt}}\par\vspace{2ex}}

\begin{document}

\maketitle

\begin{abstract}
A graph neural network (GNN) forecasting a collection of load series combines two mechanisms, sharing parameters across series and reading neighboring series, and a gain over a non-graph model may come from either; telling them apart decides whether a graph is worth maintaining. We propose a model diagnostic protocol that compares the GNN with three reference models: a per-node network, a network shared across nodes without neighbors, and the same GNN without edges. Degree-preserving rewiring tests whether neighbor identity matters. After a tuned benchmark on regional load in France and on low-voltage feeders in the UK, we apply the protocol to deliberately short node histories, as when non-stationary demand leaves only recent observations relevant: the feeders, within one week, and a public dataset of client consumption series. On the feeders, sharing parameters recovers most of the error gap between node-specific models and GNNs, about two thirds against a regularized one. Adding the edges to an otherwise identical backbone brings a smaller gain, half or more of which disappears under rewiring. On the client series, the neighbor mean degrades the per-node model on the shorter window and the edges' gain is not consistently distinguishable from zero; on the regions of France, where each node's own history is already informative, the ranking between per-node models and GNNs changes with tuning. A synthetic study points to noisy local measurements as one regime where neighbors help. In an exploratory online aggregation, graph and non-graph forecasters appear complementary, and the weight assigned to GNNs is not proportional to their contribution. In none of the three datasets do the edges account for most of a GNN's advantage over node-specific models, so a forecasting gain should be credited to the graph only after comparison with per-node and shared references.
\end{abstract}
\keywords{Graph neural networks, Model diagnostic protocol, Electricity load forecasting, Spatio-temporal models, Local and global forecasting, Parameter sharing}

% ---- one cell of the node-wise family figure: its captions, its matrix shape and its cost -------
% #1 panel center x  #2 panel origin y  #3 title  #4 subtitle
% #5 x at which the matrix shape is set  #6 shape  #7 parameter count  #8 note
% Vertical offsets are fixed here so that the 4 cells align row by row and no
% two lines of text can meet; the widths they must clear were measured once.
\newcommand{\panelframe}[8]{%
  \node[font=\bfseries\footnotesize]                     at (#1, #2+3.62) {#3};
  \node[font=\scriptsize, gray!65!black]                 at (#1, #2+3.26) {#4};
  \node[font=\scriptsize, gray!65!black, anchor=west]    at (#5, #2+1.80) {#6};
  \node[font=\footnotesize\itshape]                      at (#1, #2+0.58) {Parameters:\ \normalfont #7};
  \node[font=\scriptsize, gray!65!black]                 at (#1, #2+0.20) {#8};
}

\section{Introduction}\label{sec:intro}

Electricity demand is measured across many places at once, regions or individual feeders, and these measurements are not independent: weather, calendars, and economic activity register on neighboring meters together. Disaggregated load data are increasingly available \citep{obst2021adaptive, devilmarest2021statespace}, and hierarchical methods already turn them into gains on aggregate demand \citep{bregere2022online, antoniadis2022hierarchical}. A graph neural network (GNN) adds this spatial axis to a temporal model: one shared local rule explains each node through its own features and also those of its neighbors, so a single model covers a whole set of regions or feeders. On the half-hourly datasets, the problem is to forecast the next day at every node from a window of past observations.

In practice, the demand graph is not given: electrical topology, geographic proximity, and statistical
similarity each suggest a different node adjacency, so building a graph is part of the modeling. A second decision is the choice of reference model. Beating a model fitted to national demand does not isolate the contribution of edges: the conclusion depends on choosing the right reference model. Any such comparison assesses the value of a prescribed node adjacency under the chosen model classes, so a negative result concerns those choices only. The question pursued in this article is therefore when a prescribed node adjacency is informative enough to earn the cost of building and maintaining it, and how much of a GNN's gain over node-specific models is already
brought by parameter sharing alone.

\paragraph{Contributions.}
We propose a diagnostic protocol, built over existing architectures, that credits a forecasting gain to a candidate node adjacency only once what parameter sharing already provides has been set aside. The protocol is designed for a practitioner who must decide whether a graph is worth deploying and maintaining. It sets three reference models against the GNN: i) a model fitted separately to each node from the same local covariates, which gives the total gap a graph has to justify; ii) a network shared across nodes on the identity adjacency, without access to neighbors, which measures the edges conditional on sharing; and iii) the GNN itself with its edges removed, which separates the edges from the change of backbone. Comparing these models along both orders of introducing sharing and neighborhood access decomposes the error gap, up to the architectural differences between the MLPs and the GNN, and rewiring the graph (while keeping the degrees unchanged) assesses whether the identity of the neighbors matters beyond their number.

We work with three load datasets spanning geographic regions (\REGIONSFR), feeders (\WEAVEUK), and individual clients (\ELECTRICITY), and a synthetic study where the conditions are controlled. We run the protocol in full on \WEAVEUK and \ELECTRICITY, where we deliberately keep each node's history short, and answer four questions.\par\smallskip
\Qitem{Q1}{Does a GNN forecast load better than strong non-graph models?} It depends on the dataset: in a tuned benchmark of $8$ GNN architectures against statistical, boosting, foundation-model, and persistence baselines, no GNN beats a pooled gradient-boosted model on \REGIONSFR\ifblindrev\else\footnote{An earlier version of the \REGIONSFR benchmark appears in \citet{campagne2026leveraging}.}\fi, whereas on \WEAVEUK only $2$ of the $8$ GNNs have a lower mean RMSE than the best baseline, the zero-shot foundation model t0-gamma, both by less than one seed std.
\Qitem{Q2}{Where a GNN wins, does the gain come from the graph?} On \WEAVEUK, mostly not: sharing parameters across nodes explains most of it.
\Qitem{Q3}{At equal architecture, do neighbors help?} A little: against the same backbone without edges, the edges add a small consistent gain on \WEAVEUK in the larger-dataset comparison, half or more of which the tested rewirings remove; on \ELECTRICITY the gain is not consistently different from zero once seed variability is included.
\Qitem{Q4}{When do neighbors help?} The synthetic study identifies noisy local measurements as one mechanism. On real data, their contribution depends on the dataset and the window, and the protocol measures it case by case.\par\smallskip

An exploratory expert aggregation, which mixes graph and non-graph forecasters online and withdraws each family in turn, suggests that the two families are complementary and shows that the weight a mixture places on the GNNs is not proportional to their contribution.

\paragraph{Organization.}
Section~\ref{sec:related} reviews related work, Section~\ref{sec:preliminaries} places GNNs and node-specific models in one family, Section~\ref{sec:methodo} describes the graphs and architectures, and Sections~\ref{sec:experiments} and~\ref{sec:results} the protocol and the results; an Appendix gives further controls.

\section{Related Work}\label{sec:related}

Generalized additive models remain the operational instrument for conditional-mean load forecasting \citep{fan2010forecast, fan2011short, pierrot2011short}, with quantile and location-scale extensions for the full distribution \citep{fasiolo2021fast, gilbert2023probabilistic}. Their error grows under structural breaks: the Covid-19 crisis \citep{jiang2021impacts} and the European price surge \citep{doumeche2023human} are relevant recent cases, which motivated adaptive variants coupling online learning with filtering \citep{devilmarest2023adaptive}. These models address an aggregated signal, whereas decentralization produces spatially distributed ones \citep{williams2020electricity}. Recurrent architectures capture the temporal dependence of load series \citep{shi2017deep, marino2016building, massaoudi2021deep} without representing relations between nodes, which is the gap GNNs fill; they have carried spatio-temporal forecasting in traffic \citep{guo2019attention, chen2020dynamic} and, more recently, in power systems \citep{jiang2023buaa_bigscity, jiang2024power}.

Two lines bear directly on the comparisons run here. MTGNN \citep{wu2020connecting} and ADLGNN \citep{sriramulu2023adaptive} learn the node adjacency from the data instead of receiving it, and both report electricity benchmarks. The Appendix evaluates MTGNN under the same data splits and training protocol as the larger-dataset comparison. Attention and multi-scale variants pursue the same end by other means, coupling graph attention with recurrence and physical line attributes \citep{orji2025enhanced}, filtering in the frequency domain \citep{mo2025mrgnn}, or updating the topology from temporal context \citep{zhu2025short}. Alongside them, pretrained forecasters now enter the same benchmarks, PriceFM \citep{yu2025pricefm} embedding a graph bias for European price forecasting and the broader family surveyed by \citet{ferdaus2025foundation}, which is why the benchmark of Section~\ref{sec:seq2seq-results} carries foundation models among its baselines. Hybrid designs pursue a related goal from the statistical side, Neural Additive Models \citep{agarwal2021neural} and their probabilistic variants \citep{thielmann2024neural} carrying the feature-wise decomposition of a GAM inside a neural architecture.

Comparisons with aggregated or global non-graph models do not, by themselves, distinguish the value of sharing parameters from that of accessing neighbors. Closest to this question, \citet{montero2021principles} contrast local and global models for groups of series, \citet{shao2022spatial} show that node identity embeddings let a plain multilayer perceptron (MLP) rival spatio-temporal GNNs, and \citet{cini2023taming} study local and global components in spatio-temporal forecasting. The latter work measures how much node-specific capacity a global model needs, supplied through node-specific parameters and embeddings. We measure which part of the gap between a node-specific model and a GNN the graph explains: we decompose the error gap from a node-specific reference model along both orders of introducing parameter sharing and neighborhood access, including a non-shared model given the neighbor mean. To our knowledge, the cited studies do not report this two-path decomposition.

\section{Preliminaries and Background}\label{sec:preliminaries}

A graph $\mathcal{G} = (\mathcal{V}, \mathcal{E})$ on $N$ nodes is represented by a binary adjacency matrix $\bA$ or a weighted one $\bW$, and $\mathcal{N}_v = \{u \in \mathcal{V} \mid W_{uv} > 0\}$ is the neighborhood of node $v$. A message-passing layer \citep{gilmer2017neural} updates each node's representation from a permutation-invariant aggregate (sum, mean, maximum, or attention-weighted mean) of its neighbors' representations, so $L$ layers reach the $L$-hop neighborhood; the general formalism is recalled in Section~\ref{app:mp} of the Appendix. We denote a matrix by a bold capital letter and its elements by italic capitals, e.g. $\bA$ and $A_{ij}$, and a vector by a bold lowercase letter and its elements by italic lowercase letters, e.g. $\mathbf{y}_t$ and $y_{i,t}$. Moreover, $\mathbf{I}$ is the identity matrix, $\mathbf{1}$ is the all-ones vector of dimension $N$, and $\Ind{\,\cdot\,}$ is the indicator function.

\subsection{GNNs and MLPs: One Family of Node-wise Models}\label{sec:gnn-mlp-family}

Each node $i$ carries features $\mathbf{x}_i \in \mathbb{R}^d$, and a model returns a forecast $\hat{\mathbf{y}}_i \in \mathbb{R}^h$. Models from an independent MLP per region to an attentional GNN differ along three design decisions: 
\begin{itemize}
\item[i)] \emph{Parameter sharing}: Either use $N$ node-specific models, or a single model applied to every node. The latter divides the parameter count by $N$ and pools the data of all nodes, at the cost of imposing the same input--output rule on nodes that may differ. \item[ii)] \emph{Neighborhood access}: The support of $\bA$ induces the node adjacency, which marks the nodes whose features are involved in node $i$'s forecast. With $\bA = \mathbf{I}$, only $i$'s own features are used in its forecast; on the other extreme, $\bA = \mathbf{1}\mathbf{1}^\top$ leads to full connectivity; sparse node adjacency lies between those extremes. 
\item[iii)] \emph{Neighborhood aggregation}: a GNN aggregates neighbors symmetrically, so the aggregation is permutation-invariant, whereas an MLP reading concatenated features gives each node a fixed input slot, which is permutation-sensitive.
\end{itemize}

A simple layer illustrates the sharing and access axes:
\begin{align}
\mathbf{H} = \sigma(\bX\bW_1 + \hat{\bA}\,\bX\,\bW_2),
\end{align}
where $\bX \in \mathbb{R}^{N \times d}$ stacks the node features, $\hat{\bA}$ is a normalized adjacency matrix (specifically, $\tilde{\mathbf{D}}^{-1/2}(\bA + \mathbf{I})\tilde{\mathbf{D}}^{-1/2}$ for the graph convolutional network (GCN), the row-normalized $\mathbf{D}^{-1}\bA$ for the mean aggregation of SAGE), $\bW_1$ and $\bW_2$ shared weight matrices, and $\sigma$ an element-wise activation function. Setting $\bA = \mathbf{I}$ reduces this layer to $\mathbf{H} = \sigma(\bX \bW)$ with $\bW = \bW_1 + \bW_2$. For this layer, the identity adjacency $\mathbf{I}$ gives the same function class as a weight-shared MLP acting on each node's own features. This algebraic equivalence does not imply though that arbitrary MLP and residual GNN implementations have the same depth, parameterization or optimization behavior. 

\begin{figure}[tp]
\centering
\scalebox{0.85}{%
\begin{tikzpicture}[
  font=\small,
  >=Stealth,
  own/.style   ={circle, draw=nodegreen, fill=nodegreenf, line width=0.6pt, minimum size=4.0mm, inner sep=0pt},
  other/.style ={circle, draw=nodeblue,  fill=nodebluef,  line width=0.6pt, minimum size=4.0mm, inner sep=0pt},
  off/.style   ={circle, draw=gray!50,   fill=gray!7,     line width=0.6pt, minimum size=4.0mm, inner sep=0pt},
  fnode/.style ={circle, draw=nodegreen, fill=nodegreenf, line width=1.2pt, minimum size=4.8mm, inner sep=0pt},
  blk/.style   ={draw=nodedark, line width=0.7pt, minimum height=4.2mm, inner sep=0pt},
  ghost/.style ={draw=gray!55, line width=0.5pt, minimum height=4.2mm, inner sep=0pt},
  msg/.style   ={line width=0.7pt},
]
% frame, header band, row band, axis separators
\draw[gray!55, line width=0.8pt, rounded corners=2pt] (0,2.30) rectangle (14,11.30);
\fill[gray!8] (0,10.35) rectangle (14,11.30);
\fill[gray!8] (0,2.30) rectangle (1.15,10.35);
\draw[gray!55, line width=0.8pt] (0,10.35) -- (14,10.35);
\draw[gray!55, line width=0.8pt] (1.15,6.40) -- (14,6.40);
\draw[gray!55, line width=0.8pt] (1.15,2.30) -- (1.15,11.30);
\draw[gray!45, dashed, line width=0.6pt] (7.575,2.42) -- (7.575,10.25);

\node[font=\bfseries\footnotesize, nodedark] at (4.36,11.00) {Reads own features only};
\node[font=\scriptsize, gray!65!black]       at (4.36,10.62) {node 3 reads nothing but itself};
\node[font=\bfseries\footnotesize, nodedark] at (10.79,11.00) {Reads other nodes' features};
\node[font=\scriptsize, gray!65!black]       at (10.79,10.62) {every series, or the neighborhood};
\node[font=\bfseries\footnotesize, nodedark, rotate=90] at (0.57,8.38) {MLP, tied to node labels};
\node[font=\bfseries\footnotesize, nodedark, rotate=90] at (0.57,4.33) {GNN, one rule everywhere};

% ---------------- top left: per-node MLP -------------------------------------
\panelframe{4.36}{6.45}{Per-node MLP}{one matrix per node, own features}{4.96}{$d \times H$}{$NdH$}{$N$ copies of one block}
\foreach \k/\st in {1/off,2/off,3/own,4/off,5/off}{%
  \node[\st] (pn\k) at (4.36+\k-3,9.03) {};
  \node[font=\tiny, gray!65!black] at (4.36+\k-3,9.37) {\k};}
% the stack is offset sideways only, so nothing sits above or below the block
\node[ghost, fill=white, minimum width=0.5cm] at (4.56,8.25) {};
\node[ghost, fill=white, minimum width=0.5cm] at (4.46,8.25) {};
\node[blk, fill=nodegreenf, minimum width=0.5cm] (pnW) at (4.36,8.25) {};
\node[anchor=east, font=\footnotesize] at (3.99,8.25) {$\bW_3$};
\draw[->, nodegreen, msg] (pn3) -- (4.36,8.46);
\draw[->, nodedark, msg] (4.36,8.04) -- (4.36,7.66);
\node[fnode] at (4.36,7.53) {};
\node[anchor=west, font=\scriptsize, gray!65!black] at (4.66,7.53) {$\hat y_3$};

% ---------------- top right: all-node MLP -------------------------------------
\panelframe{10.79}{6.45}{All-node MLP}{one matrix, a block per series}{12.19}{$Nd \times H$}{$NdH$}{one copy of $N$ blocks}
\foreach \k/\st in {1/other,2/other,3/own,4/other,5/other}{%
  \node[\st] (gb\k) at (10.79+\k-3,9.03) {};
  \node[font=\tiny, gray!65!black] at (10.79+\k-3,9.37) {\k};}
\foreach \k/\c in {1/nodebluef,2/nodebluef,3/nodegreenf,4/nodebluef,5/nodebluef}{%
  \node[blk, fill=\c, minimum width=0.50cm] (gbW\k) at (10.79+0.50*\k-1.50,8.25) {};}
\node[anchor=east, font=\footnotesize] at (9.42,8.25) {$\bW$};
\foreach \k/\c in {1/nodeblue,2/nodeblue,3/nodegreen,4/nodeblue,5/nodeblue}{%
  \draw[->, \c, msg] (gb\k) -- (gbW\k.north);}
\draw[->, nodedark, msg] (10.79,8.04) -- (10.79,7.66);
\node[fnode] at (10.79,7.53) {};
\node[anchor=west, font=\scriptsize, gray!65!black] at (11.09,7.53) {$\hat y_3$};

% ---------------- bottom left: identity-graph GNN ------------------------
\panelframe{4.36}{2.40}{Identity-graph GNN}{$\bA = \mathbf{I}$, the shared matrix}{4.76}{$d \times H$}{$dH$}{$N$ times fewer than above}
\foreach \k/\st in {1/off,2/off,3/own,4/off,5/off}{%
  \node[\st] (id\k) at (4.36+\k-3,4.98) {};
  \node[font=\tiny, gray!65!black] at (4.36+\k-3,5.32) {\k};}
\node[blk, fill=nodegreenf, minimum width=0.5cm] (idW) at (4.36,4.20) {};
\node[anchor=east, font=\footnotesize] at (3.99,4.20) {$\bW$};
\draw[->, nodegreen, msg] (id3) -- (4.36,4.41);
\draw[->, nodedark, msg] (4.36,3.99) -- (4.36,3.61);
\node[fnode] at (4.36,3.48) {};
\node[anchor=west, font=\scriptsize, gray!65!black] at (4.66,3.48) {$\hat y_3$};

% ---------------- bottom right: GNN --------------------------------
\panelframe{10.79}{2.40}{GNN}{$\bA$ sparse, the shared matrix}{11.44}{$2d \times H$}{$2dH$}{$N/2$ times fewer than above}
\foreach \k/\st in {1/off,2/off,3/own,4/other,5/other}{%
  \node[\st] (gr\k) at (10.79+\k-3,4.98) {};
  \node[font=\tiny, gray!65!black] at (10.79+\k-3,5.32) {\k};}
\node[blk, fill=nodegreenf, minimum width=0.50cm] (grW1) at (10.54,4.20) {};
\node[blk, fill=nodebluef,  minimum width=0.50cm] (grW2) at (11.04,4.20) {};
\node[anchor=east, font=\footnotesize] at (10.17,4.20) {$\bW$};
\draw[->, nodegreen, msg] (gr3) -- (10.54,4.41);
\draw[->, nodeblue,  msg] (gr4) -- (10.94,4.41);
\draw[->, nodeblue,  msg] (gr5) -- (11.14,4.41);
\draw[->, nodedark, msg] (10.79,3.99) -- (10.79,3.61);
\node[fnode] at (10.79,3.48) {};
\node[anchor=west, font=\scriptsize, gray!65!black] at (11.09,3.48) {$\hat y_3$};

% ---------------- legend: three columns, widths cleared by measurement -------
\draw[gray!55, line width=0.6pt, rounded corners=2pt, fill=gray!4] (0,0.75) rectangle (14,2.15);
\node[own]   at (0.45,1.78) {}; \node[anchor=west, font=\scriptsize] at (0.83,1.78) {the features of node 3 itself};
\node[other] at (5.15,1.78) {}; \node[anchor=west, font=\scriptsize] at (5.53,1.78) {another series the model reads};
\node[off]   at (9.85,1.78) {}; \node[anchor=west, font=\scriptsize] at (10.23,1.78) {a series it cannot reach};
\node[fnode] at (0.45,1.18) {}; \node[anchor=west, font=\scriptsize] at (0.83,1.18) {the forecast $\hat y_3$ produced at node 3};
\node[ghost, fill=white, minimum width=0.44cm] at (5.31,1.18) {};
\node[ghost, fill=white, minimum width=0.44cm] at (5.23,1.18) {};
\node[blk,   fill=white, minimum width=0.44cm] at (5.15,1.18) {};
\node[anchor=west, font=\scriptsize] at (5.61,1.18) {one matrix per node predicted};
\node[blk, fill=nodegreenf, minimum width=0.26cm] at (9.72,1.18) {};
\node[blk, fill=nodebluef,  minimum width=0.26cm] at (9.98,1.18) {};
\node[anchor=west, font=\scriptsize] at (10.23,1.18) {one $d \times H$ block per segment};
\end{tikzpicture}}
\caption{The four node-wise models on the sharing (rows) and access (columns) axes. Each cell shows the series reaching the forecast $\hat y_3$, the matrix producing it and its parameter count for $N$ nodes, $d$ features, and $H$ hidden units; each segment is one $d \times H$ block.}
\label{fig:nodewise-family}
\end{figure}
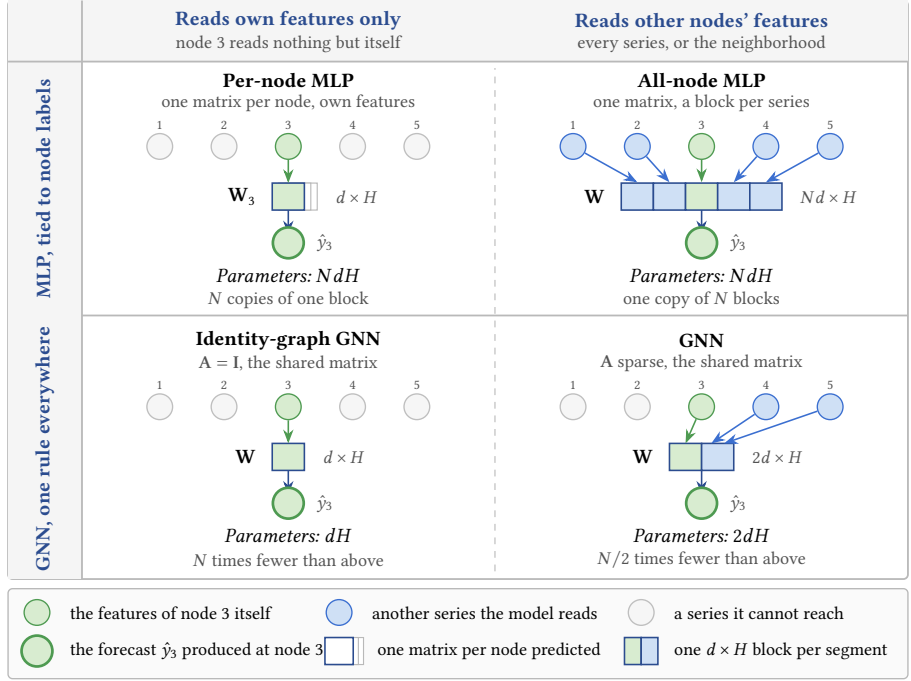

Throughout, \emph{backbone} denotes the neural architecture that surrounds the convolution, and not the graph: two models can share a backbone and differ only in their node adjacency, which is what the controls below exploit. We call \emph{no-graph GNN} the backbone run with no edges between distinct nodes, so that no node reads another, and distinguish its two realizations: the \emph{identity-graph GNN} ($\bA = \mathbf{I}$) and the \emph{empty-graph GNN} ($\bA = \mathbf{0}$). Both span the function class above, since the root term $\bX\bW_1$ keeps each node's own features; under a normalization that adds self-loops the two coincide, and otherwise they differ only in how that map is parameterized. The per-node MLP is recovered by lifting the sharing, replacing the common $\bW$ with node-specific weights $\bW_i$, which gives $\mathbf{H}_i = \sigma(\mathbf{x}_i \bW_i)$. The GNN with a nontrivial neighborhood is recovered by giving $\bA$ the support of a graph, so that row $i$ of $\hat{\bA}\bX$ is a combination of the features of node $i$'s neighbors. Figure~\ref{fig:nodewise-family} groups the $4$ resulting models by family and situates them on the two axes, the coincidence between the shared MLP and the identity-graph GNN appearing there as two readings of a single computation.

Comparing a GNN on a prescribed node adjacency with the identity-graph GNN measures the contribution of the graph edges when both models share their parameters across nodes. Sharing imposes the same input--output rule on every node, and part of what the edges add may only correct that constraint, so this comparison does not separate the information carried by the neighbors from that correction. The per-node MLP neither shares parameters nor reads neighbors, and it is our reference to compare graph-based models.

\section{Methodology}\label{sec:methodo}

The diagnostic protocol takes the graph construction and the GNN architecture as inputs; this section sets out those used here. A GNN is specified by the adjacency it reads and by the convolution that reads it; forecasts from several seeds or architectures are then combined.

\subsection{Graph Construction and Selection}\label{sec:graph-construction}

Typically, the demand graph is not given with the data: nodes are regions or feeders, and an edge must encode a dependence that drives demand, which the electrical wiring explains only in part. The spatial construction kernelizes the geodesic distance, $\mathbf{d}(u,v) = \exp\{-\mathbf{dist}(u,v)^2/\sigma_b^2\}$, with $\sigma_b$ being here the median pairwise distance. On \REGIONSFR, we keep only the closest pairs,
\begin{align}
    (\boldsymbol{W}_\lambda)_{uv}
    = \mathbf{d}(u,v) \cdot \Ind{\mathbf{d}(u,v) \geq \lambda}, \qquad u,v \in \V,
\end{align}
where $\lambda$ is the largest level that keeps the graph connected \citep{williams2020electricity}. As weather systems and regional activity drive demand and are shared by adjacent regions, the idea is to drop the distant pairs that would mostly dilute the neighbor mean, while keeping the graph connected.
The considered data-driven constructions use the DTW distance, the Pearson correlation, and the precision matrix of the load series; DistSplines weights an edge by the $L^2$ distance between the two nodes' fitted temperature-to-load splines, and GL3SR \citep{humbert2021learning} learns a smooth graph from the data. The graph is selected based on the validation RMSE.

The spatial graph keeps $28$ of the $132$ ordered pairs of regions. The data-driven constructions are used as computed, DTW keeping $130$ pairs and the others all $132$, so a convolution that reads edge weights (GCN, ChebConv, TAGConv, APPNP) sees a different operator under each of them, whereas one that reads only the support (GraphSAGE, GAT, GATv2, TransformerConv) sees the same complete graph.

\subsection{Architectures and Aggregation}\label{sec:gnn-arch}\label{sec:aggregation}

Our benchmark covers GCN \citep{kipf2017semisupervised}, GraphSAGE (SAGE) \citep{hamilton2017inductive}, APPNP \citep{gasteiger2018predict}, GAT \citep{velivckovic2018graph}, GATv2 \citep{brody2021attentive}, TransformerConv (Transformer) \citep{shi2020masked}, ChebConv \citep{defferrard2016convolutional} and TAGConv (TAG) \citep{du2017topology}, all inside the same residual backbone: a linear node encoder, residual blocks with layer normalization around each convolution, and a linear readout of the $48$ outputs. The forecasts of several seeds or architectures are combined either uniformly or by the ML-Poly algorithm \citep{gaillard2014second}, which sets the weights $p_{j,t}$ of the aggregate $\widehat y_t = \sum_{j=1}^K p_{j,t} x_{j,t}$ from past performance. A seed ensemble, which averages the forecasts of several seeds \citep{lakshminarayanan2017simple}, captures the variability due to initialization and optimization only, so the spread across seeds does not measure total predictive uncertainty.

\section{Experiments}\label{sec:experiments}

\paragraph{Design.}
The experiments evaluate GNNs at two levels:
\begin{itemize}
\item[i)] The \emph{benchmark level} asks how well architecture--graph pairs forecast after tuning, taken \emph{individually}, in \emph{seed ensembles}, and in \emph{mixtures of architectures} (Section~\ref{sec:seq2seq-results}).
\item[ii)] The \emph{diagnostic level} attributes a gain (Sections~\ref{sec:mlp-reference} to \ref{sec:real-regimes}): it fits prescribed model classes under common training rules and varies parameter sharing, neighborhood access and the identity of the neighbors across training volumes and synthetic regimes, the empty-graph GNN separating the edges from the backbone and degree-preserving rewiring separating the identity of the neighbors from their number.
\end{itemize}
Benchmark scores rank forecasters without attributing their gains; diagnostic scores attribute gains within the specified model classes without full hyperparameter tuning, and meet the benchmark only in Section~\ref{sec:mlp-reference}, where tuning changes the ranking. Outside both levels, a separate exploratory analysis aggregates older GNN configurations online with non-graph forecasters (Section~\ref{sec:aggreg-summary}). Table~\ref{table:experiment-map} lists the blocks and their fixed or distinctive settings.

\begin{table}[tp]
\centering
\caption{The experimental blocks presented in Section~\ref{sec:results} that address the questions of Section~\ref{sec:intro}.}
\label{table:experiment-map}
\footnotesize
\begin{tabular}{>{\raggedright\arraybackslash}m{2.6cm}>{\raggedright\arraybackslash}m{2.4cm}>{\raggedright\arraybackslash}m{1.8cm}>{\raggedright\arraybackslash}m{3.6cm}>{\raggedright\arraybackslash}m{2.6cm}}
\toprule
Block & Datasets & \makecell[l]{Training\\window} & Models & Fixed or distinctive settings \\
\midrule
Architecture benchmark \qquad (\S\ref{sec:seq2seq-results}; Q1) & \REGIONSFR; \WEAVEUK & $1090$ d; $9$ d & $8$ tuned GNNs, non-graph baselines, GNN and shared-MLP seed ensembles, and GNN mixtures & Split and horizon within each dataset; inputs as specified \\
Per-node reference (\S\ref{sec:mlp-reference}; Q1, Q4) & \REGIONSFR & $54$--$1090$ d & Per-node MLP ($\pm$ neighbor mean) and all-node MLP; identity-graph and SAGE GNN & SAGE and hidden width \\
Synthetic regimes (\S\ref{sec:synth-regimes}; Q4) & Synthetic \ \ \ \ denoising data & Training fraction varied & Same 4 node-wise models & Base generator, one factor varied \\
Larger datasets (\S\ref{sec:real-regimes}; Q2--Q4) & \ELECTRICITY; \WEAVEUK & $11$/$60$ d; \qquad $30$ h/$10$ d & Per-node, shared, all-node MLP; per-node and shared MLP $+$ neighbors; SAGE/ChebConv GNN, empty-graph GNN, rewired graphs & Hidden width $64$ \\
Complementarity (\S\ref{sec:aggreg-summary}; exploratory) & \REGIONSFR; synthetic data & Jan $2017$--Jun $2018$ & $6$ GNNs, feedforward network, additive model, ML-Poly mixture & Own split, $1$ seed per expert \\
\bottomrule
\end{tabular}
\end{table}

\subsection{Datasets}\label{sec:datasets}

\paragraph{\REGIONSFR.}
The French network operator (RTE) publishes the half-hourly load of the $12$ French administrative regions, from which we use the period January $2015$ to December $2019$ and call this dataset \REGIONSFR. Covariates are calendar variables (time of day, weekday, day type, holidays, position in the year, daylight saving) and weather conditions (temperature, cloud cover, wind) measured at $32$ stations and averaged within each region; no weather forecast is used. The target is each region's load in MW (megawatts), and system-wide scores are computed on their sum.

\paragraph{\WEAVEUK.}
This dataset contains the half-hourly smart-meter load of low-voltage feeders, as released by the UK data platform (Weave\footnote{\url{https://weave.energy/}}). We use $500$ feeders of Scottish and Southern Electricity Networks \weavecite{}, recorded February $14$--$28$, $2024$ and selected to have no missing values and to span heterogeneous load profiles. Only calendar variables and lagged load are available.

\paragraph{\ELECTRICITY.}
This dataset \citep{zhou2021informer} records the hourly consumption of $321$ clients from July $2016$ to July $2019$. Section~\ref{sec:real-regimes} uses $319$ of them: the client that the original benchmark designates as its forecasting target is left out, and so is one client whose consumption is constant over the selected period.

\subsection{Experimental Settings}\label{sec:exp-setting}

\paragraph{Sequence-to-sequence forecasting.}
The model maps the previous day to the next day's load profile, $\hat{\mathbf{y}}_{t:t+h-1} = \Phi_{\boldsymbol{\theta}}(\mathcal{G},\, \tilde{\mathbf{X}}_t \cup \mathbf{y}_{t-w:t-1})$ with $w = h = 48$ half-hours. Training, validation and testing use nonoverlapping windows of stride $h$, so every origin falls on a day boundary and output $k$ always denotes slot $k$ of the day (Figure~\ref{fig:seq2seq} in the Appendix); every model of the benchmark, graph or not, is trained and scored under this stride. Features and targets are scaled per node by min--max statistics of the training period (the MLP and gradient-boosted baselines of \WEAVEUK standardize their targets instead), and parameters minimize the squared error with Adam \citep{kingma2014adam}. On \REGIONSFR, the periods are $2015$--$2017$ for training, $2018$ for validation, and $2019$ for testing; on \WEAVEUK they are February $14$--$22$, $23$--$24$ and $25$--$28$, $2024$ (Figure~\ref{fig:dataset-splits} in the Appendix).

\paragraph{Evaluation metrics.}
With $s_t = \sum_{i=1}^N y_{i,t}$ and $\hat{s}_t = \sum_{i=1}^N \hat{y}_{i,t}$ the observed and forecast total sums, and $e_t = s_t - \hat{s}_t$ their difference, the two metrics used are:
\begin{align}
\text{MAPE} = \frac{100}{T} \sum_{t=1}^T \left|\frac{e_t}{s_t}\right|,
\qquad
\text{RMSE} = \sqrt{\frac{1}{T}\sum_{t=1}^T e_t^2}.
\end{align}
Section~\ref{sec:real-regimes} adds a per-node score that removes the cancellation the sum allows. Unless otherwise stated, tables report the seed-averaged error, each seed being scored separately and the errors then averaged, with standard deviations (std: sample std, divisor $n-1$) in parentheses; gray shades mark the three best entries (darker is better) of each block and column, or of each row in Tables~\ref{table:data-regime}-\ref{table:node-count}; the means are ranked without testing for significance.

\paragraph{Baselines.}
\emph{Bottom-local} fits one model per node and \emph{bottom-global} one model on the stacked samples of all nodes, each node reading its own features. The baselines are ARIMA, the additive model (per node, or on the aggregate load), per-node MLPs and shared MLPs over $10$ seeds, XGBoost under both schemes, the Chronos-Bolt variant of Chronos \citep{ansari2024chronos} zero-shot on the aggregate load, Chronos-2 \citep{ansari2025chronos2}, t0-gamma \citep{meyer2026t0}, and Timestral \citep{mistral2026timestral}, in its $209$M-parameter version (Timestral-209M), which read covariates, an ARIMA with exogenous inputs (ARIMA-X), and persistence at $D-1$ and $D-7$. The additive model follows the half-hour-specific GAM design of \citet{fan2011short} and \citet{pierrot2011short}, one model per slot, and is an implementation for this benchmark, not a production model: on \REGIONSFR it smooths current and smoothed temperatures, position in the year, cloud cover, and wind, with weekday, holiday, day type, and weekend as factors; on \WEAVEUK it reduces to a linear model per half-hour in the weekday and the previous day's load. On \REGIONSFR, ARIMA forecasts recursively over the horizon; on \WEAVEUK one ARIMA$(4,0,2)$, its order selected on the aggregate, is fitted per half-hour on the daily subseries and forecasts $1$ day ahead with frozen parameters. Chronos-Bolt (small) reads target history alone, its point forecast being the median of its quantiles, and on \WEAVEUK its context runs from the start of training to the forecast origin; Chronos-2, zero-shot on the same summed load and origins, also reads the calendar and, on \REGIONSFR, the weather as past and future covariates, the weather being observed values. t0-gamma and Timestral receive the same inputs and origins, with a context of at most the last $1024$ steps, the most t0-gamma reads in one pass. ARIMA-X is specified in Section~\ref{app:gnn-hp} of the Appendix. \REGIONSFR models read calendar, weather, and the $48$ load lags, except the additive model, whose covariates are listed above; \WEAVEUK models read calendar and lags only. The \REGIONSFR additive model, ARIMA, Chronos-Bolt and persistence rows are untuned and come from an earlier run on the same $2019$ test days and stride. The shared MLPs and per-node MLPs are plain feedforward networks without the residual backbone of the GNNs.

\paragraph{Tuning.}
Optuna \citep{akiba2019optuna} tunes every architecture with $40$ trials scored on the validation RMSE of the summed forecast, the graph construction being searched jointly with the other hyperparameters; the MLPs run $10$ seeds, the gradient-boosted models and GNNs $3$. The search spaces, the settings of the untuned baselines and the retained configurations are given in Section~\ref{app:gnn-hp} of the Appendix. The diagnostic comparisons use prescribed model classes without full hyperparameter searches, the limited validation-based selections being specified in each block; their widths and backbones are stated there, and common width should not be read as matched total capacity.

\section{Results}\label{sec:results}

\subsection{Architecture Benchmark}\label{sec:seq2seq-results}

\paragraph{Individual architectures.}
On \REGIONSFR, the $8$ GNNs span $95$~MW, from $891$~MW for GCN to $986$~MW for APPNP, against a seed dispersion of up to $37$~MW (Table~\ref{table:combined-results}). None reaches the gradient-boosted model fitted on the stacked nodes, at $859$~MW and $1.16\%$; the per-node MLP, at $922$~MW, precedes $6$ of the $8$, and the shared MLP, at $937$~MW, only APPNP. On the $500$ feeders, where each GNN learns from $9$ training days, the best baseline is t0-gamma, zero-shot, at $144.6$~kWh and $3.51\%$, the lowest MAPE of the table. Only GATv2, at $140.6$~kWh, and APPNP, at $142.8$~kWh, have a lower mean RMSE, both by less than one seed std, and Transformer follows at $146.2$~kWh. SAGE, GCN, GAT, and ChebConv lie between $156.0$ and $167.3$~kWh, around Timestral at $158.1$~kWh and ahead of Chronos-2 at $170.2$~kWh; TAG trails at $203.7$~kWh. A neighborhood is not needed to reach the level of Chronos-2: the ChebConv retained by validation has order $K = 1$, which propagates nothing, and still reaches $167.3$~kWh inside the residual backbone the GNNs share, level with Chronos-2 within its seed std.

Pooling separates the two datasets. On the regions, the per-node MLP leads the shared one ($922$ against $937$~MW), whereas pooling helps gradient boosting ($859$ against $913$~MW); on the feeders every pooled model beats its per-node counterpart ($236.5$ against $325.2$~kWh for the MLP, $263.5$ against $288.0$~kWh for gradient boosting). The ranking of convolutions does not carry over: APPNP is last on the regions and among the first three on the feeders, where those three lie within the seed dispersion.

\begin{table}[tp]
\newsavebox{\tabAggBox}\newsavebox{\tabCombBox}
\sbox{\tabAggBox}{\resizebox{0.49\linewidth}{!}{%
\begin{tabular}{llcc|cc}
\toprule
\multirow{2}{*}{Architecture} & \multirow{2}{*}{Combination} & \multicolumn{2}{c|}{\REGIONSFR, $12$ regions} & \multicolumn{2}{c}{\WEAVEUK, $500$ feeders} \\
 & & RMSE (MW) & MAPE (\%) & RMSE (kWh) & MAPE (\%) \\
\midrule
\multicolumn{6}{c}{\emph{Seed ensembles, $3$ seeds of one architecture}} \\
\midrule
GCN         & Uniform & \rkb{$832$} & \rkc{$1.09$} & $153.9$ & $3.76$ \\
            & Online  & \rkc{$838$} & $1.10$ & $153.7$ & $3.66$ \\
SAGE        & Uniform & $863$ & $1.13$ & $147.0$ & $3.97$ \\
            & Online  & $882$ & $1.14$ & $149.0$ & $4.10$ \\
GAT         & Uniform & \rka{$823$} & \rka{$1.06$} & $156.3$ & $4.05$ \\
            & Online  & $841$ & \rkb{$1.08$} & $159.7$ & $4.33$ \\
GATv2       & Uniform & $861$ & $1.11$ & \rka{$134.0$} & \rka{$3.24$} \\
            & Online  & $864$ & $1.11$ & $139.9$ & $3.65$ \\
Transformer & Uniform & $865$ & $1.12$ & \rkc{$139.2$} & $3.70$ \\
            & Online  & $884$ & $1.14$ & \rkb{$136.4$} & \rkb{$3.59$} \\
TAG         & Uniform & $866$ & $1.14$ & $190.7$ & $5.17$ \\
            & Online  & $874$ & $1.15$ & $196.5$ & $5.69$ \\
ChebConv    & Uniform & $867$ & $1.12$ & $156.1$ & \rkc{$3.64$} \\
            & Online  & $864$ & $1.11$ & $162.4$ & $4.01$ \\
APPNP       & Uniform & $922$ & $1.23$ & $139.7$ & $3.70$ \\
            & Online  & $919$ & $1.23$ & $141.8$ & $3.77$ \\
\midrule
\multicolumn{6}{c}{\emph{Mixtures of all $8$ architectures}} \\
\midrule
All & Uniform & \rkc{$814$} & \rkc{$1.05$} & \rka{$144.2$} & \rka{$3.61$} \\
    & Bottom  & \rka{$809$} & \rka{$1.04$} & \rkb{$145.0$} & \rkc{$3.70$} \\
    & Top     & \rkb{$809$} & \rkb{$1.04$} & \rkc{$145.8$} & \rkb{$3.70$} \\
\bottomrule
\end{tabular}
}}
\sbox{\tabCombBox}{\resizebox*{!}{\dimexpr\ht\tabAggBox+\dp\tabAggBox\relax}{%
\begin{tabular}{lcc|cc}
\toprule
\multirow{2}{*}{Model} & \multicolumn{2}{c|}{\REGIONSFR, $12$ regions} & \multicolumn{2}{c}{\WEAVEUK, $500$ feeders} \\
 & RMSE (MW) & MAPE (\%) & RMSE (kWh) & MAPE (\%) \\
\midrule
\multicolumn{5}{c}{\emph{Baseline models}} \\
\midrule
ARIMA               & $4527$ & $5.80$ & $860.6$ & $18.66$ \\
ARIMA-X             & $1333$ & $1.77$ & $237.9$ & $5.90$ \\
GAM (aggregate)     & $1228$ & $1.69$ & $223.5$ & $6.27$ \\
GAM (per node)      & $1200$ & $1.67$ & $329.4$ & $10.70$ \\
Shared MLP          & $937$\vr{26} & $1.28$\vr{0.06} & $236.5$\vr{15.4} & $6.62$\vr{0.32} \\
Per-node MLP        & \rkc{$922$\vr{10}} & \rkc{$1.24$\vr{0.02}} & $325.2$\vr{9.3} & $9.41$\vr{0.54} \\
XGBoost (b-global)  & \rka{$859$\vr{0.4}} & \rka{$1.16$\vr{0.00}} & $263.5$\vr{1.1} & $7.67$\vr{0.03} \\
XGBoost (b-local)   & \rkb{$913$} & \rkb{$1.20$} & $288.0$ & $8.51$ \\
Chronos-Bolt        & $2408$ & $2.99$ & $237.4$ & $6.02$ \\
Chronos-2           & $1767$ & $1.94$ & \rkc{$170.2$} & \rkc{$3.76$} \\
t0-gamma            & $1724$ & $1.83$ & \rka{$144.6$} & \rka{$3.51$} \\
Timestral-209M      & $1864$ & $2.00$ & \rkb{$158.1$} & \rkb{$3.75$} \\
Persistence--1 d    & $4507$ & $5.78$ & $272.1$ & $6.01$ \\
Persistence--7 d    & $4656$ & $6.03$ & $283.6$ & $9.42$ \\
\midrule
\multicolumn{5}{c}{\emph{Graph-based models}} \\
\midrule
GCN         & \rka{$891$\vr{26}} & \rkb{$1.19$\vr{0.03}} & $161.9$\vr{15.5} & $4.14$\vr{0.43} \\
SAGE        & $925$\vr{11} & $1.22$\vr{0.04} & $156.0$\vr{11.7} & $4.37$\vr{0.40} \\
GAT         & \rkb{$899$\vr{37}} & \rka{$1.18$\vr{0.04}} & $166.8$\vr{14.1} & $4.52$\vr{0.59} \\
GATv2       & \rkc{$923$\vr{32}} & $1.21$\vr{0.06} & \rka{$140.6$\vr{15.8}} & \rka{$3.56$\vr{0.21}} \\
Transformer & $931$\vr{9} & $1.23$\vr{0.02} & \rkc{$146.2$\vr{14.1}} & \rkc{$4.00$\vr{0.34}} \\
TAG         & $929$\vr{30} & $1.24$\vr{0.04} & $203.7$\vr{25.2} & $5.77$\vr{0.55} \\
ChebConv    & $924$\vr{18} & \rkc{$1.20$\vr{0.04}} & $167.3$\vr{5.1} & $4.15$\vr{0.14} \\
APPNP       & $986$\vr{27} & $1.33$\vr{0.05} & \rkb{$142.8$\vr{5.4}} & \rkb{$3.81$\vr{0.14}} \\
\bottomrule
\end{tabular}
}}
\begin{minipage}[t]{0.49\linewidth}
\centering
\caption{Test error of the individual models: mean across seeds and std in parentheses, none for deterministic models.}
\label{table:combined-results}
\usebox{\tabCombBox}
\end{minipage}\hfill
\begin{minipage}[t]{0.485\linewidth}
\centering
\caption{Combined GNN forecasts from Table~\ref{table:combined-results}: seed ensembles of one architecture (top) and mixtures of the $8$ architectures, each first averaged over its seeds (bottom).}
\label{table:aggregation-results}
\usebox{\tabAggBox}
\end{minipage}
\end{table}

\paragraph{Seed ensembles and mixtures.}
A seed ensemble averages the forecasts of the $3$ seeds of one architecture, while a mixture combines the $8$ architectures, each first averaged over its seeds; \emph{Bottom} mixes node by node before summation and \emph{Top} mixes the summed forecasts, and online weights are updated three times (after each of the $4$ days of the \WEAVEUK test). On the regions, the seed ensemble of an architecture gains $58$--$76$~MW, GAT from $899$ to $823$~MW, and mixing the $8$ architectures node by node reaches $809$~MW, $82$~MW below the best seed-averaged error of an individual architecture and $14$~MW below the best seed ensemble (Table~\ref{table:aggregation-results}). On both datasets, several combined GNN forecasts finish ahead of every individual model of Table~\ref{table:combined-results}; the non-graph models were not mixed, and only the shared MLP is seed-ensembled ($10$ seeds). On the feeders, the seed ensemble improves every architecture, GATv2 from $140.6$ to $134.0$~kWh, but no mixture of the $8$ passes that average, the best, uniform, reaching $144.2$~kWh. Mixing node by node and mixing the summed forecast are level on both datasets ($0.3$~MW and $0.8$~kWh), and online reweighting improves on the uniform seed ensemble for $2$ of the $8$ architectures on each dataset. The \WEAVEUK benchmark tests on $4$ days, so its rankings rest on $4$ daily errors of one aggregate: the seed ensemble of each GNN architecture beats the $10$-seed ensemble of the shared MLP on $3$ of them.

\subsection{A Per-Node Reference Model for the Contribution of the Graph}\label{sec:mlp-reference}

\begin{figure}[tp]
    \centering
    \includegraphics[width=\linewidth]{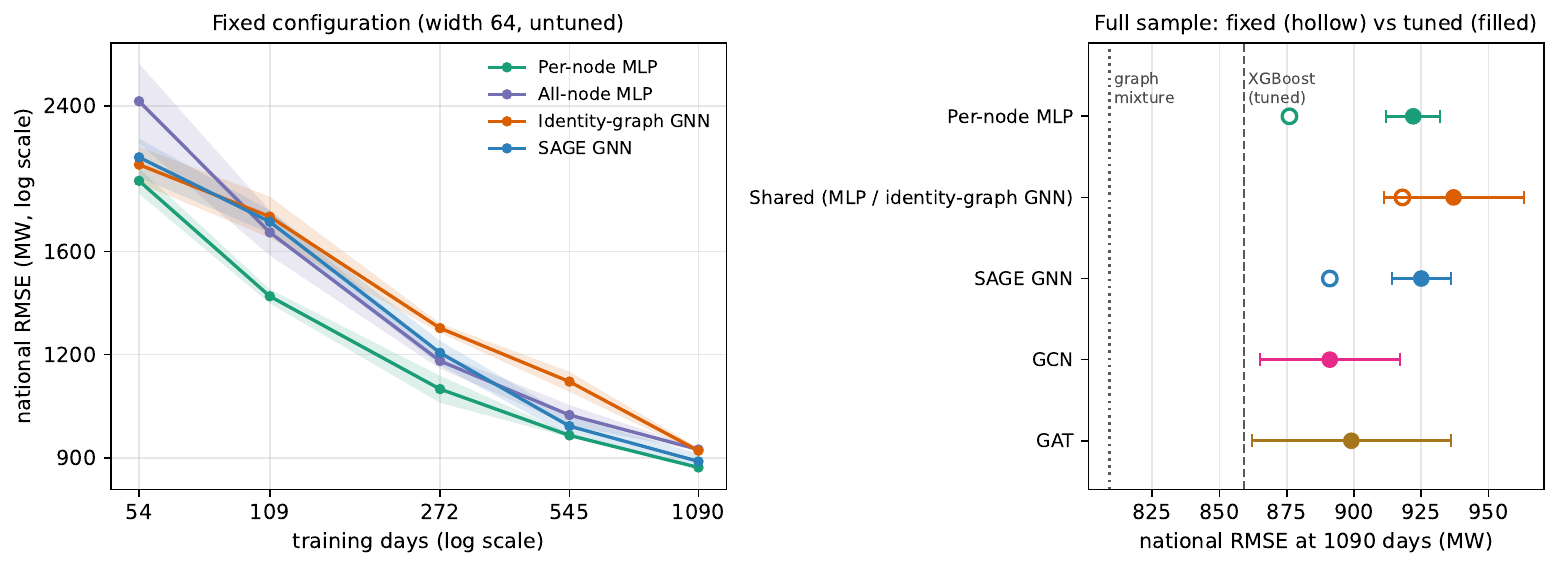}
    \caption{\REGIONSFR. \emph{Left:} national RMSE of the $4$ node-wise models at one fixed configuration (width $64$, untuned) as the training set contracts, $3$ seeds (Table~\ref{table:data-regime}, Appendix). \emph{Right:} at the full sample, each family at the fixed configuration (hollow) and tuned (filled, Table~\ref{table:combined-results}), beside the tuned pooled XGBoost and the bottom-up graph mixture.}
    \label{fig:french-regime}
\end{figure}

\paragraph{Setup.}
On \REGIONSFR, the $4$ node-wise models (Figure~\ref{fig:nodewise-family}) are also fitted at one fixed configuration, without tuning, with a common width, optimizer, and stopping rule: the MLPs have $2$ hidden layers of width $64$ and read features standardized on the training days, the identity-graph GNN and SAGE share the residual backbone of the benchmark at two blocks of width $64$, and all stop on validation (Adam, learning rate $10^{-3}$, max $300$ epochs, patience $75$). The identity-graph GNN and SAGE differ only in the edges; a contrast between an MLP and a GNN also changes the architecture. Figure~\ref{fig:french-regime} sets them beside the tuned models of Table~\ref{table:combined-results}.

\paragraph{Findings.}
At the full sample, the fixed per-node MLP reaches $876 \pm 3$~MW, the identity-graph GNN $918 \pm 7$~MW, SAGE on the spatial graph $891 \pm 19$~MW, and the all-node MLP $921$~MW, and giving the per-node MLP the mean of its neighbors' features yields about $871$~MW; none of $20$ graph and convolution configurations at this capacity reaches the per-node MLP, and the per-node error gives the same ranking ($0.1259$ for the per-node MLP, $0.1272$ for SAGE, $0.1326$ for the identity-graph GNN, and $0.1427$ for the all-node MLP). The ranking does not survive tuning: the tuned GCN and GAT, at $891$ and $899$~MW, lead the tuned per-node MLP at $922$~MW, and selection on the $2018$ validation year leaves the per-node MLP and SAGE worse on the $2019$ test than their fixed configuration. Differences of $30$--$45$~MW between families therefore depend on the configuration. \REGIONSFR supports no stable ranking between node-specific models and GNNs: the tuned per-node MLP falls within the $891$--$986$~MW range of the tuned GNNs, all above the pooled gradient-boosted model at $859$~MW and the GNN mixture at $809$~MW. The $48$ load lags also narrow the spread across GNN configurations, from $586$~MW with calendar and weather alone to about $120$~MW (validation, $1$ seed; Table~\ref{table:feature-arms}, Appendix).

At the fixed configuration, contracting the training set from $1090$ to $54$ days, drawn at random so that seasonal coverage stays complete, keeps the per-node MLP lowest at every volume; restricting each node to its own features and its own weights, the per-node against the all-node MLP, saves $484$~MW at $54$ days and $45$~MW at the full sample. Since the days are sampled across the full training period, this does not establish the ranking for a contiguous short record.

\subsection{Using Synthetic Data to Study the Contribution of the Graph}\label{sec:synth-regimes}

\begin{figure}[tp]
    \centering
    \includegraphics[width=0.9\linewidth]{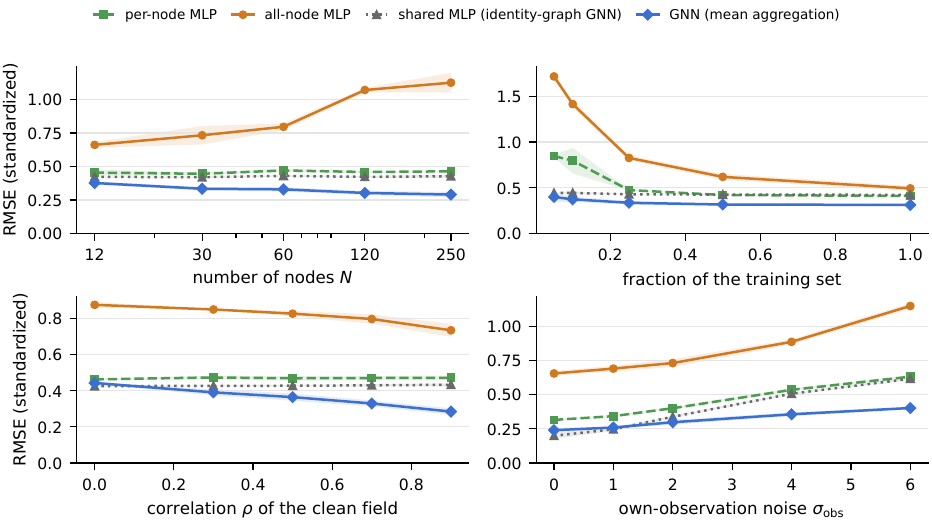}
    \caption{RMSE over nodes and test snapshots, in standardized load units, of the $4$ node-wise models on the synthetic denoising data, $3$ seeds per point with the spread shaded; each plot varies one factor around the base configuration.}
    \label{fig:synth-regimes}
\end{figure}

\paragraph{Setup.}
A denoising generator locates the regime in which the ranking can turn. A clean temperature field is drawn over a geometric graph of $N$ nodes with spatial correlation $\rho$; each node observes it through independent noise of level $\sigma_\text{obs}$, and its load is a real regional temperature-to-load spline applied to the clean field. The neighbors, whose clean temperatures are correlated and whose noise is independent, can therefore denoise a node's reading. The base configuration has $60$ nodes, $\rho = 0.7$, $\sigma_\text{obs} = 3$, and $30\%$ of the training set, and each plot of Figure~\ref{fig:synth-regimes} varies one factor. A node carries one noisy reading and no load lag, so its local input is insufficient by construction. The GNN applies a shared encoder, two residual rounds of mean aggregation over $\mathbf{D}^{-1}(\bA + \mathbf{I})$ and a shared readout at width $32$; the identity control replaces $\bA + \mathbf{I}$ by $\mathbf{I}$.

\paragraph{Findings.}
The GNN has the lowest or near-lowest error across the synthetic settings. From $12$ to $250$ nodes, it improves from about $0.38$ to $0.29$ while the all-node MLP degrades from $0.66$ to $1.13$, and at $5\%$ of the training set it stays near $0.40$ against $0.85$ for the per-node MLP and more than $1.7$ for the all-node MLP, so unrestricted access does not reproduce its gain. At $\rho = 0$, it draws level with the per-node MLP, the gap falling to about $0.02$, and at $\sigma_\text{obs} = 0$ the identity network matches it, the neighborhood having nothing to denoise. This generator makes neighbors useful by construction when local measurements are noisy. It motivates examining larger datasets with shorter histories, but does not establish that the \REGIONSFR load lags are sufficient or that node count alone determines the ranking.

\subsection{Parameter Sharing and Neighborhood Access on Two Larger Datasets}\label{sec:real-regimes}

\paragraph{Setup.}
This section runs the full diagnostic protocol on $319$ \ELECTRICITY clients and $500$ \WEAVEUK feeders. Their training windows are short by design: they test whether sharing parameters and reading neighbors compensate for a history too short to fit each node alone, a regime that non-stationary demand makes common, since structural breaks leave only recent observations representative (Section~\ref{sec:related}). All models share hidden width $64$, the same local inputs (plus the neighbor mean where stated), horizon, and split within a cell; the target is the next hour on \ELECTRICITY and the next $48$ half-hours, from origins sliding by one half-hour, on \WEAVEUK; the MLPs have $1$ hidden layer, the GNNs a linear encoder, $2$ residual graph-convolution blocks, and a linear readout, with SAGE or ChebConv ($K=3$) selected per cell on validation, and the empty-graph GNN runs the same backbone with $\bA = \mathbf{0}$, which separates the change of backbone from the edges. The graph keeps each node's $5$ strongest neighbors; writing $\mathbf{A}_\text{kNN}$ for this directed adjacency ($(\mathbf{A}_\text{kNN})_{ij}=1$ if $j$ is among those of $i$), the GNN receives $\bA = \mathbf{A}_\text{kNN}+\mathbf{A}_\text{kNN}^\top$, in which a mutual pair weighs twice as much as a one-way edge, and the edge increments are also reported on the simple graph $\max(\mathbf{A}_\text{kNN},\mathbf{A}_\text{kNN}^\top)$, with each edge stored once. The primary score is the per-node normalized RMSE (nRMSE), the root mean squared error of each node after standardizing the targets by their training standard deviation (by node and, on \WEAVEUK, by horizon), averaged over nodes; it weights every node equally, whereas system scores weight nodes by their load and reward errors that cancel in the sum; Table~\ref{table:real-regimes} reports both. Targets, windows, graphs, parameter counts, and the standardization behind nRMSE are detailed in Section~\ref{app:large-setup} of the Appendix.

\begin{table}[tp]
\centering
\caption{The node-wise models at hidden width $64$ on the full datasets, $3$ seeds, under the two system scores and the per-node error (nRMSE). The GNN uses ChebConv ($K=3$) on \ELECTRICITY at $60$ days and \WEAVEUK at $30$ hours, SAGE otherwise, as selected on validation. Equal width does not match depth or total parameter count. \WEAVEUK window lengths include the validation portion; \ELECTRICITY lengths are fitting windows, followed by $2$ validation days.}
\label{table:real-regimes}
\resizebox{\linewidth}{!}{%
\begin{tabular}{l|ccc|ccc}
\toprule
 & \multicolumn{3}{c|}{\ELECTRICITY, $319$ clients (RMSE in MW)} & \multicolumn{3}{c}{\WEAVEUK, $500$ feeders (RMSE in kWh)} \\
Model & RMSE & MAPE (\%) & nRMSE & RMSE & MAPE (\%) & nRMSE \\
\midrule
\multicolumn{7}{c}{\emph{Short window: $11$ days (\ELECTRICITY), $30$ hours (\WEAVEUK)}} \\
\midrule
Per-node MLP & \rkc{$22.3$\vr{0.7}} & \rkc{$2.34$\vr{0.11}} & $0.3030$\vr{0.0002} & $346.8$\vr{3.8} & $9.73$\vr{0.14} & $1.0509$\vr{0.0013} \\
Per-node MLP $+$ neighbors & $24.1$\vr{2.6} & $2.72$\vr{0.34} & $0.3201$\vr{0.0036} & \rka{$323.2$\vr{2.0}} & \rka{$8.89$\vr{0.08}} & $0.9995$\vr{0.0010} \\
Shared MLP & \rkb{$19.4$\vr{1.3}} & \rka{$2.10$\vr{0.15}} & \rkb{$0.2804$\vr{0.0014}} & \rkb{$329.7$\vr{11.0}} & \rkb{$9.02$\vr{0.31}} & \rkc{$0.9074$\vr{0.0103}} \\
All-node MLP & $26.0$\vr{2.4} & $2.89$\vr{0.23} & $0.4527$\vr{0.0048} & $421.8$\vr{82.3} & $12.45$\vr{2.75} & $1.0609$\vr{0.0583} \\
Empty-graph GNN & $23.5$\vr{2.9} & $2.78$\vr{0.35} & \rkc{$0.2827$\vr{0.0039}} & \rkc{$334.9$\vr{4.7}} & $9.26$\vr{0.17} & \rkb{$0.8894$\vr{0.0020}} \\
GNN & \rka{$17.9$\vr{2.5}} & \rkb{$2.23$\vr{0.41}} & \rka{$0.2775$\vr{0.0038}} & $348.6$\vr{8.0} & \rkc{$9.17$\vr{0.20}} & \rka{$0.8808$\vr{0.0047}} \\
\midrule
\multicolumn{7}{c}{\emph{Long window: $60$ days (\ELECTRICITY), $10$ days (\WEAVEUK)}} \\
\midrule
Per-node MLP & $21.3$\vr{1.6} & \rkb{$1.78$\vr{0.07}} & $0.2776$\vr{0.0009} & $359.0$\vr{1.6} & $9.60$\vr{0.02} & $0.8956$\vr{0.0004} \\
Per-node MLP $+$ neighbors & \rkc{$19.8$\vr{0.4}} & \rka{$1.76$\vr{0.09}} & \rkb{$0.2752$\vr{0.0020}} & $343.8$\vr{3.5} & $9.65$\vr{0.11} & $0.8762$\vr{0.0007} \\
Shared MLP & \rkb{$19.2$\vr{1.2}} & \rkc{$1.85$\vr{0.26}} & \rkc{$0.2770$\vr{0.0011}} & \rkc{$293.6$\vr{12.5}} & \rkc{$7.84$\vr{0.15}} & \rkc{$0.8013$\vr{0.0086}} \\
All-node MLP & $28.7$\vr{2.3} & $3.41$\vr{0.42} & $0.4336$\vr{0.0103} & $359.6$\vr{27.8} & $10.77$\vr{1.01} & $0.8787$\vr{0.0648} \\
Empty-graph GNN & $21.4$\vr{1.2} & $2.36$\vr{0.33} & $0.2797$\vr{0.0035} & \rkb{$271.0$\vr{10.0}} & \rkb{$7.63$\vr{0.15}} & \rkb{$0.7941$\vr{0.0055}} \\
GNN & \rka{$17.6$\vr{1.7}} & $2.03$\vr{0.36} & \rka{$0.2752$\vr{0.0060}} & \rka{$250.7$\vr{16.2}} & \rka{$6.61$\vr{0.58}} & \rka{$0.7745$\vr{0.0056}} \\
\bottomrule
\end{tabular}
}
\end{table}

\begin{figure}[tp]
\centering
\resizebox{\linewidth}{!}{%
\begin{tikzpicture}[
  font=\small, >=Stealth,
  mod/.style={draw=gray!60, fill=gray!4, line width=0.6pt, rounded corners=2pt,
              minimum width=4.5cm, minimum height=1.5cm, align=center},
]
\node[mod] (pn)   at (2.6,3.7)  {\footnotesize $M_\text{pn}$\ \ \scriptsize non-shared, own features\\[2pt]
  \scriptsize\textcolor{nodeblue}{$0.2776$}\quad\textcolor{nodegreen}{$1.0509$}\\[1pt]
  \tiny\textcolor{gray!70!black}{per-node error}};
\node[mod] (sh)   at (11.8,3.7) {\footnotesize $M_\text{sh}$\ \ \scriptsize shared, own features\\[2pt]
  \scriptsize\textcolor{nodeblue}{$0.2770$}\quad\textcolor{nodegreen}{$0.9074$}\\[1pt]
  \tiny\textcolor{gray!70!black}{per-node error}};
\node[mod] (pnn)  at (2.6,0.0)  {\footnotesize $M_\text{pn+n}$\ \ \scriptsize non-shared, reads neighbors\\[2pt]
  \scriptsize\textcolor{nodeblue}{$0.2752$}\quad\textcolor{nodegreen}{$0.9995$}\\[1pt]
  \tiny\textcolor{gray!70!black}{per-node error}};
\node[mod] (gr)   at (11.8,0.0) {\footnotesize $M_\text{gr}$\ \ \scriptsize shared, reads neighbors\\[2pt]
  \scriptsize\textcolor{nodeblue}{$0.2752$}\quad\textcolor{nodegreen}{$0.8808$}\\[1pt]
  \tiny\textcolor{gray!70!black}{per-node error}};

\draw[->, nodedark, line width=0.8pt] (pn) -- (sh);
\node[fill=white, inner sep=2pt, align=center, font=\scriptsize] at (7.2,4.35)
  {\textcolor{gray!70!black}{weight sharing}\\ error $\downarrow$ \textcolor{nodeblue}{$0.0006$}\ \ \textcolor{nodegreen}{$0.1435$}};
\draw[->, nodedark, line width=0.8pt] (pnn) -- (gr);
\node[fill=white, inner sep=2pt, align=center, font=\scriptsize] at (7.2,-1.05)
  {\textcolor{gray!70!black}{shared GNN, neighbors already read}\\ error $\downarrow$ \textcolor{nodeblue}{$0.0000$}\ \ \textcolor{nodegreen}{$0.1187$}};
\draw[->, nodedark, line width=0.8pt] (pn) -- (pnn);
\node[anchor=east, fill=white, inner sep=2pt, align=center, font=\scriptsize] at (2.20,1.85)
  {\textcolor{gray!70!black}{neighborhood,}\\ \textcolor{gray!70!black}{weights non-shared}\\ error $\downarrow$ \textcolor{nodeblue}{$0.0024$}\ \ \textcolor{nodegreen}{$0.0514$}};
\draw[->, nodedark, line width=0.8pt] (sh) -- (gr);
\node[anchor=west, fill=white, inner sep=2pt, align=center, font=\scriptsize] at (12.20,1.85)
  {\textcolor{gray!70!black}{neighborhood,}\\ \textcolor{gray!70!black}{weights already shared; backbone changes}\\ error $\downarrow$ \textcolor{nodeblue}{$0.0018$}\ \ \textcolor{nodegreen}{$0.0266$}};
\draw[->, darkred, dashed, line width=0.8pt] (pn.south east) -- (gr.north west);
\node[fill=white, inner sep=2pt, align=center, font=\scriptsize] at (7.2,1.85)
  {\textcolor{gray!70!black}{both axes together}\\ error $\downarrow$ \textcolor{nodeblue}{$0.0024$}\ \ \textcolor{nodegreen}{$0.1702$}};

\node[anchor=east, font=\scriptsize, nodeblue]  at (2.55,5.15) {\dsetstyle{Electricity}, $60$ d};
\node[anchor=west, font=\scriptsize, nodegreen] at (2.90,5.15) {\dsetstyle{WEAVE-UK}, $30$ h};
\end{tikzpicture}}
\caption{The two decompositions of the per-node gap on the $60$-day \ELECTRICITY window (blue) and the $30$-hour \WEAVEUK window (green) of Table~\ref{table:real-regimes}. Vertices give per-node errors; each arrow points to the better model and gives the error decrease. Both routes reach the same total, with increments differing by $0.0248$ on \WEAVEUK; arrows into the GNN also change the backbone, so this difference is not a pure factorial interaction.}
\label{fig:commuting-square}
\end{figure}
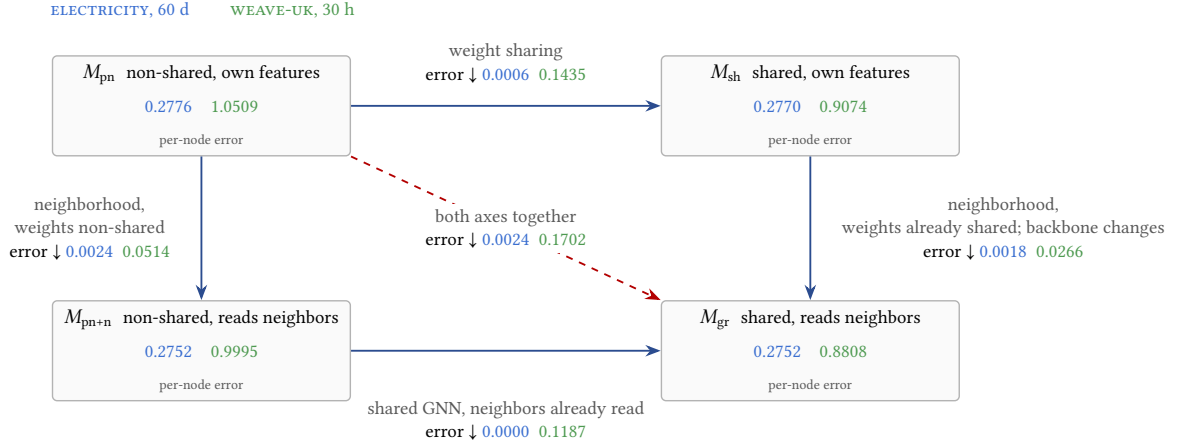

The models occupy $4$ combinations of parameter sharing and neighborhood access, giving two paths from the per-node MLP to the GNN. We call an \emph{increment} the decrease in per-node error, $E(\text{before}) - E(\text{after})$, between successive models. Write $M_\text{pn}$ for the per-node MLP, $M_\text{sh}$ for the shared MLP, $M_\text{gr}$ for the GNN, and $M_\text{pn+n}$ for the per-node model that also reads the mean features of its neighbors, and abbreviate $E_\cdot := E(M_\cdot)$.
Then,
\begin{align}%
E_\text{pn} - E_\text{gr}
&= (E_\text{pn} - E_\text{sh}) + (E_\text{sh} - E_\text{gr}) \nonumber \\
&= (E_\text{pn} - E_\text{pn+n}) + (E_\text{pn+n} - E_\text{gr}).
\end{align}%
The first line adds the sharing increment to the \emph{conditional edge increment} $E_\text{sh} - E_\text{gr}$; the second adds the \emph{non-shared edge increment} $E_\text{pn} - E_\text{pn+n}$ to the remaining gap to the GNN. Moving from the shared MLP to the GNN also changes the backbone. The empty-graph GNN, $M_\emptyset$, splits that step into a backbone increment and an edge increment at constant backbone, $E_\text{sh} - E_\text{gr} = (E_\text{sh} - E_\emptyset) + (E_\emptyset - E_\text{gr})$. The second path has no such control: the non-shared model receives a one-hop feature mean, whereas the GNN aggregates learned representations over two layers (Figure~\ref{fig:commuting-square}).

\paragraph{Findings.}
The fixed-configuration \REGIONSFR ranking reverses on both datasets: the per-node MLP finishes behind the GNN on the per-node error in every cell, by a margin that the bootstrap below does not separate from zero on the $60$-day \ELECTRICITY window, and on \WEAVEUK it also finishes behind the shared MLP, which reads no neighbor. The system scores sometimes disagree; at $30$ hours on \WEAVEUK both go to the per-node model that reads its neighbors, whose residuals cancel more in the sum of $500$ feeders.

On the first path, the sharing increment is $+0.0006$ on the $60$-day \ELECTRICITY window and $+0.1435$ on the $30$-hour \WEAVEUK window. The remaining shared-MLP-to-GNN gap ranges from $+0.0018$ to $+0.0268$ across the $4$ conditions. At constant backbone, the edges are worth $+0.0086$ $[+0.0052, +0.0120]$ and $+0.0196$ $[+0.0166, +0.0228]$ on \WEAVEUK at $30$ hours and $10$ days, and $+0.0051$ $[+0.0014, +0.0083]$ and $+0.0046$ $[+0.0025, +0.0065]$ on \ELECTRICITY at $11$ and $60$ days. These increments use the training graph, whose $k$-nearest-neighbor construction stores a mutual pair twice; with each edge stored once they become $+0.0069$, $+0.0191$, $+0.0033$, and $+0.0002$, so the $60$-day \ELECTRICITY gain vanishes (Table~\ref{table:rewire}). Once the seeds of each model are resampled as well as the nodes, both \ELECTRICITY intervals include zero, whereas both \WEAVEUK intervals stay positive (Appendix). The change of backbone accounts for the rest ($+0.0180$ and $+0.0072$ on \WEAVEUK, $-0.0023$ and $-0.0028$ on \ELECTRICITY), where the empty-graph GNN does slightly worse than the shared MLP. On the second path, the neighbor mean improves a per-node model by $+0.0194$ on \WEAVEUK at $10$ days and $+0.0514$ at $30$ hours, a sixth and three tenths of total gaps of $+0.1211$ and $+0.1702$. On \ELECTRICITY, it is worth $+0.0024$ at $60$ days and $-0.0171$ at $11$, where sharing carries $+0.0226$ of a total $+0.0255$. Reading neighbors doubles the input width of a node-specific model, so at $11$ days the price of those parameters and the neighbors' contribution are confounded. Against the true graph stored as a simple graph, a degree-matched rewiring that pays the same price over arbitrary neighbors (Section~\ref{app:controls} of the Appendix) reproduces $98\%$ and $71\%$ of the non-shared gain on \WEAVEUK and $83\%$ of the $11$-day degradation on \ELECTRICITY, whereas on the shared path it removes $52\%$ and $93\%$ of the edge gain at constant backbone on \WEAVEUK ($0.0036$ of $0.0069$, $0.0178$ of $0.0191$, one rewired graph per seed), so the identity of the neighbors matters there; over $10$ independent rewirings at $30$ hours, rewiring removes the whole gain on average (Appendix).

We assess sensitivity to the composition of the observed set of nodes by resampling them. The $6$ models of a cell share the split and $3$ seeds; errors are averaged over seeds within each node, and nodes are drawn with replacement $5000$ times, with every model scored on the same nodes in each draw. On \WEAVEUK, the sharing, non-shared-neighborhood, shared-neighborhood, and total increments have $95\%$ percentile intervals that exclude zero at both windows. At $30$ hours, these are $+0.1435$ $[+0.1364, +0.1510]$, $+0.0514$ $[+0.0468, +0.0564]$, $+0.0266$ $[+0.0208, +0.0328]$, and $+0.1702$ $[+0.1612, +0.1793]$, respectively. On \ELECTRICITY, at $11$ days the sharing increment is $+0.0226$ $[+0.0124, +0.0335]$ and the non-shared-neighborhood increment is $-0.0171$ $[-0.0219, -0.0127]$; at $60$ days the total gap, $+0.0024$ $[-0.0036, +0.0079]$, includes zero. These intervals measure stability to that composition, conditional on the fitted models and test period; they are not interpreted as sampling intervals for a population of independent nodes. They are interpreted separately for each dataset and window.

On \WEAVEUK, most of the improvement over the per-node MLP is already obtained by the shared MLP, without neighbor inputs. Adding the neighbor mean to the non-shared model recovers $16$--$30\%$ of the per-node-to-GNN gap; on \ELECTRICITY, this addition brings no consistent gain, whereas at constant backbone the edges improve the shared GNN on \WEAVEUK, by $0.0086$ and $0.0196$. On \ELECTRICITY, their gain does not differ from zero once seed variability is included, and it vanishes at $60$ days (each edge is stored once). The Appendix reports the GRU, learned-adjacency, width, and node-count controls.

\subsection{Complementarity of Graph and Non-Graph Forecasters}\label{sec:aggreg-summary}

An online expert aggregation asks whether graph and non-graph forecasters complement each other\ifblindrev\else\footnote{Earlier mixtures of GCN and SAGE variants with the additive model appear in \citet{campagne2026leveraging}.}\fi. On \REGIONSFR, ML-Poly combines the national forecasts of $8$ experts: $6$ GNNs, a weight-shared feedforward network, and the additive model; each family is withdrawn in turn (Figure~\ref{fig:aggreg-grid} in the Appendix). Forecasting each half-hour from the covariates observed at that half-hour (single-step setting), the mixture reaches $970$~MW, $279$~MW below its best expert. Withdrawing the GNNs costs $132$~MW and withdrawing the two non-graph experts $117$~MW, so each family improves the mixture; removing whole architectures, however, does not isolate the contribution of their edges. On a synthetic control whose nodes depend only on their own covariates, the GNN experts' share of the mixture's error falls from $14\%$ to $4\%$. Weights are not proportional to contribution: on a synthetic dataset forecast a day at a time, whose structure the experts cannot exploit, the GNN experts hold $0.86$ of the weight, above their count share of $0.75$, yet their withdrawal costs $34$~MW, $4\%$ of the error. The analysis uses its own split, older configurations, and $1$ seed per expert, and two of its GNN experts were selected on its test year, which biases their withdrawal cost upward; its figures are descriptive (Section~\ref{app:aggdiag}).

\section{Conclusion}\label{sec:conclusion}

We proposed a diagnostic protocol that separates the two mechanisms a GNN combines, parameter sharing and neighborhood access, by setting the GNN against a per-node model, a shared model without neighbors, and the empty-graph GNN, and by rewiring the graph at matched degrees. We ran it in full on the \WEAVEUK feeders and the \ELECTRICITY clients, after a tuned benchmark on \REGIONSFR and the feeders.
\Qflat{Q1}{Does a GNN forecast load better than strong non-graph models?} Not uniformly. On \REGIONSFR, no GNN beats a pooled gradient-boosted model, although a bottom-up mixture of the $8$ GNNs, at $809$~MW, finishes ahead of every individual model of Table~\ref{table:combined-results}; against a per-node MLP the ranking changes with tuning, the tuned per-node MLP at $922$~MW falling within the $891$--$986$~MW range of the tuned GNNs, whereas at a fixed configuration it has the lowest mean error at every training volume. On \WEAVEUK, with $9$ training days per feeder, the zero-shot t0-gamma is the best baseline and has the lowest MAPE of the benchmark; only GATv2 and APPNP have a lower mean RMSE, both by less than one seed std, and a ChebConv of order one, which reads no neighbor, is level with Chronos-2. The ranking of convolutions does not carry from one dataset to the other.
\Qflat{Q2}{Where a GNN wins, does the gain come from the graph?} On \WEAVEUK, mostly not. Sharing alone recovers $78$--$84\%$ of the gap from the per-node MLP to the GNN, a reference that overfits its short record and does worse than persistence at $30$ hours. Against a per-node ridge, the strongest node-specific reference wherever the gap is nonzero, the shared MLP still recovers $63\%$ and $71\%$ of the gap on \WEAVEUK, although this step also moves from a linear to a nonlinear model; with the model class held fixed, a shared ridge recovers $84\%$ and $41\%$ (Appendix). Adding the neighbor mean to a node-specific model recovers $16$--$30\%$, and arbitrary neighbors (maintaining the node degrees) reproduce $98\%$ and $71\%$ of that gain. Of the remaining gap, the deeper GNN backbone accounts for $+0.0180$ and $+0.0072$. On \ELECTRICITY, the share carried by sharing ranges from $-50\%$ to $111\%$ over five $11$-day windows, while the neighbor mean degrades the per-node model in all five.
\Qflat{Q3}{At equal architecture, do neighbors help?} A little, with a small, consistent gain on \WEAVEUK in the larger-dataset comparison. The empty-graph GNN keeps the backbone and the parameter count of the GNN, so this comparison holds capacity fixed. On \WEAVEUK, the edges improve the per-node error by $0.0086$ and $0.0196$, about $6$--$21\%$ of the sharing increment, and degree-matched rewiring removes half or more of the gain measured with each edge stored once; a shared MLP that reads the neighbor mean is not distinguishable from the GNN there. On \ELECTRICITY, the edge gain stays at about $+0.005$ or less and is not consistently different from zero once seed variability is included, and it vanishes at $60$ days with each edge stored once.
\Qflat{Q4}{When do neighbors help?} The synthetic study identifies noisy local measurements as one mechanism, without establishing a general boundary between regimes. On real data, their contribution depends on the dataset and the window: the neighbor mean moves the fixed \REGIONSFR per-node MLP from $876$ to about $871$~MW, recovers $16$--$30\%$ of the gap on \WEAVEUK and degrades the per-node model on the $11$-day \ELECTRICITY window, while on the feeders the edge gain at constant backbone stays positive over both windows. A forecasting gain should be credited to the graph only if it survives the comparison with a per-node model, a shared model without neighbors, and the same backbone without edges.

\paragraph{Limitations and perspectives.}
In the controlled comparisons, each cell corresponds to the mean of $3$ seeds, so differences of the order of the reported dispersion are not claimed as rankings. On the non-shared path, the neighbor mean is a fixed one-hop summary, not the learned aggregation of the GNN, so that path is not matched in architecture. The larger-dataset decomposition is run at one fixed width; its signs hold at widths $32$, $64$, and $128$ on \WEAVEUK and in the $11$-day \ELECTRICITY cell, whereas the $60$-day cell, whose total includes zero, changes sign across widths (Appendix); configurations tuned model by model are not tested there, and on \REGIONSFR tuning changes the ranking. On \WEAVEUK, the node bootstrap does not cover seasonal or weather variability. The short-window comparison also retains a graph estimated from the longer training record; re-estimating it inside the $30$-hour window gives a more accurate GNN and doubles the edge gain at constant backbone (Appendix). The decomposition is measured on one week of one distribution operator and on five $11$-day windows of the \ELECTRICITY record ending between March and July $2019$ and its last weeks ($60$ days); its shares may differ in other seasons and for longer histories, and on \ELECTRICITY the sharing increment already falls from $+0.0226$ at $11$ days to $+0.0006$ at $60$ days. Weather inputs are observed values, whereas day-ahead operation uses forecasts, which introduce noise. The synthetic study suggests that such noise could increase the value of neighbors, which these experiments do not test. Time-varying graphs
and probabilistic forecasts are possible extensions.

\paragraph{AI use statement.}%
In this work, we used generative AI tools to revise the manuscript, to help with translation and data visualization, to debug code and write the scripts of some additional controls and analyses, and to search for information and verify bibliographic references. We did not use them for the research content itself: the research questions, the choice of data, the synthetic data, the theoretical models and conceptual frameworks, the mathematical claims, or the interpretation of results. The authors wrote the original text before its AI-assisted revision and most of the code without AI assistance, have reviewed all AI-assisted work, and take responsibility for the final content, including any text, claims or artifacts produced with the aid of generative AI.

\bibliography{refs}

\clearpage
\appendix
\counterwithin{figure}{section}
\counterwithin{table}{section}
\counterwithin{equation}{section}
\renewcommand{\theHsection}{app.\Alph{section}}

\section{Message Passing Formalism}\label{app:mp}

\begin{figure}[tp]
    \centering
    \scalebox{.8}{\begin{tikzpicture}[node distance=2cm]
        \begin{scope}[xshift=-1.5cm]
            % Gray layer
            \draw[fill=gray!10, rounded corners] (-0.7,-5.7) rectangle (6,1);

            % Nodes at stage n
            \node[draw, circle, fill=gray!20] (Node_n) {$V_\ell$};
            \node[draw, circle, fill=gray!20, below of=Node_n] (Edge_n) {$E_\ell$};
            \node[draw, circle, fill=gray!20, below of=Edge_n] (Global_n) {$U_\ell$};
              
            % Nodes at stage n+1
            \node[draw, circle, fill=gray!50, right=4cm of Node_n] (Node_np1) {$V_{\ell+1}$};
            \node[draw, circle, fill=gray!50, below of=Node_np1] (Edge_np1) {$E_{\ell+1}$};
            \node[draw, circle, fill=gray!50, below of=Edge_np1] (Global_np1) {$U_{\ell+1}$};
              
            % Arrows between stages
            \draw[->, dashed, darkblue] (Node_n) -- node[midway, above] {$\phi^v$} (Node_np1);
            \draw[->, dashed, darkblue] (Edge_n) -- node[midway, above] {$\phi^e$} (Edge_np1);
            \draw[->, dashed, darkblue] (Global_n) -- node[midway, above] {$\phi^u$} (Global_np1);

            % Messages
            \draw[- Circle, dotted, teal!70!black, thick] (0.5, 0) to [bend right=50] node[midway, left] {$\rho_{V_\ell\to E_\ell}$} (2, -2.);
            \draw[-, dotted] (0.5, -4) to [bend left=50](2, -2.);
            \draw[-, dotted] (0.5, -4) to [bend right=50] (4, 0.);
            \draw[- Circle, dotted, orange!70!black, thick] (2, -2) to [bend left=50] node[midway, right] {$\rho_{E_\ell\to V_\ell}$} (4, 0.);
            \draw[- Circle, dotted, green!60!black, thick] (2, -2) to [bend right=50] node[midway, right] {$\rho_{E_\ell\to U_\ell}$} (4, -4);
            \draw[-, dotted] (0.5, 0) to [bend left=50] (4, -4.);

            \node[below] at (2.5, -5) {\textit{Message passing layer}};
        \end{scope}
    
        % Left graph
        \begin{scope}[xshift=-6cm, yshift=-3cm]
            \draw[dashed, rounded corners] (-0.7,-1) rectangle (2.7,3);
            
            \node[circle, draw, fill=blue!25, minimum size=.7cm] (Left_A) at (0,0) {};
            \node[circle, draw, fill=teal!25, minimum size=.7cm] (Left_B) at (2,0) {};
            \node[circle, draw, fill=purple!25, minimum size=.7cm] (Left_C) at (2,2) {};
            \node[circle, draw, fill=orange!25, minimum size=.7cm] (Left_D) at (0,2) {};
              
            \draw[gray!70] (Left_A) -- (Left_B);
            \draw[gray!70] (Left_B) -- (Left_C);
            \draw[gray!70] (Left_C) -- (Left_D);
            \draw[gray!70] (Left_D) -- (Left_A);
            
            \draw[-] (2.7,-1) -- (3.8,-2.6);
            \draw[-] (2.7,3) -- (3.8,3.93);
        \end{scope}
        
        % Right graph
        \begin{scope}[xshift=6cm, yshift=-3cm]
            \draw[dashed, rounded corners] (-0.7,-1) rectangle (2.7,3);
            
            \node[circle, line width=1pt, draw, fill=blue!50, minimum size=.7cm] (Right_A) at (0,0) {};
            \node[circle, line width=1pt, draw, fill=teal!50, minimum size=.7cm] (Right_B) at (2,0) {};
            \node[circle, line width=1pt, draw, fill=purple!50, minimum size=.7cm] (Right_C) at (2,2) {};
            \node[circle, line width=1pt, draw, fill=orange!50, minimum size=.7cm] (Right_D) at (0,2) {};
              
            \draw[gray!80, line width=1.5pt] (Right_A) -- (Right_B);
            \draw[gray!80, line width=1.5pt] (Right_B) -- (Right_C);
            \draw[gray!80, line width=1.5pt] (Right_C) -- (Right_D);
            \draw[gray!80, line width=1.5pt] (Right_D) -- (Right_A);

            \draw[-] (-1.5,-2.6) -- (-0.7,-1);
            \draw[-] (-1.5,3.93) -- (-0.7,3);
        \end{scope}
    \end{tikzpicture}}
    \caption{Example of a message-passing layer in a GNN. $V_\ell$, $E_\ell$, and $U_\ell$ respectively refer to node, edge, and global level at layer $\ell$. $\phi$ and $\rho$ are respectively update and propagation functions.}
    \label{fig:message_passing}
\end{figure}
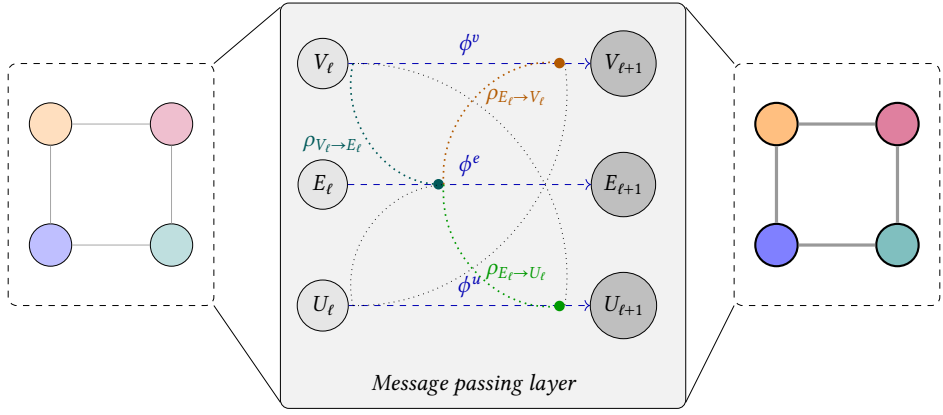

Message passing generalizes convolution to irregular domains \citep{gilmer2017neural}. In the framework of \citet{battaglia2018relational}, a message-passing layer operates at three interconnected levels, node $V_{\ell}$, edge $E_{\ell}$, and global $U_{\ell}$, jointly updated through propagation and update functions, as Figure~\ref{fig:message_passing} shows. The propagation operators $\rho$ define how information is aggregated across the graph, for instance from nodes to edges $\rho_{V_\ell \to E_\ell}$, from edges to nodes $\rho_{E_\ell \to V_\ell}$, or from both to the global representation $\rho_{V_\ell \to U_\ell}$, $\rho_{E_\ell \to U_\ell}$. The update functions $\Phi = (\phi^v, \phi^e, \phi^u)$ then transform these aggregated messages into new latent representations for nodes, edges, and the global graph state, respectively $V_{\ell+1} = \phi^v(V_\ell, \rho_{E_\ell \to V_\ell}, U_\ell)$, $E_{\ell+1} = \phi^e(E_\ell, \rho_{V_\ell \to E_\ell}, U_\ell)$ and
$U_{\ell+1} = \phi^u(U_\ell, \rho_{V_\ell \to U_\ell}, \rho_{E_\ell \to U_\ell})$.
A classical example is GCN, where the propagation operator aggregates neighboring node features through the normalized adjacency matrix $\tilde{\bA}$, and the update function is a linear transformation followed by a nonlinearity:
\begin{align*}
    \bh_v^{(\ell+1)}
    &= \sigma\!\left(\sum_{u \in \mathcal N_v \cup \{v\}} \tilde{A}_{uv}\,\bh_u^{(\ell)} \bW^{(\ell)}\right) \\
    &= \sigma\!\left(\rho_{E_{\ell} \rightarrow V_{\ell}}\left(\bh^{(\ell)}\right)\,\bW^{(\ell)}\right),
\end{align*}
where $\rho_{E_{\ell} \rightarrow V_{\ell}}$ corresponds to the normalized neighborhood aggregation, and $\phi^v(\cdot) = \sigma(\cdot\,\bW^{(\ell)})$ defines the update function. In this work, we use node-level interactions only: edge and global states are not used, and each GNN layer updates node representations by aggregating from neighboring nodes.

\section{Expert Aggregation Analysis}\label{app:aggdiag}

\paragraph{Synthetic datasets.}
To evaluate GNN forecasting under controlled spatial and temporal dependencies, we generate synthetic datasets in which the presence and strength of inter-node structure are known. These data do not provide a performance bound for the real datasets. They test whether a model uses the dependency inserted by the generator.

We generate node-level temperature and load series that mimic real-world variability. Let $T_j^\text{gen}(t)$ and $L_j^\text{gen}(t)$ denote the generated temperature and load at node $j$ and time $t$, with $T_j^\text{gen}(t) = a t + b_j (\cos \omega_1 t + \cos \omega_2 t)$, where $a$ is a linear trend coefficient, $\omega_1$ and $\omega_2$ encode daily and yearly cycles, and $\mathbf b = (b_j)_{1 \le j \le N} \sim \mathcal{N}(\hat{\boldsymbol{\mu}}, \hat{\mathbf{C}})$ with empirical mean $\hat{\boldsymbol{\mu}}$ and covariance $\hat{\mathbf{C}}$ estimated from observed temperatures. Temperatures are rescaled to match observed regional minima and maxima. Loads are then derived via a cubic spline mapping $L_j^\text{gen}(t) = \tilde f_j(T_j^\text{gen})(t) + \varepsilon_j(t)$, with $\tilde f_j$ fitted to the observed data and $\boldsymbol{\varepsilon}(t) \sim \mathcal{N}(\mathbf 0, \boldsymbol{\Sigma})$ drawn independently over time. Two covariance structures are considered. The first, $\boldsymbol{\Sigma} = \mathbf{D}_\varepsilon$, is diagonal and leaves the regions independent. The second couples them through the spatial graph by a conditional autoregression: with $\tilde{\bW}_\lambda = \mathbf{D}^{-1/2} \bW_\lambda \mathbf{D}^{-1/2}$ the symmetrically normalized adjacency and $\mathbf{D}$ the degree matrix,
\begin{align}
\boldsymbol{\Sigma} = \mathbf{D}_\varepsilon^{1/2} \mathbf{R}_\rho \mathbf{D}_\varepsilon^{1/2},
\qquad \mathbf{R}_\rho \propto (\mathbf{I} - \rho\, \tilde{\bW}_\lambda)^{-1},
\end{align}
where $\mathbf{R}_\rho$ is rescaled to unit diagonal so the marginal noise level of each region stays at $\mathbf{D}_\varepsilon$ and only the correlations change. The scalar $\rho \in [0,1)$ is the coupling coefficient: it sets how much of a node's residual its neighbors account for, and $\rho = 0$ is the independent case.

\paragraph{Single-step regression setup.}
This employs a \textit{direct regression} approach, commonly adopted in electricity load forecasting when the emphasis is placed on exogenous explanatory factors. Electricity demand at each time step $t$ is forecast based solely on exogenous covariates observed at the same time.
The model learns a mapping $\Phi_{\boldsymbol{\theta}} : (\mathcal{G}, \tilde{\mathbf{X}}_t) \mapsto \hat{\mathbf{y}}_t$, with $\tilde{\mathbf{X}}_t \in \mathbb{R}^{N\times d}$ representing normalized features for all $N$ nodes and $d$ covariates. The features include calendar indicators, weather-related variables, and regional characteristics, allowing the model to capture spatial dependencies encoded in the graph structure $\mathcal{G}$. It is an analysis tool, not an operational forecaster. The model parameters $\boldsymbol{\theta}$ are optimized by minimizing the MSE between predicted and observed loads.

\begin{figure}[tp]
    \centering
    \includegraphics[width=\linewidth]{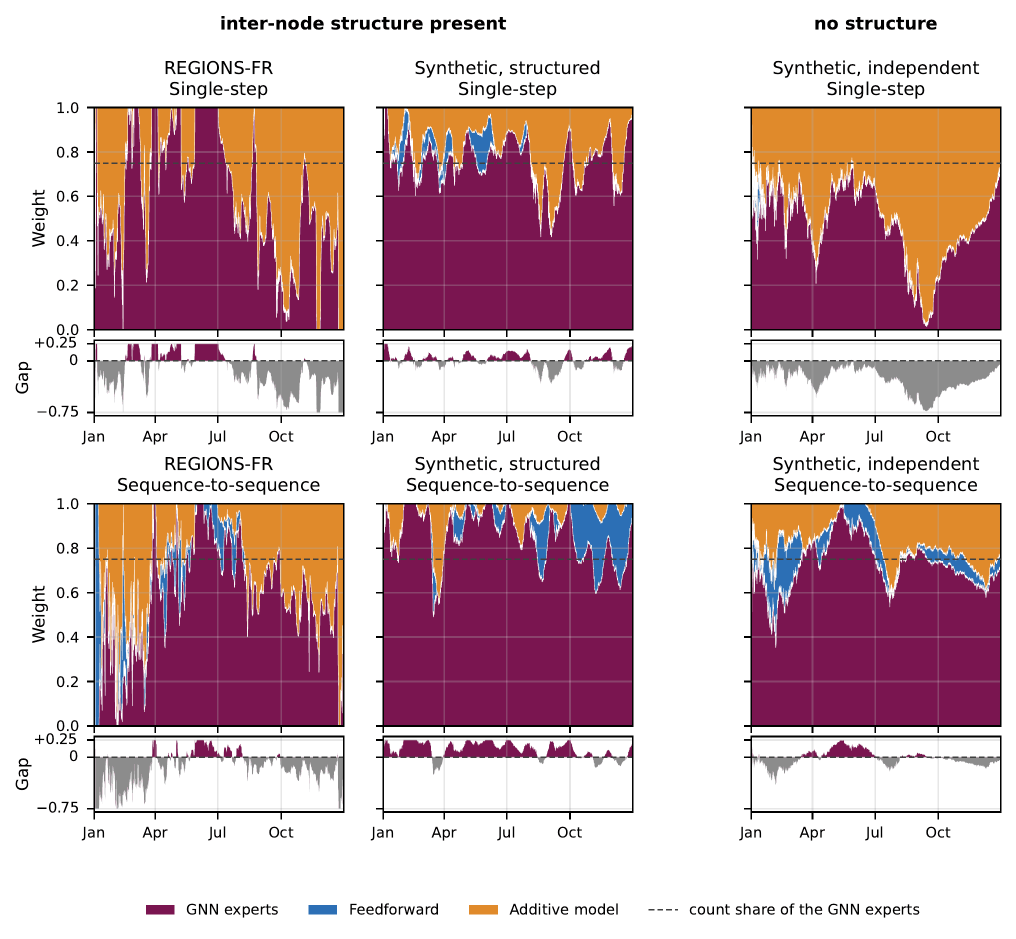}
    \caption{Weights ML-Poly plays over the test year $2019$. The two left columns are \REGIONSFR and the synthetic dataset whose residual covariance follows the spatial graph (unobservable by the experts); the third has none, its residuals being independent across nodes. Single-step forecasting is above and sequence-to-sequence forecasting below. The $6$ GNN experts are summed into one plum band, the feedforward network appearing in blue and the additive model in orange. The dashed line marks $0.75$, the share the $6$ hold among the $8$ by count alone, so a band sitting on it reports an allocation that cannot separate its experts and not an accuracy the edges bought. The strip beneath each plot carries the departure from that share, plum where the mixture buys the GNN experts more than counting them would give and gray where it buys them less.}
    \label{fig:aggreg-grid}
\end{figure}

Expert aggregation measures the complementarity of graph and non-graph experts, on condition that the measurement separates what the edges carry from the diversity any two model families produce on any data. The same protocol runs on two synthetic datasets. The generator above produces a temperature field per region and a load obtained by passing it through the region's own spline. In the first dataset, the residual noise is independent across regions, so a node's load depends on its own covariates alone. In the second, the residuals follow the conditional autoregression above carried by the spatial graph, at coupling $\rho = 0.9$, so residuals are correlated across neighbors but independent over time, and no expert observes its neighbors' current residuals. In neither synthetic dataset does a neighborhood carry information an expert can use; the two differ only in whether residual errors cancel in the national sum. Crossing the $3$ datasets with the $2$ forecasting settings gives the $6$ cells of Figure~\ref{fig:aggreg-grid} and Table~\ref{table:aggreg-cells}.

We observe that $8$ experts forecast the $12$ nodes in every cell: $6$ GNNs, a weight-shared feedforward network reading each node's own covariates, and the operationally inspired additive model fitted per node and per half-hour. This analysis predates the benchmark of Table~\ref{table:combined-results} and uses its own split and configurations. GCN, SAGE, GAT, and ChebConv use the configurations of an earlier tuning of this dataset: GCN on the spatial graph with one layer of $170$ channels, SAGE on the correlation graph with two layers of $401$ channels, GAT on GL3SR with three layers of $364$ channels and two heads, and ChebConv on DistSplines with one layer of $249$ channels and $K=3$. GatedGraph and LEConv have no tuned entry there; they were taken from a separate screening of $48$ operators that ranked them first and second by MAPE on the $2019$ test year, with a feature set that also contains the $48$ temperatures of the target slots. Their selection therefore used the year on which this analysis is scored, so their errors, weights, and withdrawal costs in this section are optimistic relative to the $6$ other experts and are read descriptively. Both are fixed at two layers of $256$ channels on the dynamic time warping graph, a width chosen on the validation months. All $8$ read the meteorological variables of the additive model, the position in the year, the day type, and the period. The sequence-to-sequence cells add the $48$ past loads of each node and commit a whole day at a time, whereas the single-step cells carry no lagged load and update the weights at every half-hour. Training covers January $2017$ to June $2018$ and validation the $6$ months that follow, $1$ seed per architecture.

\begin{table}[tp]
\centering
\caption{National root mean squared error in MW over the test year, and what each family of experts is worth inside the mixture. The $4$ cells above the rule have spatially correlated loads, the two below independent ones given their own covariates; `seq-to-seq' abbreviates the sequence-to-sequence setting. The last two columns give the error ML-Poly loses when the $6$ GNNs are withdrawn from the pool, and when the feedforward and the additive model are, in MW and as a share of the mixture's own error. The datasets sit at different load levels, so the shares compare across rows where the MW values do not.}
\label{table:aggreg-cells}
\footnotesize
\begin{tabular}{llccccc}
\toprule
Dataset & Setting & Best expert & Uniform & ML-Poly & \makecell{Without the\\GNN experts} & \makecell{Without the two\\others} \\
\midrule
\multirow{2}{*}{\dsetstyle{REGIONS-FR}}
 & single-step & $1249$ & $1144$ & $970$ & $+132$ ($14\%$) & $+117$ ($12\%$) \\
 & seq-to-seq  & $1047$ & $893$  & $844$ & $+162$ ($19\%$) & $+90$ ($11\%$) \\
\multirow{2}{*}{\makecell[l]{Synthetic\\structured}}
 & single-step & $917$ & $838$ & $799$ & $+79$ ($10\%$) & $+43$ ($5\%$) \\
 & seq-to-seq  & $800$ & $784$ & $766$ & $+34$ ($4\%$)  & $+1$ ($0\%$) \\
\midrule
\multirow{2}{*}{\makecell[l]{Synthetic\\independent}}
 & single-step & $658$ & $656$ & $624$ & $+23$ ($4\%$) & $+24$ ($4\%$) \\
 & seq-to-seq  & $604$ & $576$ & $574$ & $+19$ ($3\%$) & $+4$ ($1\%$) \\
\bottomrule
\end{tabular}
\end{table}

Take \REGIONSFR in the single-step setting first. The additive model and the ChebConv network are level at the head of the individual experts, $1249$ and $1254$~MW, GCN follows at $1273$~MW, GAT at $1298$~MW, SAGE at $1308$~MW, GatedGraph at $1314$~MW, LEConv at $1363$~MW, and the weight-shared feedforward network trails at $1428$~MW. ML-Poly reaches $970$~MW, $279$~MW below the best of its experts and $174$~MW below the uniform average, so the aggregation gains more than anything separating the experts from one another. Withdrawing the $6$ GNNs costs $132$~MW, the additive model and the feedforward network alone stopping at $1102$~MW; withdrawing the two non-graph experts costs $117$~MW, the $6$ GNNs alone stopping at $1088$~MW. Each family is worth about two thirds of the $179$~MW that separate the best and the worst individual expert. The weights place that contribution in the year: the GNN experts take the whole weight in June and $0.88$ of it in April, hold above half from February to August, and fall to $0.21$ in October and $0.32$ in December. Each GNN keeps the adjacency selected for it during hyperparameter search, so the convolution and the graph construction move together here and this run ranks neither on its own.

The independent synthetic control answers the question the first cell leaves open. Where a node's load is a function of its own temperature, the GNN experts are worth $4\%$ and $3\%$ of the mixture's error, against $14\%$ and $19\%$ on \REGIONSFR, so their share divides by $3.7$ in the single-step setting and by $5.7$ in the other. Adaptive weighting also has less to arbitrate: ML-Poly improves on the uniform average by $32$~MW in the synthetic single-step cell and by $2$~MW in the sequence-to-sequence one, where the same gaps on \REGIONSFR are $174$ and $49$~MW.

Adding the spatial covariance raises the GNN experts' contribution in the single-step setting from $23$ to $79$~MW, or from $4\%$ to $10\%$ of the mixture's error, and their mean weight rises from $0.48$ to $0.76$. Since no expert observes its neighbors' residuals, this rise reflects how residuals that no longer cancel in the national sum reweight the experts' errors, not information read from the neighbors. The sequence-to-sequence cell behaves differently: the GNN experts hold $0.86$ of the weight, yet their withdrawal costs only $34$~MW ($4\%$), close to the $19$~MW ($3\%$) of the independent control. The structured dataset thus separates a high weight assigned to accurate GNN architectures from their smaller marginal contribution.

Between the two settings on real data, the balance tips. Adding the $48$ past loads of each node raises the GNN share from $14\%$ to $19\%$ and lowers the other family's from $12\%$ to $11\%$, so the day-ahead task, which gives every node its own recent history, is where the GNN experts contribute most. The additive model loses its lead there, dropping to $1075$~MW behind SAGE at $1047$~MW: its per-half-hour form has no way to use a lag block that the networks read jointly.

A reader who stops at the bands will conclude the opposite of what the withdrawals indicate, since the GNN experts hold $0.73$ of the weight on the dataset without structure, more than the $0.61$ they hold on \REGIONSFR, while being worth $19$~MW there against $132$~MW. Weight mass is not contribution, and two mechanisms separate them.

The first is counting. GNNs make up $6$ of the $8$ experts, so an allocation unable to tell them apart returns their count share ($0.75$), which the dashed line of Figure~\ref{fig:aggreg-grid} marks in every cell. That is the situation of the unstructured sequence-to-sequence cell: its $8$ errors lie within $47$~MW of one another and its mean weights run from $0.083$ to $0.187$ around the uniform $0.125$, so the band settles on the line and reports indifference. Where the errors do separate, the allocation leaves the line, falling to $0.48$ in the unstructured single-step cell, whose experts spread over $100$~MW.

The second is that a GNN on a structureless dataset is not necessarily disadvantaged. Nothing forces it to use its neighborhood: the aggregation can drive that weight to zero, and the network then behaves as a per-node model reading its own covariates, which is exactly what this dataset favors. It forecasts well, and its weight follows from that accuracy; on that dataset the $4$ best individual forecasters are GNNs. It cannot, however, supply information absent from its neighbors. Weights describe the fitted mixture, while the withdrawal columns of Table~\ref{table:aggreg-cells} measure the contribution of the GNN family to that mixture. Removing entire experts does not isolate the contribution of their edges. The individual errors of that cell bear the second mechanism out: SAGE reaches $604$~MW, GAT and GatedGraph $607$~MW, ChebConv $608$~MW, ahead of the feedforward network at $633$~MW and the additive model at $651$~MW.

\begin{figure}[tp]
    \centering
    \includegraphics[width=\linewidth]{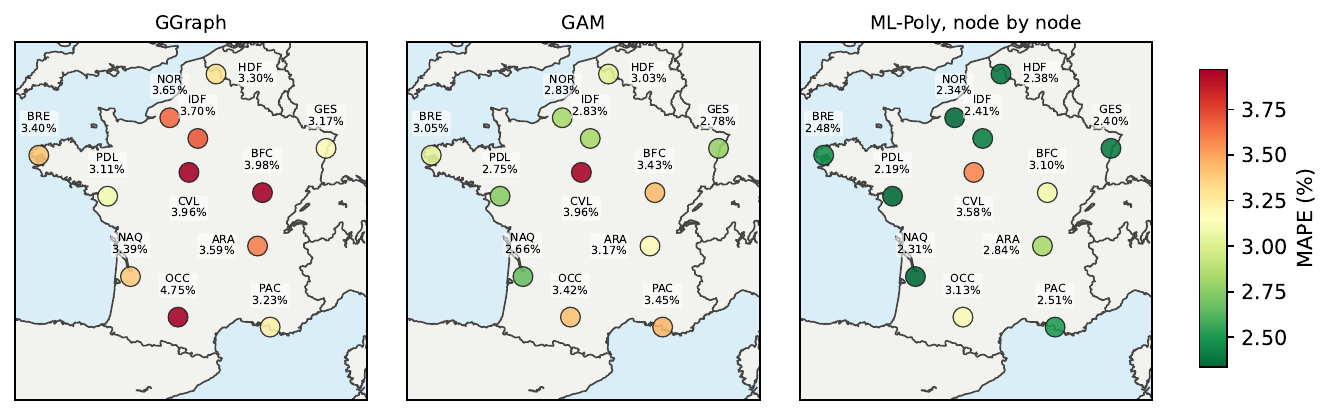}
    \caption{Mean absolute percentage error of each region's own forecast over the test year, single-step setting. Left, GatedGraph, the most accurate of the $6$ GNN experts region by region; center, the additive model; right, ML-Poly run region by region, each region carrying its own mixture and its own weights. The color scale is shared. Regions carry their official codes: ARA Auvergne-Rhône-Alpes, BFC Bourgogne-Franche-Comté, BRE Bretagne, CVL Centre-Val de Loire, GES Grand Est, HDF Hauts-de-France, IDF Île-de-France, NAQ Nouvelle-Aquitaine, NOR Normandie, OCC Occitanie, PAC Provence-Alpes-Côte d'Azur, PDL Pays de la Loire.}
    \label{fig:nodal-errors}
\end{figure}

Everything above scores the national sum, where a region's overshoot cancels another's shortfall, and no expert optimizes that sum. The neural experts minimize a squared error taken node by node, each node standardized by its own training moments, so the $12$ regions of \REGIONSFR weigh equally in the objective whatever their load; the additive model is fitted region by region on its own scale. Keeping the regions apart at evaluation reads the forecasts on the criterion they were trained for, and it changes the ranking, as Figure~\ref{fig:nodal-errors} shows.

The two levels do not order the GNN experts alike. Nationally ChebConv leads them at $1254$~MW, $5$~MW behind the additive model, and GatedGraph trails it by $60$~MW. Region by region the ranking reverses: GatedGraph reaches $3.60\%$ against $4.17\%$ for ChebConv, and the additive model keeps the lead at $3.11\%$, the $4$ remaining networks lying between $4.36\%$ and $4.47\%$, and the feedforward network at $4.55\%$. The networks' regional errors cancel in the sum to an extent the additive model's do not, so the national metric credits a network with an accuracy its regions do not show. Provence-Alpes-Côte d'Azur is the worst-served region for $4$ of the $6$ GNNs, $5.75$--$5.86\%$, Bourgogne-Franche-Comté for ChebConv, Occitanie for GatedGraph, and Centre-Val de Loire for the additive model at $3.96\%$.

Running ML-Poly inside each region, the bottom-up scheme of Table~\ref{table:aggregation-results} applied to this pool, gives the best regional accuracy, $2.64\%$ on average and $3.58\%$ on its worst region, below every expert in every region. Summing its $12$ forecasts yields $971$~MW nationally against $970$~MW for the mixture built on the national series directly, so weighting region by region costs nothing on the sum and saves half a point of relative error inside it.

The disagreement between the two levels belongs to the single-step setting. In the sequence-to-sequence cell, the two rankings agree: the additive model leads region by region at $2.53\%$, SAGE follows at $2.62\%$ as it does nationally, and the $6$ remaining experts lie between $2.77\%$ and $2.91\%$. The node-level mixture reaches $2.15\%$ there and $844$~MW nationally, again level with the national-level mixture. Centre-Val de Loire is the worst-served region for all $9$ forecasters in that cell, the region whose load is smallest and whose relative error the national sum weights least.

\section{Forecasting Protocol, Feature Sets, and Training Volume}\label{app:arms}

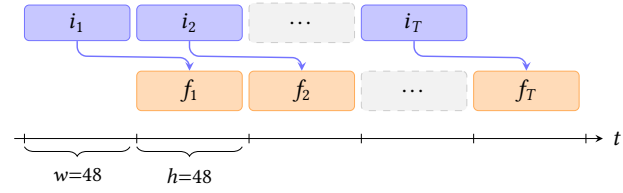
\begin{figure}[tp]
\centering
\scalebox{0.9}{
\begin{tikzpicture}[>=stealth, font=\small,
    inp/.style  ={fill=blue!22,   draw=blue!55,   rounded corners=2pt,
                  minimum width=1.55cm, minimum height=0.48cm, align=center},
    fcast/.style={fill=orange!28, draw=orange!65, rounded corners=2pt,
                  minimum width=1.55cm, minimum height=0.48cm, align=center},
    ghost/.style={fill=gray!10,   draw=gray!40,   rounded corners=2pt, dashed,
                  minimum width=0.85cm, minimum height=0.48cm, align=center},
]

\def\rx{0.}  % right-panel x origin

% ── input blocks ─────────────────────────────────────────────────────
\node[inp]   (i1) at (\rx+0.775, 2.2) {$i_1$};
\node[inp]   (i2) at (\rx+2.425, 2.2) {$i_2$};
\node[ghost, minimum width=1.55cm, minimum height=0.58cm]   (i3) at (\rx+4.075, 2.2) {\dots};
\node[inp]   (i4) at (\rx+5.725, 2.2) {$i_{T}$};

% ── forecast blocks (offset 1.65cm right = one block + gap) ──────────
\node[fcast] (f1) at (\rx+2.425, 1.2) {$f_1$};
\node[fcast] (f2) at (\rx+4.075, 1.2) {$f_2$};
\node[ghost, minimum width=1.55cm, minimum height=0.58cm] (f3) at (\rx+5.725, 1.2) {\dots};
\node[fcast] (f4) at (\rx+7.375, 1.2) {$f_{T}$};

% ── prediction arrows: i_k → f_k  (down → right → down into north)
% [yshift=0.10cm]f_k.north sits at y=1.74 = i_k.south−0.22, so middle segment is horizontal.
\draw[->, blue!55, semithick, rounded corners=4pt]
    (i1.south) -- ++(0,-0.22) -- ([yshift=0.19cm]f1.north) -- (f1.north);
\draw[->, blue!55, semithick, rounded corners=4pt]
    (i2.south) -- ++(0,-0.22) -- ([yshift=0.19cm]f2.north) -- (f2.north);
\draw[->, blue!55, semithick, rounded corners=4pt]
    (i4.south) -- ++(0,-0.22) -- ([yshift=0.19cm]f4.north) -- (f4.north);

% ── time axis ────────────────────────────────────────────────────────
\draw[->] (\rx-0.15, 0.5) -- (\rx+8.5, 0.5) node[right] {$t$};
\foreach \x in {0.00, 1.65, 3.30, 4.95, 6.60, 8.25} {
    \draw (\rx+\x, 0.43) -- (\rx+\x, 0.57);
}

% ── braces ───────────────────────────────────────────────────────────
\draw[decorate, decoration={brace, amplitude=4pt, mirror}]
    (\rx+0.00, 0.38) -- (\rx+1.55, 0.38)
    node[midway, below=5pt, font=\footnotesize] {$w{=}48$};
\draw[decorate, decoration={brace, amplitude=4pt, mirror}]
    (\rx+1.65, 0.38) -- (\rx+3.20, 0.38)
    node[midway, below=5pt, font=\footnotesize] {$h{=}48$};

\end{tikzpicture}
}
\caption{Sequence-to-sequence evaluation protocol ($48$ half-hourly steps, stride $h = 48$). Input windows $i_k$ (blue) each feed a forecast window $f_k$ (orange), and consecutive forecast windows are contiguous, so each prediction spans $w = 48$ lags into $h = 48$ future steps. Training, validation and testing all use this stride, so every forecast origin falls on a day boundary and output $k$ always denotes slot $k$ of the day. A stride of one would place the origin at an arbitrary half-hour and force a single $48$-output head to represent $48$ phase-shifted mappings, which is a different learning problem and a different test set; every model of the benchmark, graph or not, is trained and scored under the stride of this figure; Section~\ref{sec:real-regimes} states its own protocol.}
\label{fig:seq2seq}
\end{figure}

\begin{figure}[tp]
    \centering
    \scalebox{0.9}{
    \begin{tikzpicture}[font=\small, node distance=0.5cm and 0.5cm]
        % French Dataset timeline
        \node at (4.5, 2.3) {\textbf{\dsetstyle{REGIONS-FR}}};
        \draw[->, thick] (-2,1.5) -- (12,1.5);

        % French periods
        \fill[blue!20] (-2,1.3) rectangle (5.56,1.7);
        \node at (1.78,1.5) {Train};

        \fill[orange!20] (5.56,1.3) rectangle (8.08,1.7);
        \node at (6.82,1.5) {Val};

        \fill[green!20] (8.08,1.3) rectangle (10.6,1.7);
        \node at (9.34,1.5) {Test};

        \draw (-2,1.3) -- (-2,1.7);
        \draw (5.56,1.3) -- (5.56,1.7);
        \draw (8.08,1.3) -- (8.08,1.7);
        \draw (10.6,1.3) -- (10.6,1.7);

        \node at (-2,1.1) {\footnotesize Jan '15};
        \node at (5.56,1.1) {\footnotesize Jan '18};
        \node at (8.08,1.1) {\footnotesize Jan '19};
        \node at (10.6,1.1) {\footnotesize Jan '20};

        % UK Dataset timeline
        \node at (4.5, .3) {\textbf{\dsetstyle{WEAVE-UK}}};
        \draw[->, thick] (-2, -0.5) -- (12, -0.5);

        % UK periods
        \fill[blue!20] (-2,-0.7) rectangle (7,-0.3);
        \node at (2.5,-0.5) {Train};

        \fill[orange!20] (7.,-.7) rectangle (7.9,-0.3);
        \node at (7.45,-0.5) {Val};

        \fill[green!20] (7.9,-.7) rectangle (10.6,-0.3);
        \node at (9.15,-0.5) {Test};

        \draw (-2,-.7) -- (-2,-0.3);
        \draw (7.,-.7) -- (7.,-0.3);
        \draw (7.9,-.7) -- (7.9,-0.3);
        \draw (10.6,-.7) -- (10.6,-0.3);

        \node at (-2,-0.9) {\footnotesize Feb 14};
        \node at (6.9,-0.9) {\footnotesize Feb 23};
        \node at (8.1,-0.9) {\footnotesize Feb 25};
        \node at (10.6,-0.9) {\footnotesize Feb 28 (end)};
    \end{tikzpicture}}

    \caption{Temporal splits for training, validation, and testing on \REGIONSFR and \WEAVEUK.}
    \label{fig:dataset-splits}
\end{figure}

The role of the $48$ load lags is measured by removing them. The comparison covers $4$ feature sets on \REGIONSFR (Table~\ref{table:feature-arms}), each scored by the lowest validation error reached over $18$ GNN configurations ($6$ convolutions at three widths), on $1$ seed as a screening pass.

\begin{table}[tp]
\centering
\caption{Validation RMSE in MW on \REGIONSFR, by feature set. The middle column is the best of $18$ GNN configurations ($6$ convolutions at three widths), and the right column is the gap between the best and the worst of those. It uses $1$ seed, so the column reads as a screening pass and not as a ranking.}
\label{table:feature-arms}
\footnotesize
\begin{tabular}{lcc}
\toprule
Features & Best & Best to worst \\
\midrule
Calendar and weather & $1141$ & $586$ \\
$+$ temperatures of the target slots & $1133$ & $367$ \\
$+$ 48 load lags & $\mathbf{865}$ & $120$ \\
Both blocks & $\mathbf{768}$ & $119$ \\
\bottomrule
\end{tabular}
\end{table}

\begin{figure}[tp]
    \centering
    \includegraphics[width=0.9\linewidth]{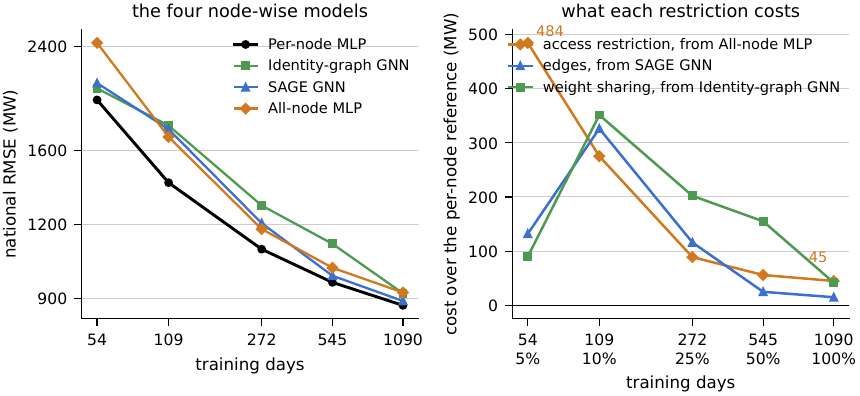}
    \caption{Table~\ref{table:data-regime} read against the training window, on a logarithmic day axis. \emph{Left:} the national error of the $4$ node-wise models, the per-node MLP leading at every volume. \emph{Right:} the same table as increments over that reference model, so a curve is the price of one restriction and a higher value is a costlier one. The access restriction falls from $484$ to $45$~MW as the window opens and the edge cost from $132$ to $15$~MW; weight sharing costs $42$--$351$~MW with no monotone trend.}
    \label{fig:gnn-data-regime}
\end{figure}

Adding load lags to calendar and weather reduces the best validation RMSE by $276$~MW; adding target-slot temperatures alone reduces it by $8$~MW. The spread across $18$ configurations falls from $586$ to $120$~MW with load lags, and to $119$~MW with both additions. The homogeneity of the GNN architectures observed in Table~\ref{table:combined-results} is consistent with a dependence on the feature set: the lags narrow the spread between configurations by a factor of $5$.
Table~\ref{table:data-regime} gives the values shown in the left plot of Figure~\ref{fig:french-regime}, and Figure~\ref{fig:gnn-data-regime} reads them as increments over the per-node MLP.

\begin{table}[tp]
\centering
\caption{National RMSE (MW) on \REGIONSFR as the training set contracts, SAGE convolution, days drawn at random, $3$ seeds.}
\label{table:data-regime}
\footnotesize
\begin{tabular*}{\linewidth}{@{\extracolsep{\fill}}lcccc}
\toprule
Training days & Per-node MLP & All-node MLP & Identity-graph GNN & SAGE GNN \\
\midrule
$54$\ \ ($5\%$)   & \rka{$1948$\vr{69}} & $2432$\vr{269} & \rkb{$2038$\vr{103}} & \rkc{$2080$\vr{112}} \\
$109$ ($10\%$)    & \rka{$1412$\vr{28}} & \rkb{$1687$\vr{105}} & $1763$\vr{101} & \rkc{$1738$\vr{55}} \\
$272$ ($25\%$)    & \rka{$1090$\vr{41}} & \rkb{$1179$\vr{25}} & $1292$\vr{17} & \rkc{$1206$\vr{41}} \\
$545$ ($50\%$)    & \rka{$958$\vr{5}}   & \rkc{$1014$\vr{29}} & $1113$\vr{31} & \rkb{$983$\vr{27}} \\
$1090$ ($100\%$)  & \rka{$876$\vr{3}}   & $921$\vr{3}    & \rkc{$918$\vr{7}} & \rkb{$891$\vr{19}} \\
\bottomrule
\end{tabular*}
\end{table}

\section{Setup of the Larger-Dataset Comparison}\label{app:large-setup}

This section details the setup of the comparison of Section~\ref{sec:real-regimes} on $319$ \ELECTRICITY clients and $500$ \WEAVEUK feeders. All models use hidden width $64$ and the same local inputs (plus the neighbor mean where stated), horizon, and split within a dataset and window. The MLPs have $1$ hidden layer, whereas the GNNs use a linear encoder, $2$ residual graph-convolution blocks, and a linear readout. The convolution, SAGE or ChebConv with $K=3$, is selected per cell on validation: ChebConv on the $60$-day \ELECTRICITY and $30$-hour \WEAVEUK windows, SAGE on the other two. Thus width is common, but depth, parameter count, and aggregation differ. The per-node MLP holds $265{,}727$ parameters on \ELECTRICITY and $3{,}416{,}000$ on \WEAVEUK, and $490{,}303$ and $5{,}240{,}000$ once it reads the neighbor mean; the shared MLP holds $833$ and $6{,}832$. The GNN holds $17{,}729$ with SAGE or $25{,}921$ with ChebConv on \ELECTRICITY, and $23{,}728$ or $31{,}920$ on \WEAVEUK, as does the empty-graph GNN. Finally, the empty-graph GNN ($\bA = \mathbf{0}$) runs the same backbone and convolution, so it propagates nothing; it separates the change of backbone from the edges.

On \ELECTRICITY, the target is the next hour, read from lags $1$, $2$, $3$, $24$, $25$, $48$, and $168$ and a cyclical encoding of hour and weekday, over fitting windows of $11$ and $60$ days followed by $2$ validation days and $4$ test days, which end the record (July $2019$); one-step hourly targets do not overlap across these boundaries. On \WEAVEUK, the target comprises the next $48$ half-hours, read from $48$ load lags, the cyclical instant, and the weekday. Unlike the benchmark, origins slide by one half-hour. The two pre-test windows contain $481$ and $60$ origins, approximately $10$ days and $30$ hours; the last fifth is reserved for validation, leaving $385$ and $48$ fitting origins. The test contains $145$ origins spanning $3$ days, with overlapping $48$-step targets. Training and test origins are built separately from $14$--$24$ and $25$--$28$ February $2024$, respectively, so no fitting or validation target falls in the test period; within the training record, the targets of the last $47$ fitting origins overlap those of the first validation origins, which affects only validation-based choices (epoch, convolution, and ridge penalties); test targets remain separate.

\ELECTRICITY graphs retain each node's $5$ strongest training correlations and both edge directions. On \WEAVEUK, the graph is taken from a cache built from the full training series, restricted to the retained nodes and reduced to at most $5$ neighbors per node before symmetrization. This cached graph is used for both window lengths, so the $30$-hour condition restricts model-fitting data but not graph-estimation data. The per-node $+$ neighbors model receives the mean features of the same in-neighbors. \ELECTRICITY is recorded hourly and summed over clients, corresponding to an average aggregate load of about $680$~MW over the test window; \WEAVEUK is summed over feeders.

The system scores let errors of opposite sign cancel across clients or feeders. We therefore also report a per-node normalized RMSE (nRMSE). Targets are standardized using training means and stds, separately for each node and, on \WEAVEUK, each forecast horizon. For each node, we take the root mean squared standardized error over test origins and horizons, then average over nodes. An nRMSE of $1$ denotes an error on the scale of the training stds; forecasting the training mean need not achieve $1$ on a test period with different moments. We take nRMSE as the primary score of the decomposition as it weights every node equally, whereas system scores weight nodes by their load and reward errors that cancel in the sum. Table~\ref{table:real-regimes} reports both, and on \WEAVEUK at $30$ hours they disagree: by system RMSE the per-node MLP with the neighbor mean leads and the GNN trails the shared MLP.

\section{Node-Count Sweep}\label{app:nodecount}

\WEAVEUK can be thinned to examine how sharing and neighborhood access vary with node count. We draw subsets of $10$, $40$, $160$, and $500$ feeders and build the same $5$-nearest graph inside each draw (Table~\ref{table:node-count}). SAGE remains the reference convolution; ChebConv is included as a control for the choice of convolution. On the $10$-day window, sharing improves nRMSE at every size, by $0.061$ at $10$ feeders and $0.094$ at $500$. Conditional on sharing, SAGE improves on the shared MLP by $0.004$ at $10$ nodes, $0.001$ at $40$, $0.024$ at $160$, and $0.027$ at $500$. The edge gain is therefore small in the two smaller sampled subsets and larger in the two larger ones.

On the $30$-hour window, sharing improves nRMSE by $0.094$ at $10$ feeders and about $0.14$ from $40$ onwards. The conditional SAGE gain does not order with node count on that window at all: $0.040$ at $10$ feeders, $0.003$ at $40$, $0.037$ at $160$, and $0.031$ at $500$. The $10$-day window is therefore the only one of the two in which the edge gain grows with node count, and even there the sweep does not locate a threshold, several of its differences being comparable with the reported dispersion. The two windows normalize by different training moments, so increments compare within a window and only their sign compares across windows.

The all-node MLP, which $12$ regions could only place at a single size, degrades at $500$ feeders on the $30$-hour window. Over $10$ days, it stays ahead of the per-node reference model at every size, by $0.017$--$0.062$, without ever reaching the shared MLP. Over $30$ hours, its advantage holds near $0.13$ up to $160$ feeders, then turns into a loss of $0.010$ at $500$, where its spread across draws ($0.058$) exceeds that of every other model. Using every series at once stops reducing the error once the dataset is large and each record short, while reading the retained neighbors still improves performance.

\begin{table}[tp]
\centering
\caption{\WEAVEUK as the dataset is thinned, three draws of nodes and $1$ seed each, the same $5$-nearest graph rebuilt inside every draw. The per-node error is computed after standardizing targets separately by node and horizon using training moments, then taking the root mean squared error over test origins and horizons and averaging over nodes. A value of $1$ denotes an error on the training-standard-deviation scale, not a test-set mean-forecast baseline. This score and MAPE compare across node counts, whereas a raw RMSE grows with the number of feeders summed and is given at $500$ of them in Table~\ref{table:real-regimes}, whose fixed width of $64$ this table shares. The stds over the draws in parentheses; at $500$ feeders the draw is the whole dataset, so the three repetitions of that row vary the fit alone. Gray shades mark the top three models of each row, darkest being the best. The GNN improves on the per-node reference model at every sampled size. Its gain over the shared MLP grows with node count on the $10$-day window and does not order with node count on the $30$-hour one; the limited repetitions do not identify a threshold on either.}
\label{table:node-count}
\footnotesize
\begin{tabular}{ccccc|cc}
\toprule
\multirow{2}{*}{Window} & \multirow{2}{*}{Nodes} & \multicolumn{3}{c|}{MLP} & \multicolumn{2}{c}{GNN} \\
 & & Per-node & Shared & All-node & SAGE & ChebConv \\
\midrule
\multicolumn{7}{c}{\emph{Per-node error, nRMSE}} \\
\midrule
$30$ h & $10$ & $1.0790$\vr{0.0194} & $0.9846$\vr{0.0578} & \rka{$0.9333$\vr{0.0135}} & \rkb{$0.9444$\vr{0.0170}} & \rkc{$0.9632$\vr{0.0049}} \\
$30$ h & $40$ & $1.0216$\vr{0.0161} & \rkc{$0.8746$\vr{0.0298}} & $0.8952$\vr{0.0100} & \rkb{$0.8719$\vr{0.0405}} & \rka{$0.8637$\vr{0.0304}} \\
$30$ h & $160$ & $1.0733$\vr{0.0190} & \rkc{$0.9315$\vr{0.0302}} & $0.9402$\vr{0.0244} & \rka{$0.8949$\vr{0.0256}} & \rkb{$0.8994$\vr{0.0164}} \\
$30$ h & $500$ & $1.0509$\vr{0.0013} & \rkc{$0.9074$\vr{0.0103}} & $1.0609$\vr{0.0583} & \rka{$0.8760$\vr{0.0015}} & \rkb{$0.8808$\vr{0.0047}} \\
\addlinespace[2pt]
$10$ d & $10$ & $0.9005$\vr{0.0285} & \rkb{$0.8400$\vr{0.0479}} & $0.8481$\vr{0.0281} & \rka{$0.8362$\vr{0.0483}} & \rkc{$0.8450$\vr{0.0520}} \\
$10$ d & $40$ & $0.8877$\vr{0.0422} & \rkb{$0.8189$\vr{0.0605}} & $0.8412$\vr{0.0516} & \rka{$0.8178$\vr{0.0652}} & \rkc{$0.8216$\vr{0.0645}} \\
$10$ d & $160$ & $0.9152$\vr{0.0130} & \rkc{$0.8295$\vr{0.0116}} & $0.8535$\vr{0.0192} & \rka{$0.8053$\vr{0.0064}} & \rkb{$0.8111$\vr{0.0066}} \\
$10$ d & $500$ & $0.8956$\vr{0.0004} & \rkc{$0.8013$\vr{0.0086}} & $0.8787$\vr{0.0648} & \rka{$0.7745$\vr{0.0056}} & \rkb{$0.7862$\vr{0.0103}} \\
\midrule
\multicolumn{7}{c}{\emph{MAPE (\%)}} \\
\midrule
$30$ h & $10$ & $12.05$\vr{0.93} & $12.96$\vr{2.40} & \rka{$11.21$\vr{0.21}} & \rkc{$11.94$\vr{0.26}} & \rkb{$11.88$\vr{0.45}} \\
$30$ h & $40$ & $12.12$\vr{1.11} & \rkc{$10.81$\vr{0.72}} & \rkb{$10.77$\vr{0.78}} & $11.27$\vr{1.18} & \rka{$10.70$\vr{0.90}} \\
$30$ h & $160$ & $10.09$\vr{0.57} & \rkc{$9.45$\vr{0.97}} & $9.55$\vr{0.14} & \rkb{$9.19$\vr{0.45}} & \rka{$9.04$\vr{0.25}} \\
$30$ h & $500$ & $9.73$\vr{0.14} & \rka{$9.02$\vr{0.31}} & $12.45$\vr{2.75} & \rkc{$9.26$\vr{0.21}} & \rkb{$9.17$\vr{0.20}} \\
\addlinespace[2pt]
$10$ d & $10$ & $11.40$\vr{0.26} & \rka{$10.07$\vr{0.76}} & $11.35$\vr{0.22} & \rkb{$10.09$\vr{0.61}} & \rkc{$10.27$\vr{0.30}} \\
$10$ d & $40$ & $9.52$\vr{0.96} & \rka{$8.81$\vr{1.03}} & $10.45$\vr{0.92} & \rkb{$9.01$\vr{1.58}} & \rkc{$9.26$\vr{1.37}} \\
$10$ d & $160$ & $9.88$\vr{0.44} & \rkc{$8.52$\vr{0.41}} & $10.29$\vr{0.91} & \rka{$7.90$\vr{0.37}} & \rkb{$7.97$\vr{0.20}} \\
$10$ d & $500$ & $9.60$\vr{0.02} & \rkc{$7.84$\vr{0.15}} & $10.77$\vr{1.01} & \rka{$6.61$\vr{0.58}} & \rkb{$7.52$\vr{0.60}} \\
\bottomrule
\end{tabular}
\end{table}

\section{Further Controls on the Decomposition}\label{app:controls}

The main comparison fixes hidden width at $64$, while depth and parameterization differ across model classes. The following check examines sensitivity to width; it does not remove those architectural differences. Repeating the $4$ cells at widths $32$ and $128$ leaves every \WEAVEUK increment with its sign at all three widths, and the totals within $0.013$ of one another, while the share the non-shared neighborhood recovers rises with capacity, from $27\%$ to $31\%$ at $30$ hours and from $12\%$ to $19\%$ at $10$ days. A larger model extracts slightly more from the same neighbors. On the \ELECTRICITY clients, the $11$-day cell is equally stable, its non-shared neighborhood costing between $-0.0156$ and $-0.0171$ across the three widths, while the $60$-day cell changes the sign of both its sharing arrow and its total from one width to the next.

The shared MLP reads the lag block as a flat vector, whereas a recurrent model would read it as a sequence, and no recurrent baseline appears elsewhere in this paper. A shared GRU over the lags, joined to the calendar features at its final state and emitting the targets of the cell, enters the same protocol at the same width. It reaches $0.3322$ and $0.2816$ on the \ELECTRICITY clients at $11$ and $60$ days and $0.9099$ and $0.8068$ on the feeders at $30$ hours and $10$ days, so it trails the shared MLP in all $4$ cells. Only on the $11$-day \ELECTRICITY window does that margin ($0.0518$) exceed its own dispersion across seeds ($0.0035$); in the other three the two models are level within that spread, and the GNN stays ahead of both everywhere. The tested GRU provides no clear improvement over the shared MLP under this protocol; this does not establish that recurrent modeling is generally unhelpful. The recurrence is also the least stable model in the comparison: its spread on the $60$-day \ELECTRICITY window ($0.0095$) is $4$ times the per-node-to-GNN gap of that cell.

The GNN of the decomposition receives a fixed graph. A model that learns its own adjacency may use the neighborhood differently. MTGNN \citep{wu2020connecting} enters the same $4$ cells under the same datasets, splits, seeds, and optimizer, fitted twice against itself: once with its graph learner active and once with the message passing switched off, so that the pair differs in the learned graph and in nothing else. Its capacity is cut from the published defaults until the parameter count sits below the per-node reference model of the same cell. The learned graph is worth $+0.0127$ on the feeders at $30$ hours, $+0.0072$ at $10$ days, $+0.0068$ on the \ELECTRICITY clients at $11$ days and $-0.0059$ at $60$. The $60$-day \ELECTRICITY estimate is the largest relative to its seed dispersion ($-0.0059$, paired std $0.0036$, $3$ seeds), and it is a degradation. The $4$ estimates average $+0.0052$ with a std of $0.0079$; these are descriptive summaries, since windows within a dataset are not independent replications. Against the models of Table~\ref{table:real-regimes} the learned-adjacency network reaches the GNN in no cell: $0.8857$ against $0.8808$ at $30$ hours, $0.7962$ against $0.7745$ at $10$ days, and on the \ELECTRICITY clients it trails the shared MLP as well. This MTGNN comparison provides no clear improvement over the fixed-GNNs under the tested protocol.

The all-node MLP provides a further comparison with unrestricted access to all series. It also changes the parameterization: a common hidden layer feeds the outputs for all nodes, whereas the per-node MLP has a separate hidden layer for each node. On \ELECTRICITY, the per-node model improves nRMSE by $0.1497$ at $11$ days and $0.1560$ at $60$ relative to the all-node MLP. On \WEAVEUK, it improves nRMSE by $0.0100$ at $30$ hours but loses $0.0169$ at $10$ days. Thus unrestricted access is not consistently beneficial, and these comparisons do not isolate information access from the way that information is represented.

\paragraph{Homophily and what remains to predict.}
In graph learning, homophily is the tendency of adjacent nodes to carry similar features or targets. Neighbor aggregation is naturally suited to such graphs, whereas low homophily can make conventional message passing ineffective \citep{zhu2020beyond}. For the continuous load target $\mathbf{x}_t$, similarity over the graph can be measured by Moran's $I$ \citep{moran1950notes}:
\begin{align}
I(\mathbf{x}_t) &= \frac{N}{S_0}\,
\frac{\sum_{i,j} W_{ij} (x_{i,t} - \overline{x}_t)(x_{j,t} - \overline{x}_t)}{\sum_i (x_{i,t} - \overline{x}_t)^2}.
\end{align}
Here, $W_{ij}$ is the adjacency weight and $S_0 = \sum_{i,j} W_{ij}$. We compute $I$ at every evaluation time and report its temporal mean. A positive $I$ value
suggests that adjacent nodes tend to lie on the same side of the cross-sectional mean; a value near the random-label baseline (approx. zero for these node counts) suggests that adjacency does not organize the values. 

The target is homophilous by this measure: mean Moran's $I$ reaches $+0.555$ on \ELECTRICITY and $+0.192$ on \WEAVEUK. The same conclusion appears without a graph statistic: the mean temporal correlation of pairs joined by an edge is $0.870$, against $0.529$ for arbitrary pairs, on \ELECTRICITY, and $0.386$, against $0.134$, on \WEAVEUK. Target homophily alone does not imply that neighbors improve a forecast, because two loads can share seasonal variation already visible in each node's own history. To test what remains after that history is used, we fit one ridge model per node on its own features and apply Moran's $I$ to its residuals. The values fall to $+0.079$ and $+0.063$. Adjacent loads are therefore similar, while their unexplained errors are aligned only weakly in magnitude (Moran's $I\le0.08$) though far from random under the residual-energy test ($z\ge10$, Table~\ref{table:rewire}): the neighbor mostly repeats information already supplied by the node's own past.

\paragraph{A graph of the same degrees, joining arbitrary pairs.}
The edge increments of Section~\ref{sec:real-regimes} measure the value of receiving the $5$ neighbors each node selects (about $10$ after symmetrization); they do not determine whether the identities of those series matter. We isolate edge identity by replacing the true graph $\mathcal{G}$ with a graph $\mathcal{G}'$ obtained by degree-preserving double edge swaps. Two edges $(a,b)$ and $(c,d)$ are replaced by $(a,d)$ and $(c,b)$ whenever the result remains simple, and this operation is repeated $10|\mathcal{E}|$ times. The true graph stores mutual nearest neighbors twice ($18\%$ of directed entries on \ELECTRICITY, $6\%$ on \WEAVEUK), whereas the rewired graph is simple, so the comparison is made against the true graph stored as a simple graph; the vertex set, edge count, degree sequence, model, features, targets, splits, initialization, and training budget then remain unchanged. Only the assignment of neighbors changes: the retained neighbors of a node are replaced by as many nodes, constrained by the same degree sequence.

Before comparing forecasts, we check that this rewiring has removed the residual alignment described above. Let $\mathbf{r}_t \in \mathbb{R}^N$ be the observed residual vector of the per-node ridge and $\bL_\mathcal{H}$ the normalized Laplacian of the graph $\mathcal{H}$ being checked, either the true graph $\mathcal{G}$ or its rewired counterpart $\mathcal{G}'$. Define its residual energy by
\begin{align}
s_\mathcal{H}(\mathbf{r}) = \frac{\sum_t \mathbf{r}_t^\top \bL_\mathcal{H} \mathbf{r}_t}{\sum_t \mathbf{r}_t^\top \mathbf{r}_t}.
\end{align}
This quantity is small when residuals connected by an edge of $\mathcal{H}$ are similar. To calibrate it, let $\Pi$ be the uniform distribution over permutations of the node labels and let $\mathbf{r}^\pi$ denote the same residual series reassigned to nodes by $\pi \sim \Pi$. The graph $\mathcal{H}$ remains fixed during these permutations. With $\mu_\mathcal{H} = \EE_{\pi \in \Pi}[s_\mathcal{H}(\mathbf{r}^\pi)]$ and $\sigma_\mathcal{H} = \operatorname{sd}_{\pi \in \Pi}(s_\mathcal{H}(\mathbf{r}^\pi))$, the standardized residual-alignment score is
\begin{align}
z_\mathcal{H} = \frac{\mu_\mathcal{H} - s_\mathcal{H}(\mathbf{r})}{\sigma_\mathcal{H}}.
\end{align}
A large positive $z_\mathcal{H}$ suggests that the observed residual assignment is smoother along the fixed graph $\mathcal{H}$ than random assignments; $z_\mathcal{H}$ near zero suggests that its edges do not align with the residuals beyond chance. We compute this score separately on $\mathcal{G}$ and on $\mathcal{G}'$. The permutations serve only to define the reference distribution and cannot change any topological quantity. The two-hop count instead compares the topologies of $\mathcal{G}$ and $\mathcal{G}'$ themselves, which is why it changes after rewiring even though it would remain constant under every residual permutation.

\begin{table}[tp]
\centering
\caption{Paired degree-preserving rewiring control, averaged over $3$ seeds, against the true graph stored as a simple graph. Forecast columns report rewired minus true performance, so positive values are degradations; the last forecast column gives the edge increment at constant backbone on that simple graph. Structural columns show how the residual-alignment score and two-hop neighborhood size change when only the edge identities are randomized.}
\label{table:rewire}
\footnotesize
\begin{tabular}{lcccc|cc}
\toprule
\multirow{2}{*}{Dataset and window} & \multicolumn{3}{c}{Forecast change: rewired $-$ true} & Edge gain & \multicolumn{2}{c}{Graph diagnostic: true $\rightarrow$ rewired} \\
 & RMSE & MAPE points & nRMSE & nRMSE & $z_\text{residual}$ & Two-hop nodes \\
\midrule
\ELECTRICITY, $11$ d & $+2.82$~MW & $+0.31$ & $+0.0046$ & $+0.0033$ & $23.3 \rightarrow 1.5$ & $39.5 \rightarrow 72.8$ \\
\ELECTRICITY, $60$ d & $-3.29$~MW & $-0.35$ & $-0.0082$ & $+0.0002$ & $21.0 \rightarrow 1.9$ & $43.7 \rightarrow 76.3$ \\
\WEAVEUK, $30$ h    & $-33.6$~kWh & $-0.40$ & $+0.0036$ & $+0.0069$ & $13.1 \rightarrow 1.6$ & $135.1 \rightarrow 144.0$ \\
\WEAVEUK, $10$ d    & $+24.3$~kWh & $+1.21$ & $+0.0178$ & $+0.0191$ & $10.3 \rightarrow 0.7$ & $135.1 \rightarrow 144.0$ \\
\bottomrule
\end{tabular}
\end{table}

The diagnostics confirm that the control changes more than the names of the endpoints. Rewiring drives $z$ from $10.3$--$23.3$ on the true graphs to $0.7$--$1.9$, close to the random-label reference. At the same time, the two-hop neighborhood grows on \ELECTRICITY (from about $40$ to $73$ nodes), where the random edges act as shortcuts, and barely on \WEAVEUK ($135$ to $144$ nodes). The rewired graph retains the degree of every node but largely removes the preference for joining residual series that move together.

Rewiring costs $+0.0044$ of per-node error on average. The $12$ paired runs comprise three fits in each of $4$ dataset-window conditions: $+0.0046$ and $-0.0082$ on \ELECTRICITY at $11$ and $60$ days, $+0.0036$ and $+0.0178$ on \WEAVEUK at $30$ hours and $10$ days. The $4$ cell means have std $0.0106$. Windows within a dataset use overlapping data, so these $4$ values are descriptive comparisons, not independent replications for an inferential test. The MAPE points the same way on average ($+0.19$ of a point), its $4$ cell values running from $-0.40$ to $+1.21$. The system RMSE cannot be pooled at all, the two datasets carrying different units. The sign turns on the metric as much as on the dataset: on \WEAVEUK at $30$ hours rewiring improves both system scores and degrades the per-node one. Next to the edge increment at constant backbone on the same simple graph, rewiring costs $52\%$ on \WEAVEUK at $30$ hours and $93\%$ at $10$ days, exceeds the whole increment on \ELECTRICITY at $11$ days, and at $60$ days the rewired graph does better than the true one. On the shared path, the identity of the neighbors therefore matters in some cells, and the control does not reduce the edge increment to aggregation alone.

The same substitution applies to the non-shared path, comparing true and rewired neighbors at the same parameter count (Section~\ref{sec:real-regimes}). Giving the per-node model the rewired neighbor mean reproduces $98\%$ and $71\%$ of the gain obtained with the true mean, on the simple graph, on \WEAVEUK: $+0.0505$ against $+0.0515$ at $30$ hours and $+0.0140$ against $+0.0197$ at $10$ days. On \ELECTRICITY, the corresponding increments are $-0.0023$ against $+0.0026$ at $60$ days and $-0.0147$ against $-0.0177$ at $11$ days. The advantage of the true graph over the rewired graph ranges from $-0.0030$ to $+0.0057$ across these $4$ conditions. On the $11$-day \ELECTRICITY window, both true and rewired neighbors worsen the forecast. This is consistent with a cost from the enlarged input or estimation problem, but does not identify parameter count as the sole cause. On \WEAVEUK, most of the improvement survives rewiring; the true neighbor identities account for only $2\%$ and $29\%$ of the original non-shared gains in this control.

\paragraph{Stronger node-specific references and seed-level intervals.}
The per-node MLP of Table~\ref{table:real-regimes} carries thousands of weights per node for a few dozen training origins. Table~\ref{table:local-refs} sets it beside references that cannot overfit in the same way, on the same cells, splits, inputs, and score: a ridge regression per node, its penalty chosen node by node on the validation window, a ridge shared by all nodes, and persistence at the same slot of the previous day. The per-node ridge beats the per-node MLP in $3$ of the $4$ cells, and on \WEAVEUK at $30$ hours the per-node MLP does worse than persistence. Against the stronger of the two node-specific references, the shared MLP still recovers $79\%$ of the gap to the GNN on \ELECTRICITY at $11$ days (July window; Table~\ref{table:periods} shows how this share varies across windows), and $63\%$ and $71\%$ on \WEAVEUK at $30$ hours and $10$ days; at $60$ days on \ELECTRICITY the total gap does not differ from zero. That step also changes the model class. Within the linear class, sharing recovers $84\%$ and $41\%$ of the gap on \WEAVEUK and degrades the per-node ridge on \ELECTRICITY, so the \ELECTRICITY share comes from the nonlinear model more than from sharing.

\begin{table}[tp]
\centering
\caption{Per-node error (nRMSE) of the decomposition's models beside stronger node-specific references, on the cells of Table~\ref{table:real-regimes}. Ridge regressions are fitted in closed form on the same standardized inputs; the per-node ridge selects its penalty node by node on the validation window, the shared ridge on the pooled validation error. Persistence repeats the same slot of the previous day. The last two rows give the share of the gap between the stronger node-specific reference and the GNN that the shared MLP and the shared ridge recover.}
\label{table:local-refs}
\footnotesize
\begin{tabular*}{\linewidth}{@{\extracolsep{\fill}}lcccc}
\toprule
 & \multicolumn{2}{c}{\dsetstyle{Electricity}} & \multicolumn{2}{c}{\dsetstyle{WEAVE-UK}} \\
Model & $11$ d & $60$ d & $30$ h & $10$ d \\
\midrule
Persistence & $0.5573$ & $0.5768$ & $0.9776$ & $0.9600$ \\
Per-node MLP & $0.3030$ & $\mathbf{0.2776}$ & $1.0509$ & $0.8956$ \\
Per-node ridge & $\mathbf{0.2909}$ & $0.2839$ & $\mathbf{0.9519}$ & $\mathbf{0.8662}$ \\
Shared ridge & $0.2980$ & $0.2942$ & $0.8922$ & $0.8286$ \\
Shared MLP & $0.2804$ & $0.2770$ & $0.9074$ & $0.8013$ \\
GNN & $0.2775$ & $0.2752$ & $0.8808$ & $0.7745$ \\
\midrule
Share recovered by the shared MLP & $79\%$ & -- & $63\%$ & $71\%$ \\
Share recovered by the shared ridge & $<0$ & -- & $84\%$ & $41\%$ \\
\bottomrule
\end{tabular*}
\end{table}

The intervals of Section~\ref{sec:real-regimes} resample nodes after averaging, within each node, the separately scored errors of the $3$ seeds of each model, so they ignore seed variability. Resampling the seeds of each model before the nodes widens them. The edge increment at constant backbone keeps a positive interval on \WEAVEUK, $[+0.0029, +0.0146]$ at $30$ hours and $[+0.0119, +0.0277]$ at $10$ days, and includes zero on \ELECTRICITY, $[-0.0013, +0.0111]$ at $11$ days and $[-0.0023, +0.0109]$ at $60$ days. On \WEAVEUK, every other increment keeps its sign except the backbone change at $10$ days; on \ELECTRICITY every edge-related increment includes zero except the degradation the neighbor mean causes at $11$ days.

Re-estimating the graph inside the $30$-hour \WEAVEUK window, from its training origins alone, gives the GNN an nRMSE of $0.8722$ against $0.8808$ with the graph of the longer record, over the same $3$ seeds and with mutual pairs stored twice as in the main table ($0.8735$ with each edge stored once), and the per-node MLP with the neighbor mean $0.9980$ against $0.9995$. The empty-graph GNN reads no graph, so the edge increment at constant backbone rises from $+0.0086$ to $+0.0172$: the graph of the longer record, if anything, understates the edge gain at $30$ hours.

The rewiring control of Table~\ref{table:rewire} draws one rewired graph per training seed, so the draw and the initialization are confounded. At $30$ hours on \WEAVEUK, we drew $10$ independent rewirings of the graph with each edge stored once and crossed each with the $3$ training seeds. The true graph reaches an nRMSE of $0.8825$, below every rewired graph, which range from $0.8862$ to $0.8954$ with a mean of $0.8905$; on average rewiring removes $116\%$ of the edge gain at constant backbone, $53$--$188\%$ depending on the draw, and $7$ of the $10$ rewired graphs do worse than the empty-graph GNN. On the non-shared path, the same rewirings reproduce $96$--$99\%$ of the gain the neighbor mean brings to the per-node MLP, so that gain does not depend on which neighbors are averaged.

The shared MLP of Table~\ref{table:real-regimes} reads no neighbor. Giving it the neighbor mean completes a factorial design within the MLP family (Table~\ref{table:mlp-square}). On \WEAVEUK, sharing improves the per-node error in both orders, by $+0.1436$ without neighbors and $+0.1054$ with them at $30$ hours, and by $+0.0943$ and $+0.1054$ at $10$ days, all with seed-and-node intervals clear of zero. The shared MLP with the neighbor mean reaches $0.8941$ and $0.7707$, against $0.8808$ and $0.7745$ for the GNN; the differences, $+0.0133$ $[-0.0070, +0.0302]$ and $-0.0038$ $[-0.0101, +0.0016]$, include zero. On \ELECTRICITY, the neighbor mean degrades the shared MLP, by $0.0129$ and $0.0116$, as it degrades the per-node one at $11$ days.

\begin{table}[tp]
\centering
\caption{Per-node error (nRMSE) of the MLP family crossed on sharing and neighbor access, beside the GNN, on the cells of Table~\ref{table:real-regimes}; $3$ seeds, same inputs, graph, width, and training routine. The neighbor mean is that of the per-node MLP with neighbors.}
\label{table:mlp-square}
\footnotesize
\begin{tabular*}{\linewidth}{@{\extracolsep{\fill}}lcccc}
\toprule
 & \multicolumn{2}{c}{\ELECTRICITY} & \multicolumn{2}{c}{\WEAVEUK} \\
Model & $11$ d & $60$ d & $30$ h & $10$ d \\
\midrule
Per-node MLP & $0.3030$ & $0.2776$ & $1.0509$ & $0.8956$ \\
Per-node MLP $+$ neighbors & $0.3201$ & $0.2752$ & $0.9995$ & $0.8762$ \\
Shared MLP & $0.2804$ & $0.2770$ & $0.9074$ & $0.8013$ \\
Shared MLP $+$ neighbors & $0.2933$ & $0.2886$ & $0.8941$ & $0.7707$ \\
GNN & $0.2775$ & $0.2752$ & $0.8808$ & $0.7745$ \\
\bottomrule
\end{tabular*}
\end{table}

The $11$-day \ELECTRICITY cell ends with the record. Repeating it on $4$ disjoint windows ending $30$, $60$, $90$, and $120$ days earlier (Table~\ref{table:periods}) leaves two results unchanged and moves a third. The neighbor mean degrades the per-node model in all $5$ windows, by $0.0107$--$0.0215$, with intervals below zero, and the edge increment at constant backbone stays between $-0.0014$ and $+0.0051$. The share of the per-node-to-GNN gap carried by sharing ranges from $-50\%$ to $111\%$: in the window ending in May $2019$ the shared MLP does worse than the per-node MLP while the GNN does better, so the advantage of the GNN there comes primarily from its backbone.

\begin{table}[tp]
\centering
\caption{Increments of per-node error on the $11$-day \ELECTRICITY cell over $5$ disjoint windows, $3$ seeds; seed-and-node $95\%$ intervals in brackets. The first row is the window of Table~\ref{table:real-regimes}.}
\label{table:periods}
\footnotesize
\begin{tabular*}{\linewidth}{@{\extracolsep{\fill}}lcccc}
\toprule
Window end & Total gap & Sharing (share) & Neighbors, unshared & Edges, same backbone \\
\midrule
Jul 2019 & $+0.026$ & $+0.023$ ($89\%$) & $-0.017$ & $+0.005$ $[-0.001, +0.011]$ \\
Jun 2019 & $+0.017$ & $+0.014$ ($80\%$) & $-0.013$ & $-0.001$ $[-0.005, +0.002]$ \\
May 2019 & $+0.046$ & $-0.023$ ($-50\%$) & $-0.011$ & $+0.004$ $[+0.000, +0.007]$ \\
Apr 2019 & $+0.028$ & $+0.031$ ($111\%$) & $-0.012$ & $+0.001$ $[-0.004, +0.006]$ \\
Mar 2019 & $+0.017$ & $+0.004$ ($23\%$) & $-0.022$ & $-0.001$ $[-0.005, +0.002]$ \\
\bottomrule
\end{tabular*}
\end{table}

On the same CPU, fitting the GNN of a cell takes $42$--$365$ seconds per seed on average, against $2$--$14$ for the shared MLP, so the GNN takes $12$--$35$ times longer to train than the shared MLP.

\section{Descriptive Analysis of Learned Representations}\label{app:gnn-repr}

The main text compares the graph with its per-node reference model and decomposes the observed gaps. This block describes what the fitted GNNs encode. It does not supply further evidence for the graph's predictive contribution. In the single-step regression framework, we inspect localized graph explanations; in the sequence-to-sequence setting, we compare latent representations and analyze the attention mechanisms emerging from temporal and spatial dependencies.

Beyond forecasting accuracy, a trained model can be inspected with interpretability tools to probe which spatial and feature-level factors it relies on. We treat these tools as a complementary analysis and read their output with caution, since post-hoc explanations remain indicative \citep{jain2019attention}.

\paragraph{GNNExplainer.} GNNExplainer \citep{ying2019gnnexplainer} identifies the subgraphs and node features that influence a model's decision the most. The method optimizes the mutual information between the forecast and a compact explanatory subgraph, distinguishing relevant from irrelevant connections. In practice, this produces a mask over the adjacency matrix and feature space, which reads as the regional links and the covariates used by the prediction. For electricity demand, GNNExplainer indicates which of the links encoded in the input graph a forecast relies on, for instance between climatic zones with similar load behaviors; it cannot reveal a relation that the graph does not contain.

\paragraph{Attention-based interpretability.} Attention mechanisms in GATs assign learnable weights to neighbors during message passing. These coefficients describe how the fitted layer combines its inputs and can vary over time; they are not, by themselves, measures of predictive importance \citep{jain2019attention}. We first project sequences of attention matrices with principal component analysis (PCA) or uniform manifold approximation and projection (UMAP) to identify recurrent temporal patterns. We then test predictive importance directly by retaining edges according to their attention rank and measuring the resulting forecast error under a fixed downstream model.

\begin{figure}[tp]
    \centering
    \includegraphics[width=0.72\linewidth]{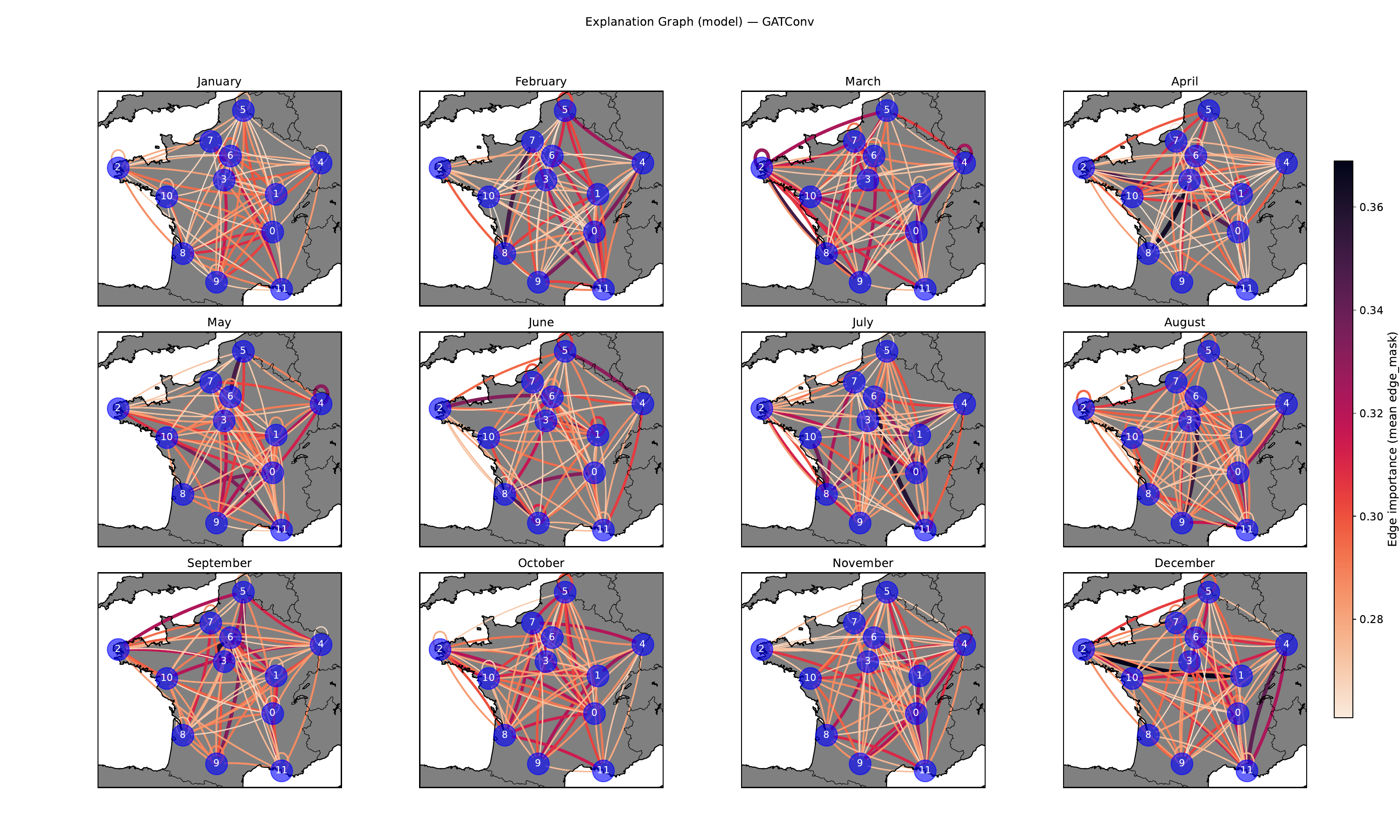}
    \caption{Explanation graphs for the GAT convolutional layer on \REGIONSFR.}
    \label{fig:exp-graphs}
\end{figure}

\paragraph{Single-step regression setup.}
To interpret GNNs trained on \REGIONSFR, we use GNNExplainer \citep{ying2019gnnexplainer}, which identifies the subgraph most influential for a prediction. For each month, daily explanatory subgraphs are aggregated by summing and normalizing their edge masks over all half-hour intervals. The learned mask $\mathbf{M}$ defines the relevance-weighted subgraph $\bW_\lambda \odot \sigma(\mathbf{M})$, where $\odot$ is the Hadamard product and $\sigma$ maps the mask values to $[0,1]^{N \times N}$. Figure~\ref{fig:exp-graphs} reports these monthly aggregates for GATConv on the real regional dataset.

The resulting subgraphs show recurrent spatial patterns. Connections along the Atlantic seaboard and among Hauts-de-France, Normandie, and Île-de-France receive high relevance in several months. These masks describe which edges the fitted GATConv uses for its predictions; they do not by themselves establish that the connected regions contain information absent from each region's own history, which is the predictive question examined in the main text.

\begin{figure}[tp]
    \centering
    \includegraphics[width=\linewidth]{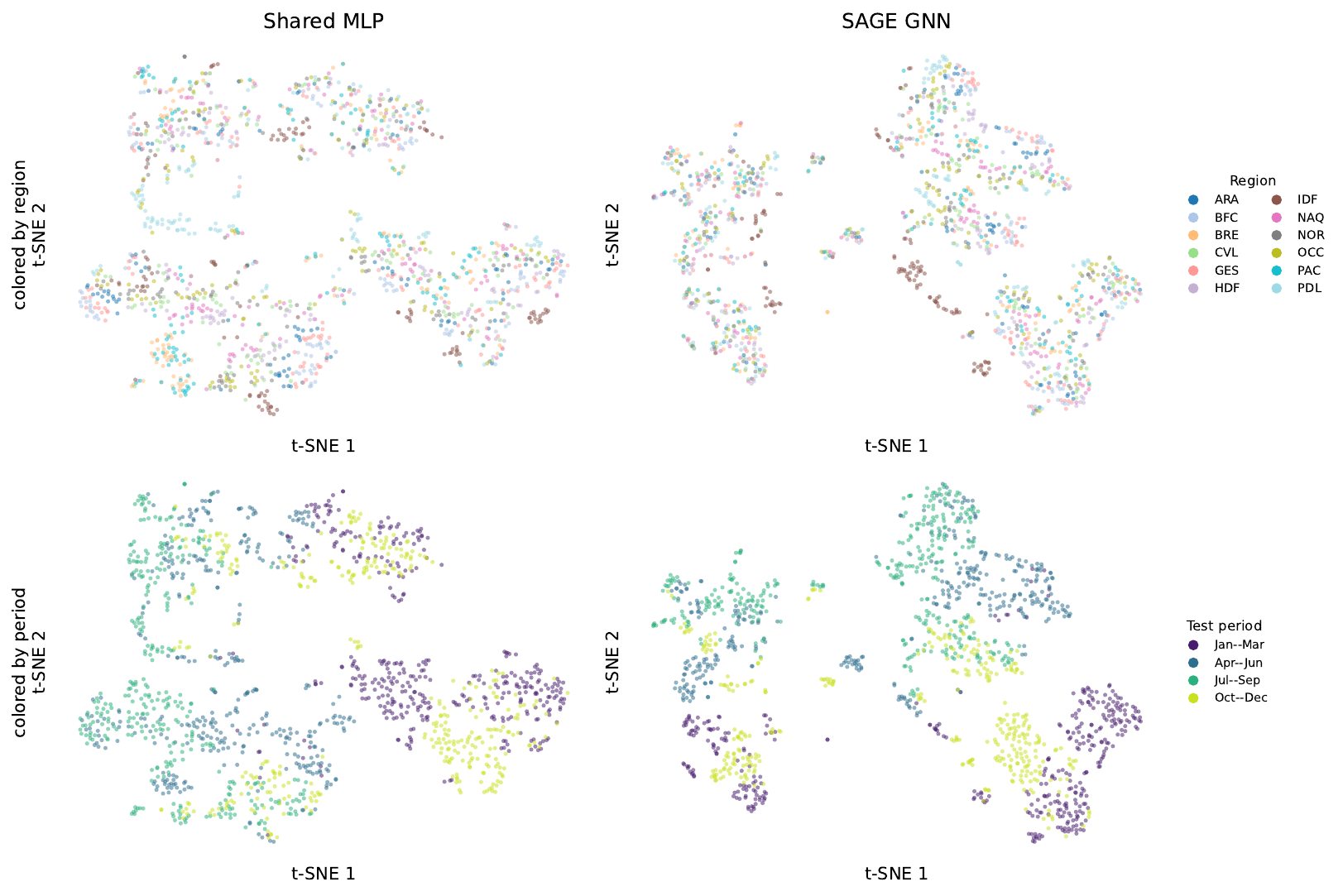}
    \caption{t-SNE projections of the final $64$-dimensional latent representations of the matched shared MLP and SAGE GNN on the \REGIONSFR test set. Every third test day and all $12$ regions are retained, giving $1452$ identical observations in each plot; the upper row colors them by region and the lower row by quarter of the test period. The projections are fitted separately with perplexity $40$ and a fixed seed, so only within-plot neighborhoods are read.}
    \label{fig:mlp-gnn-tsne}
\end{figure}

\paragraph{Sequence-to-sequence forecasting setup.}
The shared MLP and the SAGE GNN can be compared before the readout, where both return a $64$-dimensional representation for each pair formed by a test day and a region. They have the same encoder architecture, depth and width; the shared MLP is the identity-adjacency model and SAGE alone aggregates the spatial graph. Figure~\ref{fig:mlp-gnn-tsne} projects the same $1452$ pairs into each space. Each t-SNE is fitted separately, so its axes, orientation, and distances to the other plot carry no meaning; the plots show only the neighborhood geometry internal to one model.

The visual similarity is borne out in the original spaces. After each latent coordinate is standardized over the test observations, linear CKA is $0.945$. Region centroids account for $12.1\%$ of the shared MLP's latent variance and $9.9\%$ of SAGE's, while the $4$ periods account for $25.9\%$ and $26.1\%$. The neighborhood therefore changes the representation less than the seasonal position common to both models, and slightly reduces the separation between regions. On these $12$ richly observed nodes, SAGE does not learn a new spatial organization absent from the shared MLP; this agrees with the small predictive increment assigned to the edges in Section~\ref{sec:mlp-reference}.

\begin{figure}[tp]
    \centering
    \includegraphics[width=0.92\linewidth]{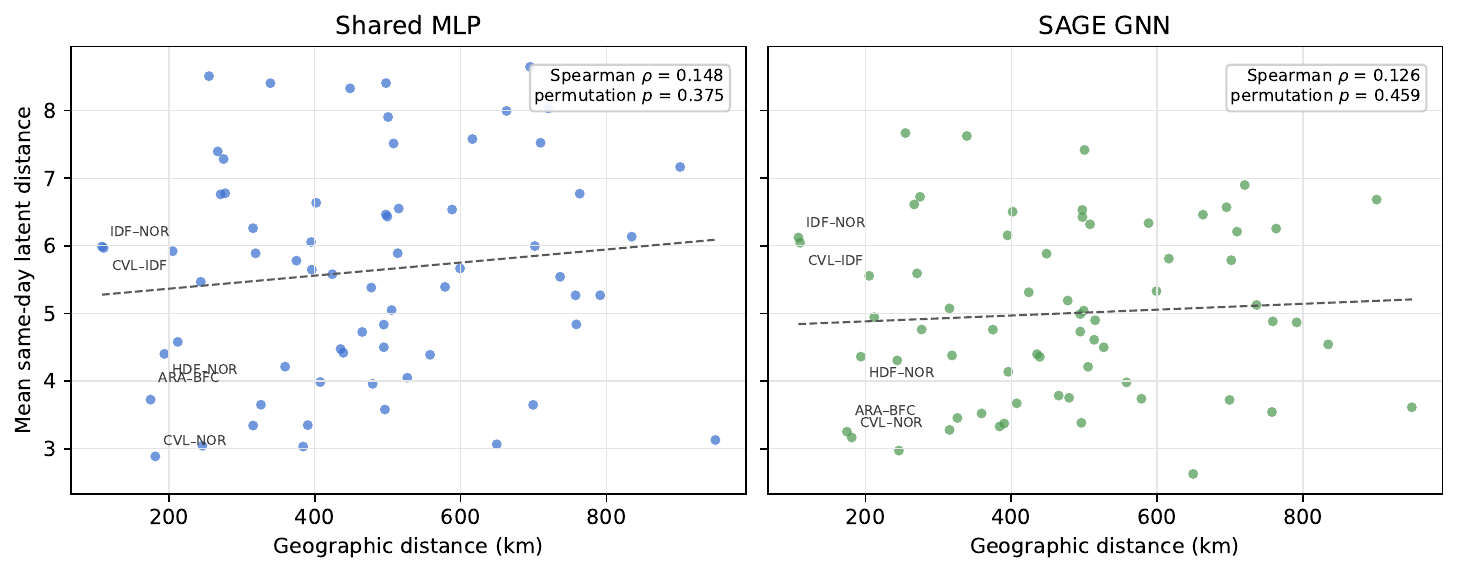}
    \caption{Geographic distance against mean same-day latent distance for all $66$ pairs of regions, in the original standardized representation spaces. The dashed line is a linear visual guide; the reported statistic is Spearman's rank correlation, with a two-sided $p$-value obtained from $9999$ permutations of the region labels to preserve the dependence among pairwise distances. The $5$ pairs separated by less than $200$~km are labeled.}
    \label{fig:france-latent-geography}
\end{figure}

\paragraph{Does latent proximity follow geographic proximity?}
A t-SNE projection cannot answer this directly because it preserves local neighborhoods selectively and distorts global distances. Figure~\ref{fig:france-latent-geography} therefore returns to the standardized $64$-dimensional spaces. For each of the $66$ pairs of regions, it averages their Euclidean latent distance over the same $362$ test days and compares that value with the great-circle distance between the regional reference points used to build the spatial graph.

No geographic ordering is detectable: Spearman's $\rho$ is $0.148$ for the shared MLP and $0.126$ for SAGE, with permutation $p$-values of $0.375$ and $0.459$. Some close pairs do have close representations, including Auvergne-Rhône-Alpes and Bourgogne-Franche-Comté, but this relation does not extend across the $66$ pairs. The regional colors in Figure~\ref{fig:mlp-gnn-tsne} must consequently be read as identities, not as evidence that the latent space reproduces the map. The adjacency makes that reading conservative rather than fragile. SAGE is taken here on the spatial graph; Table~\ref{table:gs1and2} tunes it on the DTW graph, precisely because geography is then the one structure message passing pushes the representation toward: two regions joined by an edge exchange states at every layer, so a latent space ordered by distance would be the expected outcome and not a surprise. It does not order itself that way, and calendar, weather, and each region's load history still determine most of that geometry.

\begin{figure}[tp]
    \centering
    \includegraphics[width=\linewidth]{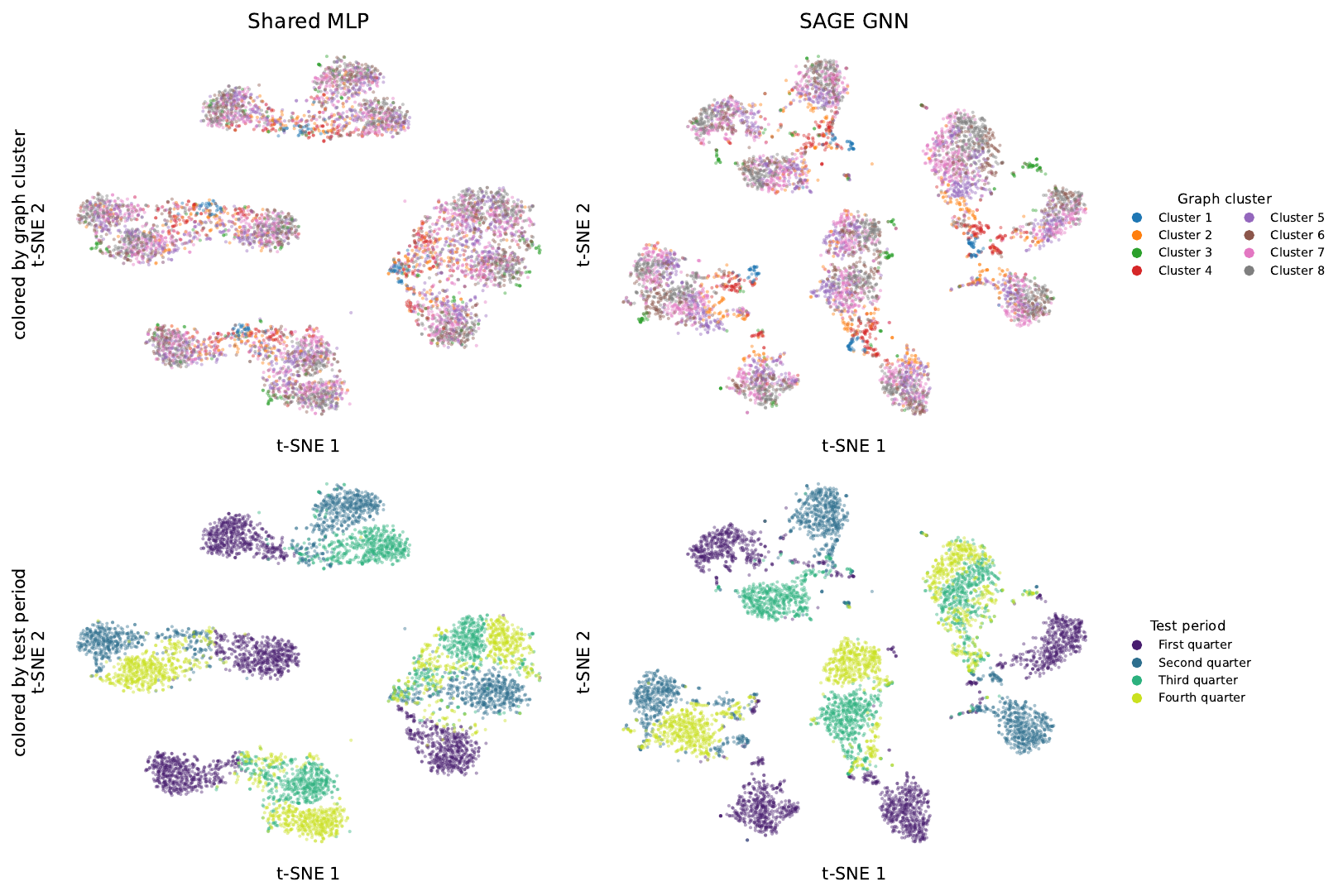}
    \caption{t-SNE projections of the final $64$-dimensional states of the shared MLP and SAGE GNN on \WEAVEUK at $500$ feeders. All feeders and every 12th test origin are retained, giving $6500$ identical observations per plot; the upper row colors them by $8$ spectral clusters of the fixed graph and the lower row by quarter of the test window. The projections are fitted separately with perplexity $50$ and a fixed seed, so only within-plot neighborhoods are read.}
    \label{fig:weave-mlp-gnn-tsne}
\end{figure}

\paragraph{\WEAVEUK at $500$ feeders.}
The same extraction is applied to the shared MLP and SAGE fits of the $500$-feeder, $10$-day cell in Table~\ref{table:node-count}. Both expose their final $64$-dimensional state before the $48$-output readout, yielding $145 \times 500 = 72500$ matched pairs of a test origin and a feeder. Spectral clustering of the fixed adjacency, computed once, gives $8$ clusters that provide a readable graph label where $500$ feeder identities would not. Figure~\ref{fig:weave-mlp-gnn-tsne} projects both spaces with these labels.

Here the coordinate-standardized spaces differ more than on \REGIONSFR: linear CKA is $0.847$ against $0.945$ there. Feeder centroids explain little variance in either model, $2.9\%$ for the shared MLP and $3.1\%$ for SAGE, and the $4$ test periods explain $8.0\%$ and $6.3\%$. After the same standardization, the mean squared distance between retained graph neighbors, divided by the corresponding distance to random feeders, falls from $0.631$ in the shared MLP to $0.475$ in SAGE. Linked feeders thus have more similar representations in SAGE. The associated predictive comparison is from $0.8013$ to $0.7745$ in per-node error (Table~\ref{table:node-count}). Since the \WEAVEUK backbones also differ, neither the representation difference nor the prediction gap can be attributed solely to message passing.

\begin{figure}[tp]
    \centering
    \includegraphics[width=\linewidth]{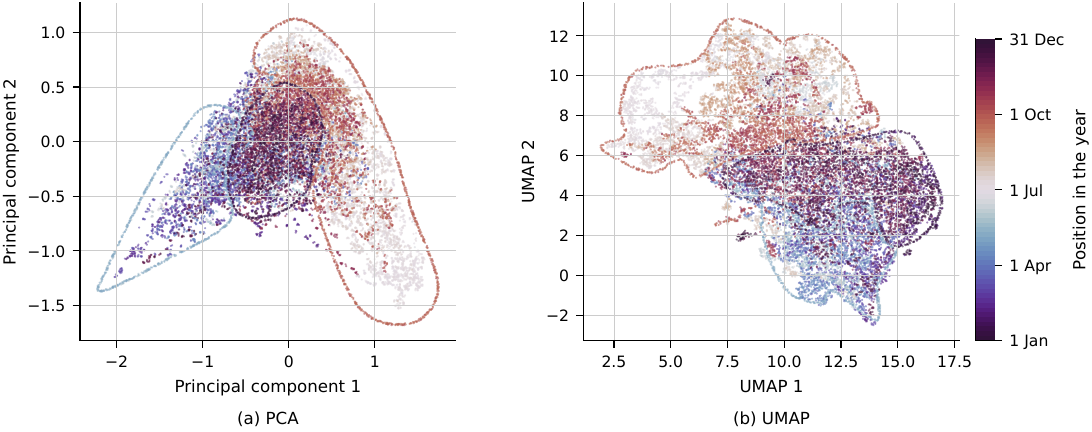}
    \caption{Two projections of the GAT attention matrices on \REGIONSFR, first layer and first head, one point per half-hour of the test year and colored by its position in that year. The partial separation of successive periods places the attention matrices along the seasonal cycle, which moves temperature, daylight, and activity together.}
    \label{fig:pca-umap-comparison}
\end{figure}

In the multi-step setting, we read the internal attention mechanisms of GAT-based architectures to gain insight into how temporal and spatial dependencies are encoded. For each time step, the model computes a full attention matrix describing how the layer weights inter-regional inputs for forecasting the next day's load. To analyze the latent structure of these matrices, we apply PCA and UMAP, which respectively capture linear variance and nonlinear manifold structure. As shown in Figure~\ref{fig:pca-umap-comparison}, both projections order the matrices by season, winter, summer, and transitional periods occupying distinct regions. The season moves temperature, daylight, and activity together, so the projection does not say which of them the attention follows.

A third synthetic generator, distinct from the denoising generator of Section~\ref{sec:synth-regimes} and from the generator of Section~\ref{app:aggdiag}, serves to study attention mechanisms. It produces $N$ node-level time series formed by a smooth trend, a periodic component, and strong high-frequency noise to prevent trivial per-node fitting. Node interactions are introduced by multiplying the input matrix $\bX \in \mathbb{R}^{N \times T}$ with a coupling matrix $\bA \in \mathbb{R}^{N \times N}$. Three variations are generated: \emph{single coupling}, where a fixed, sparse, full-rank matrix $\bA$ is applied at all time steps; \emph{explicit switching}, where two distinct matrices $\bA_1$ and $\bA_2$ are alternated according to a binary switching signal provided as an additional input; and \emph{ambiguous switching}, where the same alternation occurs, but two categorical inputs encode the active coupling only jointly, mimicking latent factors that are not directly observable. The supervised task is to reconstruct the original signals, so that predictive performance relies on the inter-node dependencies encoded in $\bA$.

\begin{figure}[tp]
    \centering
    \includegraphics[width=\linewidth]{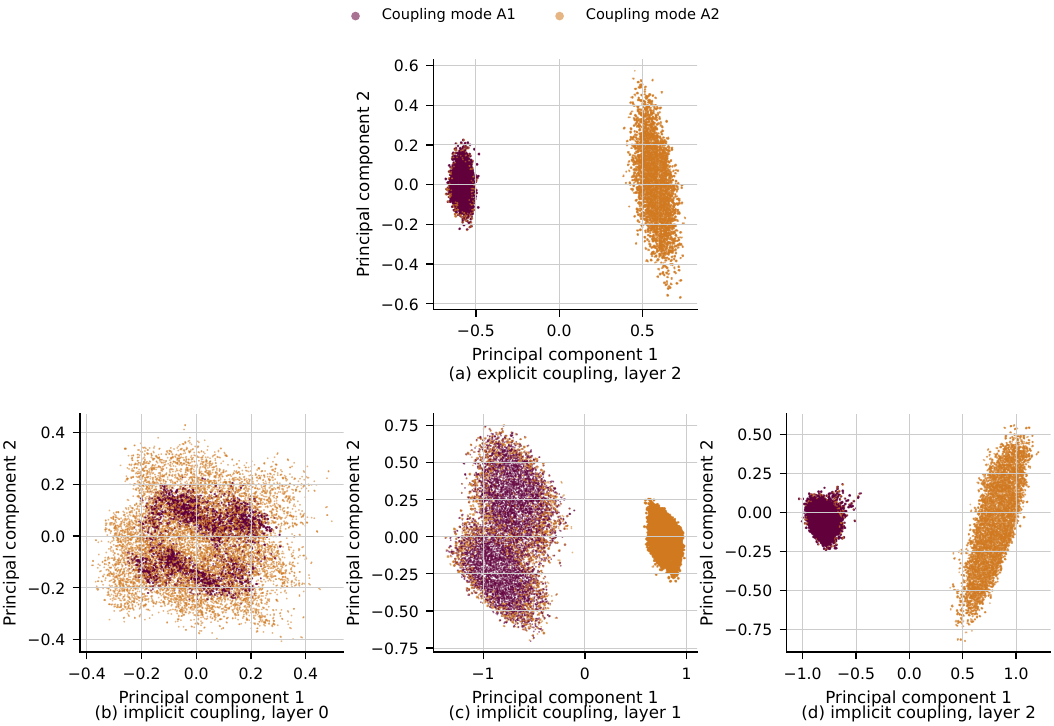}
    \caption{PCA of the attention matrices on the synthetic dataset with two coupling modes $\bA_1$ and $\bA_2$, colored by the mode in effect. In (a), the mode is handed to the model as a binary input and the two clusters separate at once. In (b) to (d), it is only implied by the joint value of two binary inputs, and the separation is built layer by layer: the first layer mixes the two modes, the second splits them partially, the third recovers them.}
    \label{fig:pca-hidden-coupling}
\end{figure}

We also inspect attention on synthetic data with known coupling matrices. With a single coupling, the weights are stable over time but do not reproduce the true adjacency. When two coupling modes alternate and the active mode is supplied as a binary input, the projected attention vectors form two clusters (Figure~\ref{fig:pca-hidden-coupling}). When the mode is only implied by two categorical inputs, the first layer mixes the modes and the later layers separate them progressively. The attention matrices therefore encode the active regime in this generator. This projection does not establish that they recover the physical adjacency or that the largest coefficients identify the most useful edges; Section~\ref{sec:attn-prune} tests the latter claim directly.

\subsection{Selection by Attention Weights}\label{sec:attn-prune}

The projections above show that attention moves with the temporal regime. They do not show that the edges carrying the largest coefficients improve the forecast. That narrower claim is tested here by deleting edges and measuring what the forecast loses, on the \WEAVEUK $500$-feeder, $10$-day cell of Table~\ref{table:node-count}. A GATv2 is fitted on the full retained graph, which holds $4712$ distinct directed edges ($5000$ with mutual pairs counted twice), about $10$ in-neighbors per node, and each edge receives one score: its attention coefficient averaged over the $145$ test origins. Averaging over the test origins makes the ranking retrospective: it describes the fitted model and is not a selection rule fixed before evaluation.

\begin{figure}[tp]
\centering
\begin{tikzpicture}[
  font=\small,
  src/.style={circle, draw=gray!60, fill=white, line width=0.5pt, minimum size=2.0mm, inner sep=0pt},
  tgt/.style={circle, draw=nodeblue, fill=nodeblue!25, line width=0.7pt, minimum size=4.4mm, inner sep=0pt},
]
\node[font=\bfseries\footnotesize] at (3.2,3.15) {Global ranking: degree is confounded};
\node[tgt] (li) at (4.3,2.25) {\tiny $i$};
\node[tgt] (lj) at (4.3,0.75) {\tiny $j$};
\node[anchor=west, font=\scriptsize, nodedark] at (4.62,2.25) {$d_i=3$};
\node[anchor=west, font=\scriptsize, nodedark] at (4.62,0.75) {$d_j=6$};
\foreach \y in {2.05,2.35,2.65}{\node[src] (a\y) at (1.7,\y) {};
  \draw[nodegreen, line width=1.0pt] (1.82,\y) -- (li);}
\foreach \y in {-0.05,0.27,0.59,0.91,1.23,1.55}{\node[src] (b\y) at (1.7,\y) {};
  \draw[gray!60, line width=0.6pt] (1.82,\y) -- (lj);}
\node[font=\scriptsize, nodegreen!80!black] at (2.95,1.98) {$\alpha \approx 1/3$};
\node[font=\scriptsize, gray!70!black] at (2.95,0.15) {$\alpha \approx 1/6$};
\node[font=\scriptsize, gray!70!black] at (3.2,-0.60) {one threshold keeps the low-degree node and empties the other};

\draw[gray!45, line width=0.4pt] (6.55,-0.35) -- (6.55,3.25);

\node[font=\bfseries\footnotesize] at (10.4,3.15) {Within-node pruning: edge identity is isolated};
\node[tgt] (ri) at (11.5,2.25) {\tiny $i$};
\node[tgt] (rj) at (11.5,0.75) {\tiny $j$};
\node[anchor=west, font=\scriptsize, nodedark] at (11.82,2.25) {$d_i=3$};
\node[anchor=west, font=\scriptsize, nodedark] at (11.82,0.75) {$d_j=6$};
\foreach \y/\c in {2.05/nodegreen,2.35/nodegreen,2.65/gray}{\node[src] at (8.9,\y) {};
  \draw[\c!70!black, line width=0.8pt] (9.02,\y) -- (ri);}
\foreach \y/\c in {-0.05/nodegreen,0.27/nodegreen,0.59/nodegreen,0.91/gray,1.23/gray,1.55/gray}{\node[src] at (8.9,\y) {};
  \draw[\c!70!black, line width=0.8pt] (9.02,\y) -- (rj);}
\node[font=\scriptsize, gray!70!black] at (10.4,-0.60) {the same share is retained at every node};
\end{tikzpicture}
\caption{Why pruning is performed separately for every target node. A global ranking mostly separates node degrees because attention is normalized over each incoming neighborhood. Within-node pruning fixes the retained degree and changes only the identity of the retained edges. Green edges are kept.}
\label{fig:attention-pruning-protocol}
\end{figure}
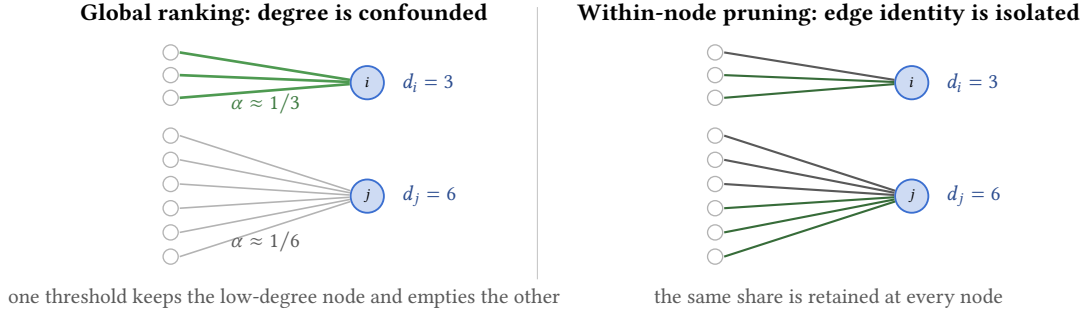

Those scores cannot be compared across target nodes. Attention is normalized separately over the edges entering each node, so a mechanism close to uniform gives about $1/d_i$ to every edge entering a node of in-degree $d_i$, and the score then reports the degree rather than the edge. Two measurements place the fitted model in that regime. Dividing the entropy of the attention entering a node by $\log d_i$ gives $1$ when the node weighs its neighbors equally and $0$ when it loads one of them; averaged over the $500$ feeders of \WEAVEUK, that ratio is $0.989$, $0.990$, and $0.991$ on the $3$ seeds. The coefficients themselves correlate at $0.94$ with $1/d_i$. One threshold applied to all $5000$ scores would therefore keep the neighborhoods of low-degree nodes and empty those of high-degree ones, comparing degree profiles instead of edges. Figure~\ref{fig:attention-pruning-protocol} contrasts that selection with the within-node selection used below.

Every node keeps a fixed share of its incoming edges: the highest-ranked, the lowest-ranked, or a draw that ignores the ranking. The three variants of a row therefore have the same in-degree at every node and differ only in the identity of the retained edges.

GATv2 is used only to rank the edges. It is not retrained on the pruned graphs. For each subgraph, we instead fit one MLP per feeder. The reference MLP receives that feeder's own features and reaches $0.8956$ nRMSE, the per-node value Tables~\ref{table:real-regimes} and~\ref{table:node-count} report for this cell. The evaluated MLP receives the same features plus one additional vector, the mean of the retained neighbors' features. Since both models remain node-specific, their comparison changes neighborhood access without introducing parameter sharing or a learned aggregation. Table~\ref{table:attn-prune} reports the test nRMSE of this second model directly; lower values indicate better forecasts.

\begin{table}[tp]
\centering
\caption{Direct test nRMSE, lower is better, of the per-node MLP augmented by the mean features of its retained neighbors. Values are averaged over $3$ seeds. Each value in parentheses gives the relative change from the row above to the row below, $100 (E_\text{below} - E_\text{above}) / E_\text{above}$. Red positive values denote an increase in error; green negative values denote an improvement. GATv2 ranks the edges on the full retained graph; it does not produce the errors in the table. The full-graph and no-neighbor values are repeated across the three columns because edge ranking is irrelevant when all or no edges are retained. Seed dispersion is $0.0003$--$0.0033$.}
\label{table:attn-prune}
\footnotesize
\begin{tabular}{lcccc}
\toprule
Edges kept & Count & Highest attention & Random & Lowest attention \\
\midrule
All edges     & $5000$ & $0.8763$ & $0.8763$ & $0.8763$ \\
One half      & $2412$ & $0.8768$ \worse{$+0.06\%$} & $0.8835$ \worse{$+0.82\%$} & $0.8912$ \worse{$+1.70\%$} \\
One quarter   & $1235$ & $0.8862$ \worse{$+1.07\%$} & $0.8938$ \worse{$+1.17\%$} & $0.9079$ \worse{$+1.87\%$} \\
One tenth     & $654$  & $0.8931$ \worse{$+0.78\%$} & $0.9031$ \worse{$+1.04\%$} & $0.9218$ \worse{$+1.53\%$} \\
No neighbors & $0$    & $0.8956$ \worse{$+0.28\%$} & $0.8956$ \better{$-0.83\%$} & $0.8956$ \better{$-2.84\%$} \\
\bottomrule
\end{tabular}
\end{table}

The full retained neighborhood improves nRMSE from $0.8956$ to $0.8763$. Keeping the best-ranked half leaves that performance almost unchanged at $0.8768$; a random half reaches $0.8835$, and the least-ranked half $0.8912$. The same ordering holds at one quarter and one tenth. Thinning eventually removes too much information: the best tenth reaches $0.8931$, close to the no-neighbor reference, while the random and least-ranked tenths perform worse than that reference.

The non-shared rewiring control (preceding section) gives the comparison at full density, about $10$ in-neighbors: from the $0.8956$ reference, arbitrary neighbors reach $0.8816$ against $0.8762$ for the true graph ($0.8763$ in Table~\ref{table:attn-prune}). At half density, a random half of the true edges reaches $0.8835$ and the attention-selected half $0.8768$. Under this fixed mean-summary protocol, generic neighbor averaging explains most of the improvement and the attention ranking accounts for a smaller additional difference.

\paragraph{Architecture-dependent control.}
The test in Table~\ref{table:attn-prune} deliberately fixes how the subgraphs are used: whatever the retained edges, each feeder receives the arithmetic mean of its neighbors' features. It therefore measures whether the GATv2 ranking selects useful inputs when the downstream model and its neighbor summary are held constant. The ordering is clear under this controlled protocol: highest-ranked edges outperform random edges, which outperform lowest-ranked edges. A final test determines whether the same ordering persists when a GNN is trained end to end on each pruned topology. SAGE is trained from scratch on the highest-ranked, random and lowest-ranked quarter-graphs and reaches respectively $0.7811$, $0.7790$, and $0.7817$ nRMSE. Random pruning is numerically best, but the largest gap ($0.0027$) is of the same order as the seed dispersion ($0.0020$); these runs therefore do not establish a reliable ordering among the three graphs. This result does not invalidate Table~\ref{table:attn-prune}: it limits its scope. The attention scores identify useful neighbors for the fixed mean-summary experiment, but they do not define a subgraph that every architecture will rank in the same way after retraining.

\section{Hyperparameters of the Benchmarked Architectures}\label{app:gnn-hp}

Optuna \citep{akiba2019optuna} tunes the depth, width, batch size, and learning rate of every architecture, with the order of ChebConv and TAGConv, the heads of GAT, GATv2, and TransformerConv, and the steps $K$ and teleportation probability $\alpha$ of APPNP. Each architecture trains for $300$ epochs and the epoch with the lowest validation loss is restored. On both datasets, the graph is searched jointly with the other hyperparameters, among the $6$ constructions on \REGIONSFR (spatial, DTW, correlation, precision, DistSplines, GL3SR) and, on \WEAVEUK, among correlation, DTW, and precision graphs in which each node keeps its $5$ most similar nodes before symmetrization, for mean degrees $6.4$--$9.6$. The TPE sampler (seed $42$) runs $40$ trials per architecture, each trained for $200$ epochs with patience $40$ and scored on the validation RMSE of the summed forecast, over depths $1$--$3$, widths $32$--$256$, and learning rates $10^{-4}$--$10^{-2}$ (both log-scaled), batch sizes $\{8,16,32\}$ on the regions and $\{2,4,8\}$ on the feeders, heads $\{1,2,4\}$, orders $K\le4$ (TAG) and $K\le5$ (ChebConv), and APPNP steps $K\in\{2,4,6,8,10\}$ with $\alpha\in\{0.05,0.1,0.15,0.2,0.3\}$. On both datasets, the two MLPs and the pooled XGBoost receive the same $40$-trial search; XGBoost (b-local) keeps fixed settings (learning rate $0.1$, depth $4$, $120$ trees), and the additive model and ARIMA are not tuned. ARIMA-X fits one model per half-hour on the daily national subseries of \REGIONSFR, with weather, heating and cooling degree terms, holidays, weekday, and position in the year as exogenous inputs; its ARMA orders are selected by AIC on $2015$--$2017$, and its exogenous design is chosen among $4$ on the $2018$ validation year (from $1960$~MW with linear weather terms to $1506$~MW), then re-estimated on $2015$--$2018$ and applied $1$ day ahead with frozen parameters, its weather inputs being the observations averaged over the regions. On \WEAVEUK, whose $9$ training days leave $9$ points per half-hour and which has no weather, a single ARIMA-X runs on the half-hourly summed load with daily Fourier terms and a weekend indicator, forecasting $48$ steps from each origin with frozen parameters (retained: order $(3,1,2)$, $6$ harmonics); its order is selected by AIC on the training days, its design among $3$ on the validation days, and it is re-estimated on both before the test. Chronos-2, t0-gamma, and Timestral are used zero-shot under a scheme fixed in advance, one $48$-step median forecast of the summed load at each daily origin. t0-gamma was scored in $2$ settings fixed before scoring, without post-processing (reported) and with the $4$ adjustments its model card recommends ($1747$~MW and $149.0$~kWh); Timestral keeps its package defaults. Variants run as side information score worse on the test: on \WEAVEUK, one model per feeder ($224.9$~kWh) and the $500$ feeders read as one multivariate series ($198.3$~kWh); on \REGIONSFR, the $12$ regions read as one multivariate series ($1769$~MW). The MLPs run $10$ seeds, the gradient-boosted models and GNNs 3. The retained configurations are listed in Table~\ref{table:gs1and2} of the Appendix. The diagnostic comparisons use prescribed model classes without full hyperparameter searches, the limited validation-based selections being specified in each block. Their widths and backbones are stated in each experimental block; common width should not be read as matched total capacity.

Table~\ref{table:gs1and2} lists, for each of the 8 convolutions of Table~\ref{table:combined-results} and each dataset, the graph construction and the hyperparameters retained by validation in the sequence-to-sequence setup.

\begin{table}[tp]
\centering
\caption{Hyperparameters of the GNN models for \REGIONSFR and \WEAVEUK in the sequence-to-sequence forecasting setup. The \emph{Graph} column gives the construction of Section~\ref{sec:graph-construction} retained by validation; on \REGIONSFR the dense constructions share one (near-)complete support, which SAGE, GAT, GATv2, and Transformer read identically (Section~\ref{sec:graph-construction}).}
\label{table:gs1and2}
\footnotesize
\begin{tabular}{llccccl}
\toprule
Dataset & Model & Graph & Batch size & Layers & Width & Learning rate \\
\midrule
\multirow{8}{*}{\REGIONSFR}
 & GCN         & Space & $8$ & $3$ & $68$ & $2 \cdot 10^{-3}$ \\
 & SAGE        & DTW & $8$ & $2$ & $196$ & $1 \cdot 10^{-3}$ \\
 & GAT         & Space & $8$ & $2$ & $75$ & $2 \cdot 10^{-3}$ \\
 & GATv2       & Space & $8$ & $3$ & $155$ & $6 \cdot 10^{-4}$ \\
 & Transformer & Correlation & $8$ & $2$ & $88$ & $2 \cdot 10^{-3}$ \\
 & TAG         & Space & $8$ & $3$ & $98$ & $4 \cdot 10^{-3}$ \\
 & ChebConv    & Space & $8$ & $2$ & $105$ & $6 \cdot 10^{-4}$ \\
 & APPNP       & GL3SR & $8$ & $2$ & $189$ & $7 \cdot 10^{-4}$ \\
\midrule
\multirow{8}{*}{\WEAVEUK}
 & GCN         & Correlation & $2$ & $1$ & $86$ & $3 \cdot 10^{-3}$ \\
 & SAGE        & Correlation & $2$ & $1$ & $126$ & $7 \cdot 10^{-3}$ \\
 & GAT         & Precision & $2$ & $3$ & $76$ & $4 \cdot 10^{-3}$ \\
 & GATv2       & Precision & $2$ & $1$ & $102$ & $9 \cdot 10^{-3}$ \\
 & Transformer & DTW & $4$ & $2$ & $115$ & $1 \cdot 10^{-3}$ \\
 & TAG         & Precision & $2$ & $3$ & $123$ & $1 \cdot 10^{-3}$ \\
 & ChebConv    & Precision & $8$ & $1$ & $54$ & $6 \cdot 10^{-3}$ \\
 & APPNP       & Precision & $2$ & $3$ & $94$ & $1 \cdot 10^{-2}$ \\
\bottomrule
\end{tabular}

\smallskip
\footnotesize
Architecture-specific hyperparameters. \REGIONSFR: GAT and GATv2 use $1$ attention head and Transformer $4$; TAG uses $K=4$; ChebConv uses $K=2$; APPNP uses $(K,\alpha)=(4,0.05)$. \WEAVEUK: GAT uses $4$ heads, GATv2 $2$, and Transformer $1$; TAG uses $K=2$; ChebConv uses $K=1$; APPNP uses $(K,\alpha)=(2,0.05)$. With $K=1$, ChebConv performs no propagation on \WEAVEUK.
\end{table}

\section{Code and Data Availability}\label{app:code}

The scripts and configurations that run the experiments of this article are available on Github\footnote{\url{https://github.com/eloicampagne/gnn4elf}}. They build on the \textit{GraphToolbox} \citep{campagne2026graphtoolbox}, an open-source Python package for GNN forecasting that implements data preprocessing, graph construction, and training, itself built on the \emph{PyTorch Geometric} library \citep{fey2019fast, fey2025pyg}. The processed datasets (\REGIONSFR and \WEAVEUK) and the synthetic benchmarks are deposited on Zenodo \citep{campagne2026weave}.\footnote{\url{https://doi.org/10.5281/zenodo.21977576}}

\end{document}